\documentclass[letterpaper, 10 pt, conference]{ieeeconf}  % Comment this line out if you need a4paper

\IEEEoverridecommandlockouts                              % This command is only needed if 
\usepackage{booktabs}
\usepackage{graphicx}
\usepackage{subcaption}
\usepackage{graphbox}
\usepackage[export]{adjustbox}
\usepackage{tikz}

\usepackage{amssymb}
\usepackage{amsmath}

\usepackage{hyperref}
\usepackage{cleveref}
\creflabelformat{equation}{#2#1#3} % remove parentheses around equation reference

\usepackage{multirow}

\usepackage{siunitx}

\title{\LARGE \bf
VLEM: Real-Time 3D Vision-Language Embedding Mapping
}

\author{Christian Rauch, Björn Ellensohn, Linus Nwankwo, Vedant Dave, Elmar Rueckert%
\thanks{All authors are with the Technical University of Leoben, Franz Josef-Straße 18, 8700 Leoben, Austria. $^{*}$Corresponding author: \texttt{Christian.Rauch@unileoben.ac.at}.}%
}

\begin{document}

\maketitle
\thispagestyle{empty}
\pagestyle{empty}

\begin{abstract}

% A metric-accurate semantic 3D representation is essential for many robotic tasks. This work investigates strategies to integrate 2D vision-language embeddings in real-time into a 3D representation for various robotic tasks. We provide a framework to integrate image-embeddings, as well as vision transformer patch-embeddings, and analyse the effect of different Vision-Language Model embeddings and their suitability for real-time applications. Based on our analysis, this work proposes a simple, yet powerful, method to integrate the 2D vision-language embeddings in a metric-accurate 3D representation in real-time for robot tasks that require the localisation of objects-of-interest via natural language.

% We evaluate our approach against the real-time approaches Open-Fusion, RayFronts, and a variation of ConceptFusion, on ScanNet and custom RGB-D sensor sequences, and demonstrate its efficiency and versatility in resource-constrained robotic tasks, such as interactive handheld mapping, mobile robotics, and manipulation tasks.

Semantic scene understanding in robotics requires representations that are both metric-accurate and queryable via natural language in real-time. While recent Vision-Language Models enable powerful 2D image-text alignment, their integration into real-time 3D mapping systems remains challenging due to their requirements on ground truth poses, computational cost, and memory constraints. We present VLEM (Vision-Language Embedding Mapping), a real-time framework for integrating pixel-aligned 2D vision-language embeddings from various backends into a globally consistent, metric-accurate 3D representation, requiring only a raw RGB-D stream.
Compared to ConceptFusion, Open-Fusion, and RayFronts, VLEM provides better open-set segmentation performance and a more compact representation.
We further demonstrate VLEM's versatility in interactive real-time robotic manipulation tasks and mobile mapping scenarios.

% We evaluate VLEM against ConceptFusion, Open-Fusion, RayFronts, and LangSplat, on ScanNet and custom RGB-D sequences and further validate its versatility in interactive real-time robotic manipulation tasks and mobile mapping scenarios.

% Extensive evaluation on ScanNet and custom RGB-D sequences demonstrates that VLEM achieves a superior trade-off between segmentation performance, spatial embedding resolution, and resource consumption compared to ConceptFusion, Open-Fusion, RayFronts, and LangSplat.

% We further validate our approach in real-time robotic manipulation and mobile mapping scenarios, showing that VLEM enables online open-set object localisation for interactive task execution via natural language queries.

% Website: \url{https://rtvlem.github.io}.

\end{abstract}

\section{Introduction}

Many robotic downstream tasks, such as manipulation and navigation, require a metric-accurate understanding of the 3D environment as well as semantic information about objects and places to interact with. Classically, this problem has been approached by closed-set semantic mapping~\cite{McCormac2017,Brasch2018,Schönberger2018}, combining semantic segmentation with dense 3D reconstruction \cite{Whelan-RSS-15}.
Closed-set segmentation uses a fixed set of classes and labels, limiting the application of these approaches to certain domains and specific query patterns. Implicit rigid motion assumptions~\cite{Rauch2022}, on the other hand, allow segmenting arbitrary objects, but do not provide semantics.

Vision and Text Transformers~\cite{Radford2021}, trained on large datasets, enable open-set image and text relations within a common embedding space. Since the majority of these models are trained on internet-scale datasets with predominantly 2D images, they are not directly applicable to 3D. For an application relying on a stream of 2D images, this also implies that only the current field-of-view (FoV) can be considered at any point in time, excluding mobile tasks, where objects-of-interest are outside the current FoV.

Recent work uses implicit neural representations~\cite{Mildenhall2022,Kerbl2023} to merge and render embeddings in place of colour channels~\cite{Kerr_2023_ICCV,Qin_2024_CVPR}, or rely on various offline or online explicit representations~\cite{Jatavallabhula-RSS-23,Yamazaki2024,Werby2024,Alama2025}. Often, these approaches are not real-time capable due to their training regimes requiring multiple dependent stages operating globally on the entire dataset, requiring ground truth camera poses, or using prohibitively expensive Vision-Language Models (VLMs) or scene representations, that prevents their use in most robotic downstream tasks.

\begin{figure}
    \centering
    \begin{tikzpicture}
        \node[anchor=south west,inner sep=0] at (0,0) {\includegraphics[width=0.5\linewidth,frame]{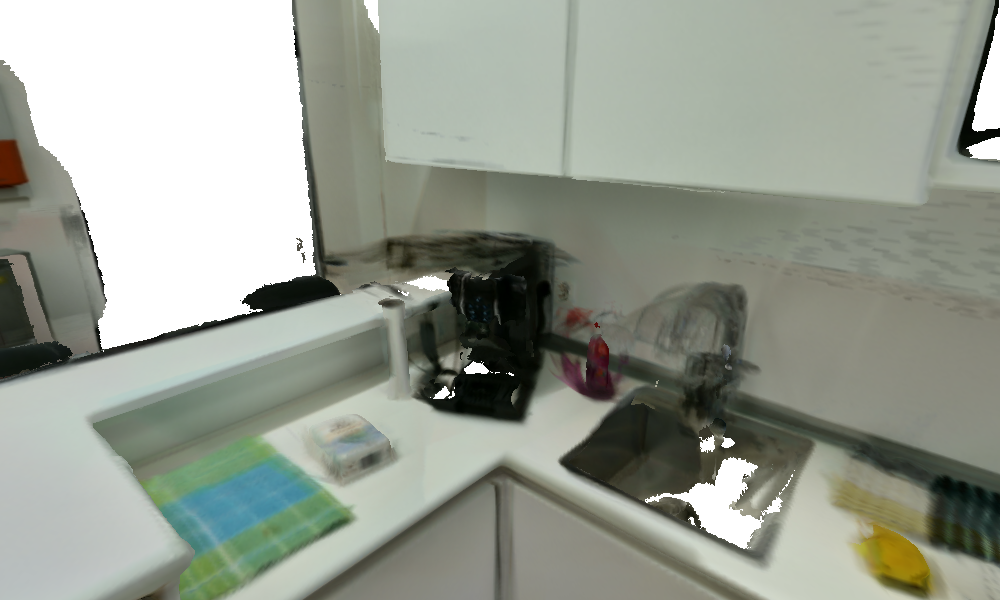}};
        \node[fill=white,below right] at (current bounding box.north west) {kitchen};
        \draw[red!80,thick,rounded corners=0.2mm] (1.75,0.75) rectangle (2.55,1.7); % coffee machine
    \end{tikzpicture}%
    \begin{tikzpicture}
        \node[anchor=south west,inner sep=0] at (0,0) {\includegraphics[width=0.5\linewidth,frame]{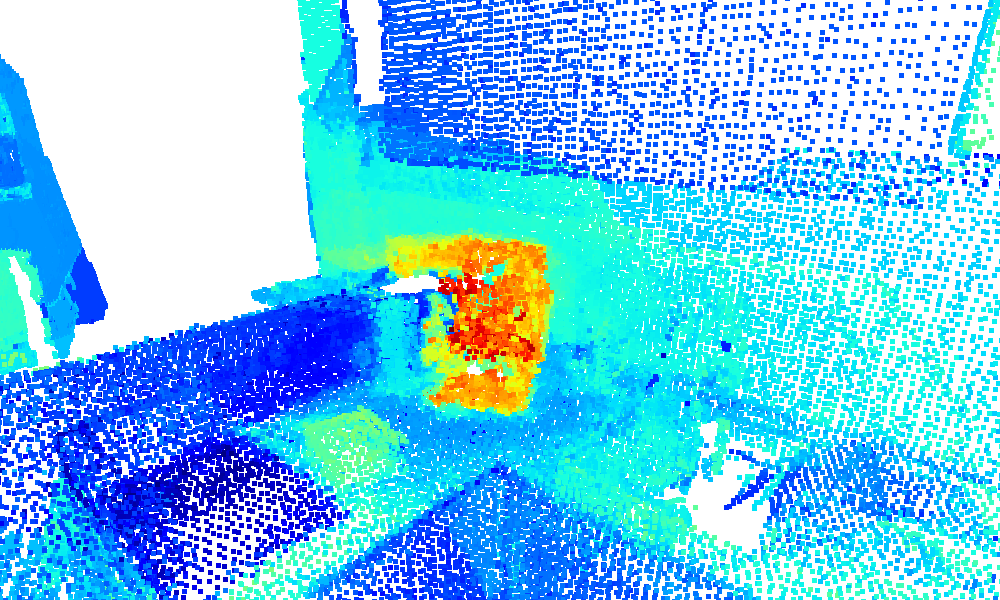}};
        \node[fill=white,below right] at (current bounding box.north west) {\textit{coffee machine}};
        \draw[black,thick,rounded corners=0.2mm] (1.75,0.75) rectangle (2.55,1.7); % coffee machine
    \end{tikzpicture}%
    
    \begin{tikzpicture}
        \node[anchor=south west,inner sep=0] at (0,0) {\includegraphics[width=0.5\linewidth,frame]{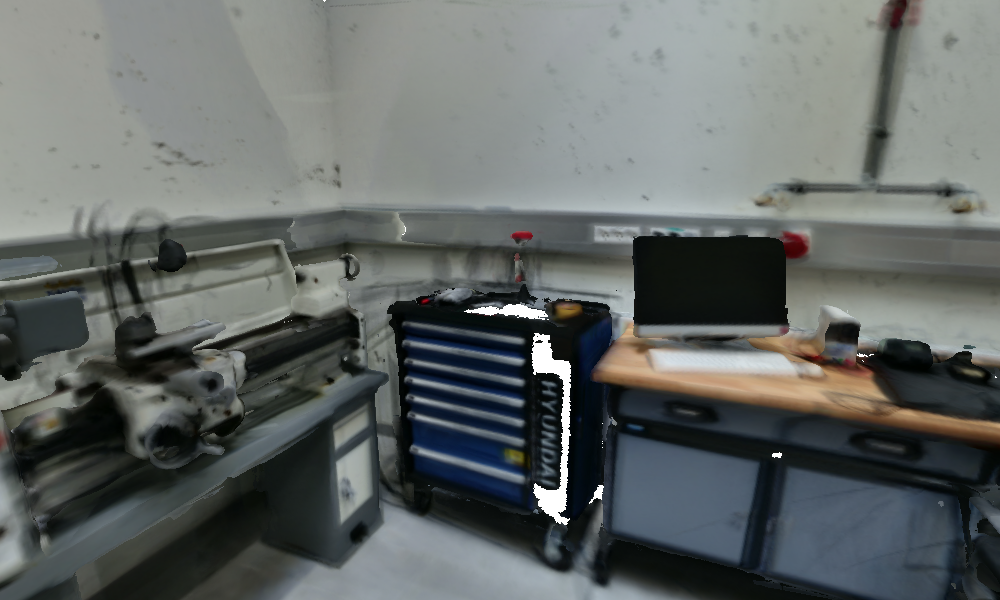}};
        \node[fill=white,below right] at (current bounding box.north west) {workshop};
        \draw[red!80,thick,rounded corners=0.2mm] (1.65,0.1) rectangle (2.8,1.5); % toolbox
    \end{tikzpicture}%
    \begin{tikzpicture}
        \node[anchor=south west,inner sep=0] at (0,0) {\includegraphics[width=0.5\linewidth,frame]{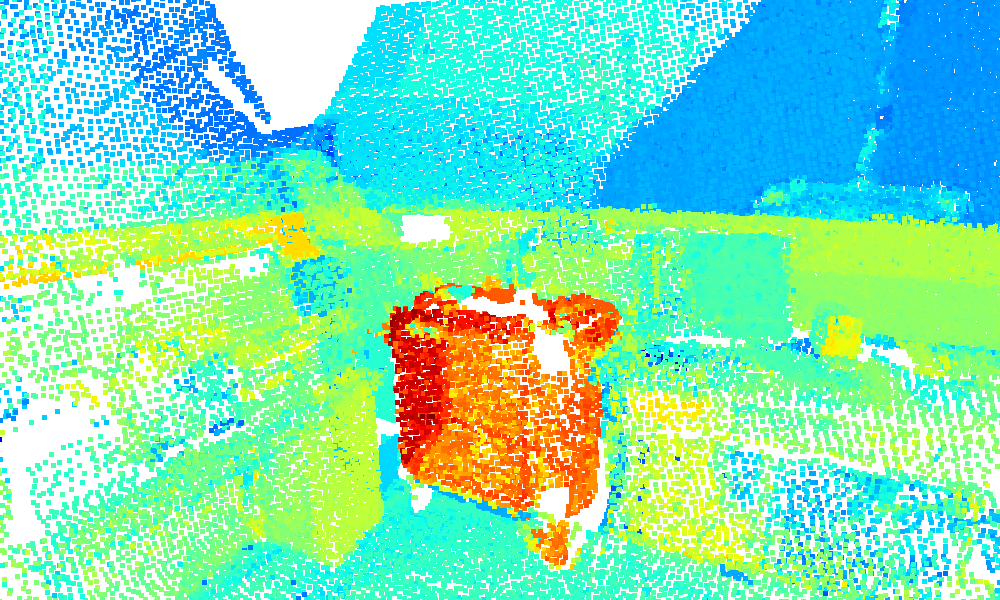}};
        \node[fill=white,below right] at (current bounding box.north west) {\textit{toolbox}};
        \draw[black,thick,rounded corners=0.2mm] (1.65,0.1) rectangle (2.8,1.5); % toolbox
    \end{tikzpicture}%
    
    \begin{tikzpicture}
        \node[anchor=south west,inner sep=0] at (0,0) {\includegraphics[width=0.5\linewidth,frame]{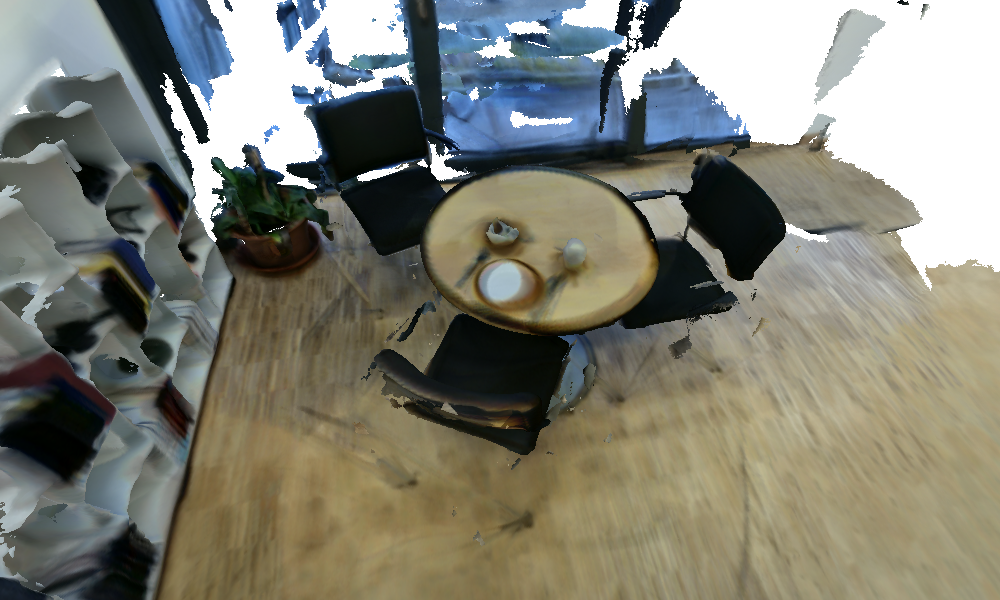}};
        \node[fill=white,below right] at (current bounding box.north west) {table};
        \draw[red!80,thick,rounded corners=0.2mm] (2.35,1.4) rectangle (2.75,1.7); % cup
    \end{tikzpicture}%
    \begin{tikzpicture}
        \node[anchor=south west,inner sep=0] at (0,0) {\includegraphics[width=0.5\linewidth,frame]{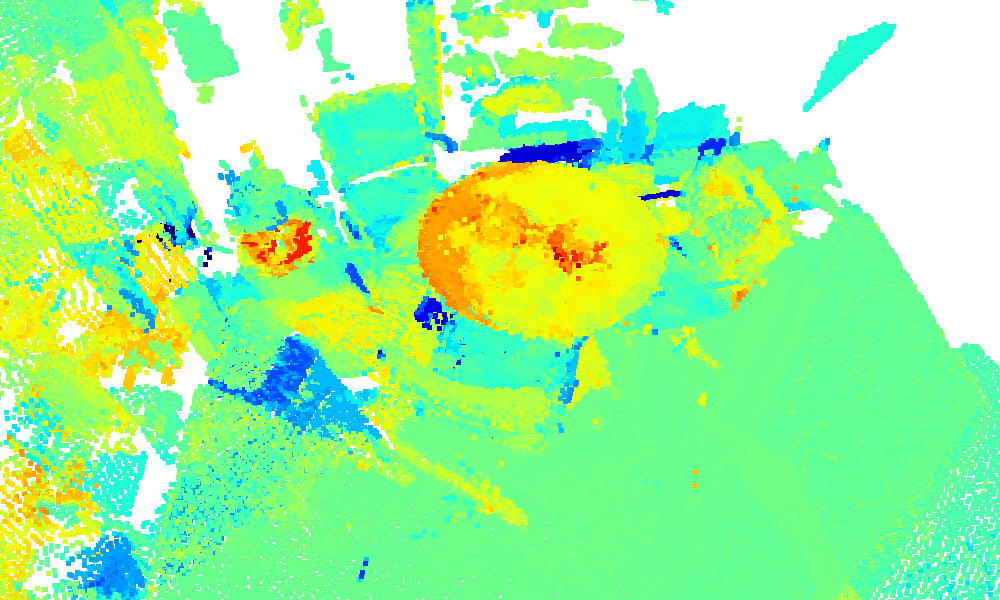}};
        \node[fill=white,below right] at (current bounding box.north west) {\textit{cup}};
        \draw[black,thick,rounded corners=0.2mm] (2.35,1.4) rectangle (2.75,1.7); % cup
    \end{tikzpicture}%
    
    \caption{3D reconstruction of various environments and their vision-language embeddings in real-time (left), and the response to text queries (right).}
    \label{fig:overview}
    \vspace{-0.5cm}
\end{figure}

This work investigates different strategies to integrate vision-language embeddings in 3D and proposes a method, VLEM (Vision-Language Embedding Mapping), for the real-time mapping of image embeddings in 3D for open-set robotic downstream tasks.
We only require a stream of colour and depth images to reconstruct an environment with image embeddings (\Cref{fig:overview}). We do not require ground truth camera poses or any pre-training on the target environment. This makes our method environment- and task-agnostic and serves as a building block for other tasks which require a language interface to robotic applications.

% In summary, this work contributes:
% \begin{itemize}
%     \item a framework for the real-time integration of VLMs in 3D, requiring only raw RGB-D images,
%     \item an analysis of various VLMs backends and integration approaches, and their effect on segmentation performance, resource consumption, and task representation,
%     \item an efficient vision-language embedding integration approach, that is more memory efficient and performant than state-of-the-art baselines, and its demonstration for interactive robotic tasks.
%     % \item demonstrations of a novel real-time embedding integration approach for interactive language instructions in handheld, manipulation and mobile robotic settings.
% \end{itemize}

% In summary, this work contributes a:
% \begin{itemize}
%     \item a framework for the real-time integration and query of various image- and patch-aligned vision-language embeddings,
%     \item a efficient approach for integrating image-aligned embeddings into ... better segm perf
%     \item a compact scene representation that 
%     \item an efficient semantic scene representation, that is memory efficient ... less memory, can use SLAM in paralle on less than 12GiB VRAM
% \end{itemize}

In summary, this work contributes 1) a framework for the real-time integration and query of various image- and patch-aligned vision-language embeddings, 2) a novel mask refinement approach for integrating image-aligned embeddings into a 3D point cloud, and 3) a compact semantic scene representation that provides superior segmentation performance in resource constrained setups.

The efficiency of our approach is demonstrated by the parallel real-time camera localisation and embedding mapping in interactive robotic tasks using less than 12 GiB VRAM.

\section{Related Work}

\subsection{Vision-Language Models}

VLMs map visual and textual information into a common embedding space and are the foundation of many open-set language-semantic tasks. Initial works, such as CLIP \cite{Radford2021} and OpenCLIP \cite{Cherti2023}, mapped entire images into the embedding space. This single embedding combines the semantic information of all objects in the image, and thus limits the granularity of localising individual objects. A common strategy to increase the granularity of the embeddings is the sampling of regions-of-interest in the image and extracting region-based embeddings. These regions can for example be extracted randomly (LERF~\cite{Kerr_2023_ICCV}), or based on a class-agnostic segmentation (LangSplat~\cite{Qin_2024_CVPR}, ConceptFusion~\cite{Jatavallabhula-RSS-23}).

Recent VLMs, such as DINOv3~\cite{Siméoni2025}, and the combination of NACLIP~\cite{Hajimiri2025} with RADIO~\cite{Heinrich2025} as proposed in RayFronts~\cite{Alama2025}, provide semantically meaningful patch-wise embeddings based on the Vision Transformer (ViT) image patches. Thus, they provide a more fine-grained localisation at the cost of segment consistency.

\subsection{Scene Representations}

Explicit representations encode a scene directly in a metric-accurate space and have been used as dense model representations for scene reconstruction and camera tracking \cite{Whelan-RSS-15,Koide2024}. In contrast, implicit representations encode a scene indirectly via parameters of a learned model, such as in the form of a neural network~\cite{Mildenhall2022} or as an ensemble of Gaussians~\cite{Kerbl2023}, focus on high-fidelity visual reconstruction instead of metric accurate representations, and rely on ground truth camera poses.

% While explicit representations focus on accurate metric representations and the application to real-time camera localisation, implicit representations, originating from the graphics community, focus on visual fidelity and the aim for accurate view synthesis.
% Most baseline implicit representation implementations rely on ground truth camera poses and do not consider depth or range images for a metric-accurate representation, which makes them less applicable for robotic applications requiring an accurate Cartesian task representation.

\subsection{Embedding Integration}

% The majority of VLMs are trained on, and thus, are only applicable to the 2D image domain. This limited field-of-view excludes their use in many common robotic applications requiring a holistic representation of the environment. A common approach in literature is the integration of multi-view embeddings into a common 3D representation.

VLMs operating in the 2D image domain have a limited field-of-view, which makes them impractical for many robotic applications that require a holistic representation of the environment. Building up such a holistic representation requires integrating embeddings from multiple views.

Implicit representation approaches \cite{Kerr_2023_ICCV,Qin_2024_CVPR} use their continuous scene representation to synthesise high-fidelity 2D views of vision-language embeddings in place of colour. This has the advantage that 2D views can be synthesised smoothly at arbitrary resolutions, but requires synthesising multiple views for a global localisation of objects in a scene.

% In parallel, approaches using explicit representations emerged, using point clouds \cite{Jatavallabhula-RSS-23,Gu2024}, voxel grid hash-maps \cite{Yamazaki2024}, or static voxel maps \cite{Alama2025} as 3D representation. While some earlier approaches focus on offline processing of already registered scene views \cite{Takmaz2023,Werby2024} for robotic tasks, such as navigating via a scene graph, recent methods focused on online processing of colour and depth data \cite{Yamazaki2024,Alama2025}.

Explicit approaches, using point clouds \cite{Jatavallabhula-RSS-23,Gu2024}, voxel grid hash-maps \cite{Yamazaki2024}, or static voxel maps \cite{Alama2025} as 3D representation, integrate embeddings directly into a Cartesian task space. Earlier work focused on offline processing of already registered scene views \cite{Takmaz2023,Werby2024} for robotic tasks, such as navigating via a scene graph, while recent methods focused on online processing of colour and depth data \cite{Yamazaki2024,Alama2025}.

\subsection{Summary}

We are similarly motivated to construct a metric-accurate semantic scene representation by processing RGB-D data online. In contrast to Open-Fusion~\cite{Yamazaki2024} and RayFronts~\cite{Alama2025}, which require ground truth camera poses and maintain embeddings on coarse segments or voxels, respectively, we optionally estimate the camera pose and integrate embeddings per 3D point in an adaptively resampled point cloud to trade-off size, and semantic and spatial granularity, in order to enable real-time robotic manipulation tasks.

\section{Method}

\subsection{Problem Formulation}
\label{sec:problem_formulation}

For a continuous stream of colour and depth image pairs $(\mathbf{C}, \mathbf{D})_i : \{ \mathbf{C} : \mathbb{R}^{w \times h} \mapsto \mathbb{R}^3, \mathbf{D} : \mathbb{R}^{w \times h} \mapsto \mathbb{R} \}$ over a time index $i \in \mathbb{N}_{\geq 0}$, we are looking for a metric-accurate 3D representation that encodes geometric information in form of coordinate points $\mathbf{p} \in \mathbb{R}^3$ and semantic information in form of an arbitrary embedding vector $\mathbf{e} \in \mathbb{R}^d$:

\begin{equation}
    \mathcal{M}_i : \{ (\mathbf{p}, \mathbf{e})_j \, | \, j \in \left[ 0, N_{\mathcal{M}} \right] \} \, .
    \label{eq:map}
\end{equation}

We further impose the following constraints: Ground truth camera poses are not available. The number of RGB-D image pairs is unlimited and $N_{\mathcal{M}} := \left| \mathcal{M}_i \right|$ is the upper bound of points. $\mathcal{M}_i$ must be sufficiently dense to enable common robotic manipulation tasks. At any time index $i$, $\mathcal{M}_i$ can only contain data up to this point and data is integrated in chronological order. $\mathcal{M}_i$ can be accessed at any point in time.

\begin{figure*}
\centering
\includegraphics[width=\linewidth]{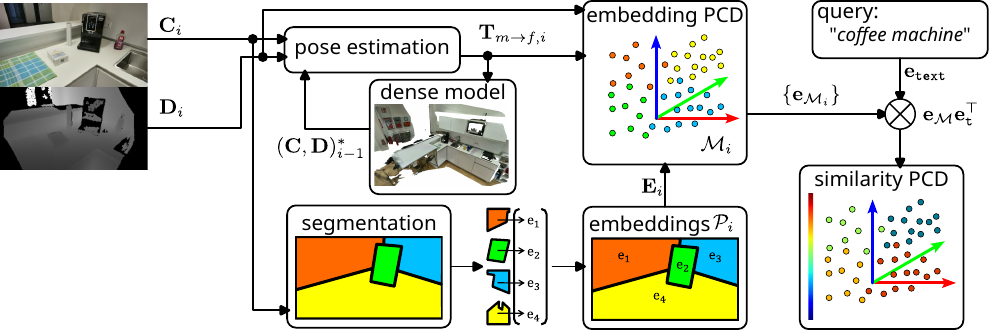}
\caption{Segment-embedding integration pipeline. For a pair of colour and depth images, we estimate the camera pose and reconstruct a dense colour model. In parallel, segment-wise vision-language embeddings are extracted from the colour images, and eventually integrated in a 3D point cloud. At any point in time, when a text query is received, the cosine similarity between the query and all points in the embedding point cloud is computed and used for various downstream tasks.}
\label{fig:pipeline_overview}
\vspace{-0.5cm}
\end{figure*}

In summary, the 3D representation is maintained in real-time for an unlimited time using only the raw sensor data and can be accessed or queried in real-time as well. This requires a constant integration and compression of image data and local embeddings and camera localisation (\Cref{fig:pipeline_overview}).
Our approach is modular, and the camera pose can optionally be provided by other means. In any case, the image and embedding integration is done in a chronological order in real-time.

% First, we estimate the camera pose via frame-to-model tracking (\Cref{sec:localisation}). Second, in parallel, mask-level embeddings are extracted per camera view (\Cref{sec:vl_embeddings_2d}). Third, the mask-level embeddings are integrated over multiple views using the estimated camera pose (\Cref{sec:embedding_integration_3d}).

\subsection{Localisation}
\label{sec:localisation}

The first step in our pipeline is the estimation of the camera pose. This part relies on a baseline reference implementation and is therefore only described briefly.

Given two consecutive image pairs $(\mathbf{C}, \mathbf{D})_a$ and $(\mathbf{C}, \mathbf{D})_b$, and the camera intrinsics $\mathbf{K}$, the visual odometry estimates the rigid transformation $\mathbf{T}_{a \rightarrow b} \in \mathrm{SE}(3)$ between the camera frames at these two points in time, using a coarse-to-fine multiscale approach on an image pyramid with a hybrid photometric and geometric loss~\cite{Steinbrücker2011}.

% \cite{Steinbrücker2011,Park2017}

We follow Open-Fusion~\cite{Yamazaki2024} and use a multivalued hash map, with blocks and voxels as key and the signed distance, integration weight, and colour as value, as a dense model. In contrast to \cite{Yamazaki2024}, which relies on ground truth camera poses, we use this map to reduce the drift in the frame-to-model camera pose estimation. After initialising this hash map with $(\mathbf{C}, \mathbf{D})_0$, consecutive estimations of $\mathbf{T}$ will use synthesised colour and depth renderings $(\mathbf{C}, \mathbf{D})_a^*$ at the previous point in time $a$ in place of real sensor data.

% Integrating the colour and depth into the hash map can be done at a much higher frame rate than integrating the post-processed embeddings due to the lower dimensionality of colour and the additional overhead of the mask-level embedding extractions. Hence, this parallelisation and decoupling allows us to better trade off the tracking accuracy with the density and dynamic range of $\mathcal{M}_i$. Hereby, the pose estimation is of greater concern due to its permanent effect on map and embedding degradation.

\subsection{Pixel-Aligned Vision-Language Embeddings}
\label{sec:vl_embeddings_2d}

Pixel-level embeddings are extracted from the colour image, assigning one embedding vector to every colour pixel coordinate: $\mathbf{C}_i \in \mathbb{R}^{w \times h \times 3} \mapsto \mathbf{E}_i \in \mathbb{R}^{w \times h \times d}$.
Following previous work, we investigate image-level embeddings~\cite{Kerr_2023_ICCV,Jatavallabhula-RSS-23,Qin_2024_CVPR,Yamazaki2024} as well as ViT-patch-level embeddings~\cite{Alama2025}.

\subsubsection{Segment-Embeddings}
\label{sec:embed_segm}

ViTs where only the class token carries semantic meaning, such as used by CLIP~\cite{Radford2021} for image and text alignment, provide semantic meaningful embeddings only on a global image level. This is not sufficiently discriminative for most downstream applications.
We broadly follow \cite{Jatavallabhula-RSS-23,Kerr_2023_ICCV,Qin_2024_CVPR} and extract vision-language embeddings for regions of the image.

\paragraph{Regions-of-Interest}

A region extraction method provides a potentially empty set of bounding boxes $\mathcal{B}_i : \{\mathcal{R}_k : (u, v, r_w, r_h) \}$, from which a set of cropped image segments $\mathbf{C}_k : \mathbb{R}^{r_w \times r_h} \rightarrow \mathbb{R}^3$, located at pixel coordinates $(u, v)$ in the original image $\mathbf{C}_i$, can be extracted and passed to the vision encoder, $V : \mathbf{C}_k \mapsto \mathbf{e}_k \in \mathbb{R}^d$, to obtain the embedding vector $\mathbf{e}_k$ for that segment.

While LERF~\cite{Kerr_2023_ICCV} samples cropped regions along a randomly sampled ray, we follow LangSplat~\cite{Qin_2024_CVPR} and ConceptFusion~\cite{Jatavallabhula-RSS-23}, and extract semantically more meaningful regions using Segment Anything~(SAM)~\cite{Kirillov_2023_ICCV}. Using the additional pixel-wise binary foreground/background mask $\mathbf{M}_k : \mathbb{R}^{r_w \times r_h} \rightarrow \mathbb{B}$, $\mathbf{e}_k$ can be associated directly to pixel coordinates in $\mathbf{C}_i$. On top of this baseline, we apply a mask refinement and visual encoder masking strategy.

\paragraph{Refinement}

For areas of $\mathbf{C}_i$, that have not been segmented, we uniformly sample a fixed amount of coordinates within these non-segmented areas and provide these as query points yet again to SAM~\cite{Kirillov_2023_ICCV}. The resulting fill-in segments are added to $\mathcal{B}_i$. While this process could be repeated, we found one additional iteration a sufficient trade-off and more efficient than using a larger model.

\paragraph{Masking}

% The naive approach of directly passing the cropped images $\mathbf{C}_k$ to the vision encoder and assigning the resulting embedding vector via the mask to the final image~\cite{Jatavallabhula-RSS-23} has the drawback that overlapping regions are potentially assigned similar embedding vectors, and thus respond similar to text queries (\Cref{fig:segment_masking}).

\begin{figure}
    \centering
    \subfloat[colour]{\includegraphics[width=0.33\linewidth]{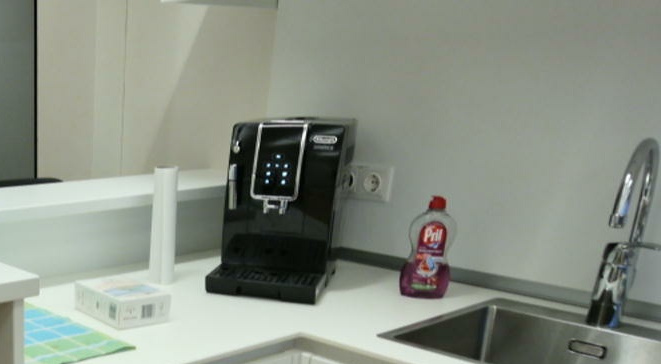}\label{fig:segment_masking_colour}}
    \subfloat[ConceptFusion]{\includegraphics[width=0.33\linewidth]{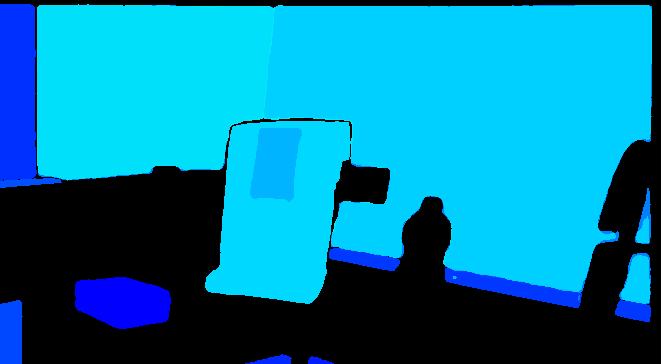}\label{fig:segment_masking_cf}}
    \subfloat[VLEM]{\includegraphics[width=0.33\linewidth]{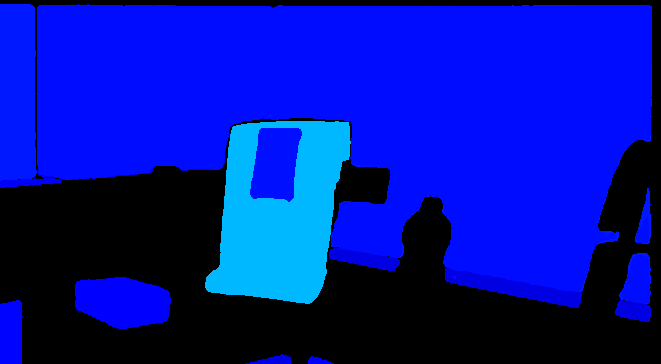}\label{fig:segment_masking_xf}}
    \caption{
    Similarity to query \textit{coffee machine} for different masking strategies (jet colourmap, dark blue: low similarity, light blue: high similarity).
    \textit{ConceptFusion:} Bounding boxes around the white background segments (\protect\subref{fig:segment_masking_colour}) contain parts of the coffee machine. Embeddings for these background segments are very close to the \textit{coffee machine} query (\protect\subref{fig:segment_masking_cf}, background in light blue / high similarity).
    \textit{VLEM:} Masking out unrelated segments reduces the query distance (\protect\subref{fig:segment_masking_xf}, white background in dark blue / low similarity) and provides a better separation to the actual coffee machine.}
    \label{fig:segment_masking}
    \vspace{-0.5cm}
\end{figure}

A common approach for extracting segment embeddings in related work is passing the cropped images $\mathbf{C}_k$ directly to the vision encoder \cite{Kerr_2023_ICCV,Qin_2024_CVPR,Jatavallabhula-RSS-23}. This naive approach has the drawback that overlapping cropped regions contain image data from neighbouring segments, thus diluting the semantic specificity of a segment embedding and its query response (\Cref{fig:segment_masking}).

Masking the cropped segment $\mathbf{C}_k$ and replacing non-segmented background areas
$$
\mathbf{C}_k'(u, v) :=
\begin{cases} 
\mathbf{v}, & \text{if} \; \mathbf{M}_k(u, v) = 0, \\
\mathbf{C}_k(u, v), & \text{if} \; \mathbf{M}_k(u, v) = 1,
\end{cases}
$$
avoids this effect by extracting visual embeddings that are more closely related to the segmented image data.

%Ideally, $\mathbf{v}$ is a value that appears the least frequent in the original data. Since this is intractable, we set $\mathbf{v} := (1, 0, 1)$ in practice.

We found that merging global image embeddings with the local segment embeddings has a similar effect on the similarity of queries and, therefore, refrain from this merging strategy and only use local segment embeddings.

All extracted regions are bicubically resampled to the same resolution and encoded as a single batch for efficiency.

\subsubsection{ViT-Patch-Embeddings}
\label{sec:embed_patch}

DINOv3~\cite{Siméoni2025}, and the combination of NACLIP~\cite{Hajimiri2025} with RADIO~\cite{Heinrich2025}, provide semantically meaningful embeddings $\mathbf{e}_k$ for each ViT patch.
Essentially, the set of regions $\mathcal{B}_i : \{\mathcal{R}_k : (u, v, r_w, r_h) \}$ is a regular tessellation of the image into square patches. The size of these squares $(r_w, r_h)$ is an inherent property of the ViT.
Instead of propagating the embeddings to the full resolution via the region mask, we bilinearly upsample and normalise the patch embeddings \cite{Alama2025}. While this produces blurred results with less sharp semantic segmentation boundaries, these approaches are faster as they do not rely on a dedicated pre-segmentation stage.

% and thus results in low performance in 2D, their integration in 3D results in equally good results without the requirement for pre-segmentation phase.

\subsection{3D Embedding Integration}
\label{sec:embedding_integration_3d}

% The local segment embeddings $\mathbf{E}_i : \mathbb{R}^{w \times h} \mapsto \mathbb{R}^d$ are only available for the current FoV at time index $i$ and many robotic applications require localisation of objects outside the current FoV. The embedding lifting integrates the embedding vectors of the local segments into the 3D map of point coordinates and embeddings (\cref{eq:map}).

To provide a globally consistent semantic representation of the vision-language embeddings, we integrate the individual local embeddings, $\mathbf{E}_i : \mathbb{R}^{w \times h} \mapsto \mathbb{R}^d$, into a 3D map of point coordinates and embeddings (\cref{eq:map}).

For each depth image $\mathbf{D}_i$ and the camera matrix $\mathbf{K}$, we back-project the pixel coordinates $(u, v)$ to point coordinates in the camera frame
$\mathbf{p} := \mathbf{K}^{-1} \cdot \mathbf{D}_i(u, v) \cdot \left[ u, v, 1 \right]^{\top}$
and use the same relation to associate the embeddings in $\mathbf{E}_i$ to the same points, such that all pixels with valid depth readings in the image space produce a set of tuples of point coordinates and embedding vectors in the current camera frame:
\begin{equation}
    \mathcal{P}_i : \{ (\mathbf{p}, \mathbf{e})_j \, | \, j \in \left[ 0, N_{\mathcal{P}} \right], N_{\mathcal{P}} \leq w \cdot h \} \, .
    \label{eq:map_cam}
\end{equation}
% We compute the confidence $c \in [0, 1]$ of an embedding vector based on how close the normal $\hat{\mathbf{n}}$ of its point $\mathbf{p}$ is to the line of sight:
% \begin{equation}
% c(\mathbf{p}) = (\hat{\mathbf{n}} \cdot \hat{\mathbf{p}} + 1) / 2
% \end{equation}
% to relate to how well a point was visible when its segment-level embedding vector was extracted.
The points are then transformed from the camera frame to the map frame using the current camera pose. If $i=0$, we initialise $\mathcal{M}_0 := \mathcal{P}_0$, otherwise, $\mathcal{M}_i$ is iteratively downsampled to a uniform point density (points per $\mathrm{m}^3$) such that $\lvert \mathcal{M}_i\rvert \leq N_{\mathcal{M}}$.
This limits the memory consumption while avoiding further sparsifying regions with low density.

Given the current points and their embeddings $\mathcal{P}_i$, and the integrated map up to the previous view, $\mathcal{M}_{i-1}$, we integrate new points and embeddings, by first establishing a set $\mathcal{C}$ of dense point correspondences $\{(\mathbf{q}, \mathbf{p})\}$ in Cartesian space by a nearest neighbour search:
\begin{equation}
\mathcal{C}_i : \{(\mathbf{q}, \mathbf{p}^*) \, | \,  \mathbf{p}^* = \underset{\mathbf{p} \in \mathcal{P}_i}{\arg\min} \lVert \mathbf{p} - \mathbf{q} \rVert_2, \, \mathbf{q} \in \mathcal{M}_{i-1}\} \, .
\end{equation}

For each point coordinate in the current map $\mathcal{M}_{i-1}$, we find its closest point coordinate in the current input $\mathcal{P}_i$ within a certain maximum search distance. Note that this is not a symmetric 1:1 relation and one point in $\mathcal{P}_i$ can be closest to many points in $\mathcal{M}_{i-1}$ and there may be points in $\mathcal{M}_{i-1}$ without a corresponding match.

% We integrate the embedding vector $\mathbf{e}_j$ and confidence $c_j$ for every $\mathbf{p}^* \in \mathcal{C}_i$ into its associated $\mathbf{q} \in \mathcal{M}$
% \begin{align}
% c_{\mathcal{M}} & := [ (1 - w) \cdot c_{\mathcal{P}}] + [w \cdot c_{\mathcal{M}}] \\
% \mathbf{e}_{\mathcal{M}} & := [ (1 - w) \cdot \mathbf{e}_{\mathcal{P}}] + [w \cdot \mathbf{e}_{\mathcal{M}}]
% \end{align}
% with confidence-based weight
% \begin{align}
% w = 1 - \left[ \left( 1 - \frac{c(\mathbf{p})}{c(\mathbf{p})+c(\mathbf{q})} \right) \cdot \left(1 - w_{\mathrm{new}} \right) \right] \,
% \end{align}
% after renormalisation. $w_{\mathrm{new}}$ is a fixed weight defining how strongly new data is to be integrated into the map. $\mathbf{p}$ in $\mathcal{P}$ without a correspondence in $\mathcal{M}$ are integrated with their embedding vector and confidence without further weighting.

The embedding vector $\mathbf{e}_j$ for every $\mathbf{p}^* \in \mathcal{P}_i$ is integrated into its associated $\mathbf{q} \in \mathcal{M}$ by the weighted average
$\mathbf{e}_{\mathcal{M}} := [ (1 - w) \cdot \mathbf{e}_{\mathcal{P}}] + [w \cdot \mathbf{e}_{\mathcal{M}}]$
and renormalised. $w := 0.5$ is a fixed weight defining to which extent the model embeddings should be updated by new observed embeddings.
$(\mathbf{p}, \mathbf{e})$ in $\mathcal{P}$ without a correspondence are integrated directly.

\subsection{3D Query Segmentation}
\label{sec:segmentation}

The text encoder provides one embedding vector $\mathbf{e}_{\mathrm{text}}$ for a text query containing an arbitrary number of words. The normalised cosine similarity,
$s_{\mathrm{query}} := (\mathbf{e}_{\mathrm{text}} \cdot \mathbf{e}_{\mathcal{M}}^{\top} + 1) / 2$, $s \in [0, 1]$
between this text embedding and the visual embedding vectors associated with points in $\mathcal{M}$ provides a 3D heatmap (\Cref{fig:overview}).
At any point in time $i$, we can query for arbitrary many $\mathbf{e}_{\mathrm{text}}$, yielding a list of similarities $[s_{1}, s_{2}, \dots, s_{N}]$ for each point $\mathbf{p} \in \mathcal{M}_i$.
For assigning labels independently to individual points, we rely on binary thresholding of the similarity score.

% At any point in time $i$, we can query for arbitrary many $\mathbf{e}_{\mathrm{text}}$, yielding a list of similarities $s_{1}, s_{2}, \dots, s_{N}, s_{\mathrm{\textit{object}}}$ for each point $\mathbf{p} \in \mathcal{M}_i$. Selecting the maximum similarity from this list separates our model $\mathcal{M}$ into disjoint point sets $\mathcal{Q}_{1}, \dots, \mathcal{Q}_{N}$, with $\mathcal{Q}_k : \{ (\mathbf{p}, s)_j \}$, for the task-specific text queries, and $\mathcal{Q}_{\mathrm{\textit{object}}}$, which is neglected and not used further.

For robotic tasks that rely on a closed set of 3D points, such as grasping an object, we apply clustering and use the cluster with the highest similarity.
To cluster a 3D point set by similarity, we additionally query for a generic \textit{object}, such that each $\mathbf{p} \in \mathcal{M}$ is associated with similarities for an actual task-specific query $s_{\mathrm{query}}$ and a generic baseline similarity $s_{\mathrm{\textit{object}}}$. The additional query for \textit{object} is inspired by LangSplat~\cite{Qin_2024_CVPR} and makes sure that our target is more relevant than a generic \textit{object}.
Selecting the maximum query similarity from $[s_{1}, s_{2}, \dots, s_{N}, s_{\mathrm{\textit{object}}}]$ partitions our model $\mathcal{M}$ into disjoint point sets $\mathcal{Q}_{1}, \dots, \mathcal{Q}_{N}$, with $\mathcal{Q}_k : \{ (\mathbf{p}, s)_j \}$, for the task-specific text queries, and $\mathcal{Q}_{\mathrm{\textit{object}}}$, which is neglected and not used further.

We further remove unstable points by clustering the similarity and coordinates to remove outliers and only retain points with a high similarity. For similarity clustering, we are fitting a Gaussian Mixture Model with three components, initialised between the minimum and maximum similarity, and select the cluster with the highest mean value. Coordinate clustering is done via DBSCAN (Density-Based Spatial Clustering of Applications with Noise).

\section{Evaluation}

\subsection{Setup}

% We evaluate the open-set segmentation performance and the application to robotic tasks on several custom RGB-D sequences and ScanNet. The custom sequences and robotic applications use an Orbbec Femto Bolt Time-of-Flight camera in handheld mode, wrist-mounted to a robot arm (Franka Emika Robot), and attached to a mobile base (Segway RMP Lite 220). Colour and registered depth images are gathered at a resolution of $1280 \times 720$ at $30$ Hz, whereas ScanNet evaluations use the native resolution of $640 \times 480$.

% We evaluate the open-set segmentation performance and the application to robotic tasks on several real, non-synthetic RGB-D, sequences from an Orbbec Femto Bolt Time-of-Flight camera and the ScanNet~\cite{Dai2017} dataset.

\subsubsection*{Data}
We evaluate the open-set segmentation performance on several RGB-D sensor sequences from an Orbbec Femto Bolt Time-of-Flight camera ($1280 \times 720$ at $30$ Hz) and the ScanNet~\cite{Dai2017} dataset ($640 \times 480$ at $30$ Hz).

\subsubsection*{Protocol}
For evaluations on ScanNet, we follow RayFronts and evaluate the mIoU on all sequences of the ScanNet scenes 11, 50, 231, 378, and 518 using the ground truth camera trajectory and labels. ScanNet evaluations run in real-time on an NVIDIA GeForce RTX 4090 with 24GiB VRAM and process frames as they arrive, potentially skipping frames.

In robotic experiments, the RGB-D camera is wrist-mounted to a robot arm (Franka Emika Robot) or attached to a mobile base (Segway RMP Lite 220) and its pose is estimated online in parallel to the embedding mapping. Robot experiments run in real-time on an NVIDIA GeForce RTX 4070 with 12GiB VRAM.

We compute the average mIoU over the full range of similarity thresholds and associate the 3D points of the ground truth model to the estimated embedding map by nearest neighbour search within 5cm.

% and ScanNet. The custom sequences and robotic applications use an Orbbec Femto Bolt Time-of-Flight camera in handheld mode, wrist-mounted to a robot arm (Franka Emika Robot), and attached to a mobile base (Segway RMP Lite 220). Colour and registered depth images are gathered at a resolution of $1280 \times 720$ at $30$ Hz, whereas ScanNet evaluations use the native resolution of $640 \times 480$.

\subsubsection*{Baselines}

We evaluate VLEM with various VLM backends (CLIP ViT-B/16, DINOv3 dino.txt ViT-L/16, NACLIP + RADIOv2.5-B (NARADIO)) against ConceptFusion~(CF)~\cite{Jatavallabhula-RSS-23} and Open-Fusion~(OF)~\cite{Yamazaki2024} with CLIP ViT-H/14, and RayFronts~(RF)~\cite{Alama2025} with NACLIP + RADIOv2.5-B (NARADIO). We also compare to LangSplat~\cite{Qin_2024_CVPR} with CLIP ViT-B/16 as a non-realtime upper performance bound.

\subsection{VLM Integration Ablation Study}

% | model                               | WS |  mIoU  |
% |-------------------------------------|----|--------|
% | m_cf                                | 11 | 0.924% |
% | m_dino                              | 11 | 0.915% |
% | m_naradio                           | ME | 0.904% |

% | xf_s_fillin_refine_m_fixed          | 13 | 0.943% |
% | xf_s_fillin_refine_m_mean_image     | 13 | 0.943% |
% | xf_s_fillin_refine_m_mean_segment   | 13 | 0.941% |
% | xf_s_fillin_refine_m_none           | 12 | 0.957% |
% | xf_s_fillin_refine_m_random         | 11 | 0.931% |

% | xf_s_original_m_fixed               | 12 | 0.908% |
% | xf_s_original_m_mean_image          | 12 | 0.907% |
% | xf_s_original_m_mean_segment        | 13 | 0.906% |
% | xf_s_original_m_none                | 12 | 0.914% |
% | xf_s_original_m_random              | 11 | 0.900% |
% |-------------------------------------|----|--------|

First, we investigate different approaches for extracting 2D pixel-aligned vision-language embeddings, and their 3D integration via VLEM. The segmentation masks are extracted with and without refinement and the background is replaced by \textit{random} data, with a \textit{static} colour of $\mathbf{v} := (1, 0, 1)$, with the \textit{mean image} colour $\mathbf{v} := \text{mean}(\mathbf{C}_i)$, the \textit{mean segment foreground} $\mathbf{v} := \text{mean}(\mathbf{C}_k \, | \, \mathbf{M}_k(u, v) = 0)$, or we do not replace the background (\textit{none}).

The 2D evaluation is carried out non-real-time on every 30th image of the ScanNet sequences and averages the mIoU for all NYU40 labels, using the label as query. \Cref{tab:scannet_2d_segm,tab:scannet_2d_vlm} show that the individual 2D segmentation on ScanNet images is generally very challenging, while CLIP embeddings of refined segments in VLEM outperform alternative patch-based embeddings and baselines.

\begin{table}
% \newcolumntype{C}[1]{>{\centering\arraybackslash}p{#1}}
\centering
\caption{2D VLEM-CLIP segmentation and mask embedding strategies: Iterative mask refinement improves performance while background replacement is not beneficial.}
\begin{tabular}{lccccc}
\toprule
\multirow{2}{0.5cm}{mask ref.}  & \multicolumn{5}{c}{segment background} \\
                & \textit{none}     & \textit{random} & \textit{static} & \textit{mean image}   & \textit{mean segm.} \\
\midrule
\textit{none}   & 0.914\%           & 0.900\%         & 0.908\%         & 0.907\%               & 0.906\% \\
\textit{with}   & \textbf{0.957\%}  & 0.931\%         & 0.943\%         & 0.943\%               & 0.941\% \\
\bottomrule
\end{tabular}
\label{tab:scannet_2d_segm}
\end{table}

\begin{table}
\centering
\caption{VLM comparison on 2D ScanNet: The proposed mask refinement outperforms patch-based embeddings.}
\begin{tabular}{lcccc}
\toprule
\textbf{VLM} & \textit{CLIP} & \textit{DINO}  & \textit{NARADIO} \\
\midrule
 % & 0.924\% & \textbf{0.957\%} & 0.915\% & 0.904\% \\
 \textbf{ConceptFusion} & 0.924\%           & ---       & --- \\
 \textbf{VLEM}          & \textbf{0.957\%}  & 0.915\%   & 0.904\% \\
\bottomrule
\end{tabular}
\label{tab:scannet_2d_vlm}
\vspace{-0.5cm}
\end{table}

% \subsection{Ablation: ScanNet 3D}

% NYU40					
% 	                 none	fixed	random	mean image	mean segment
% original	         0.259	0.259	0.261	0.256	    0.258
% fillin + refine	 0.257	0.258	0.261	0.257	    0.256

The real-time integration of embeddings in 3D significantly improves the segmentation performance (\Cref{tab:scannet_3d_segm}).
The integration of multiple views has two effects:
First, segments, and their embeddings, that are detected in one image are propagated into a global model. This improves the robustness, as we do not have to detect each segment in every individual image, and alleviates detection failures in individual images.
Second, the multi-view integration of individual segments behaves similar to the segment refinement on individual images, and thus is less beneficial the more image views and their segments are integrated into the global model.

\begin{table}
\newcolumntype{C}[1]{>{\centering\arraybackslash}p{#1}}
\centering
% \caption{3D ablation mIoU NYU40}
\caption{3D VLEM-CLIP segmentation and mask embedding strategies: Masking segments with random background provides the best performance, while there is no benefit of mask refinement when integrating in 3D.}
\begin{tabular}{lcccC{1cm}C{1cm}}
\toprule
mask ref.       & \multicolumn{5}{c}{segment background} \\
                & \textit{none}     & \textit{random} & \textit{static} & \textit{mean image}   & \textit{mean segm.} \\
\midrule
\textit{none}   & 25.9\% & \textbf{26.1\%} & 25.9\% & 25.6\% & 25.8\% \\
\textit{with}   & 25.7\% & \textbf{26.1\%} & 25.8\% & 25.7\% & 25.6\% \\
\bottomrule
\end{tabular}
\label{tab:scannet_3d_segm}
\vspace{-0.5cm}
\end{table}

\subsection{Open-Set 3D Segmentation}

% We select the best ablated VLEM configuration and compare its ScanNet segmentation performance and peak VRAM consumption against the baselines 

Second, we evaluate the real-time 3D segmentation performance of VLEM with different VLM backends (CLIP, DINO, NARADIO) against the baselines ConceptFusion~(CF), Open-Fusion~(OF), and RayFronts~(RF). VLEM-CLIP uses the best ablated configuration with mask refinement and random segment background.

% We exclude LangSplat~\cite{Qin_2024_CVPR} as it is not real-time capable. Following RayFronts, we evaluate on all RGB-D sequences of the ScanNet scenes 11, 50, 231, 378, and 518. Since the baseline methods rely on ground truth camera poses, we skip the pose estimation in VLEM and also use the ScanNet ground truth instead.

% We test our method VLEM with embeddings from CLIP (ViT-B/16), DINOv3 (ViT-L/16), and the NACLIP + RADIOv2.5(-B) implementation (NARADIO) from RayFronts. In order to run ConceptFusion in real-time, we combine its embedding extraction with our online integration.

% From the segmentation and memory comparison in \Cref{tab:scannet_miou_mem} and \Cref{fig:vram_vs_miou}, we can see that 1) the segmentation performance mainly depends on the choice of VLM, and 2) the memory consumption mainly depends on the choice of 3D representation and integration method.

\begin{figure}
    \centering
    \includegraphics[width=\linewidth]{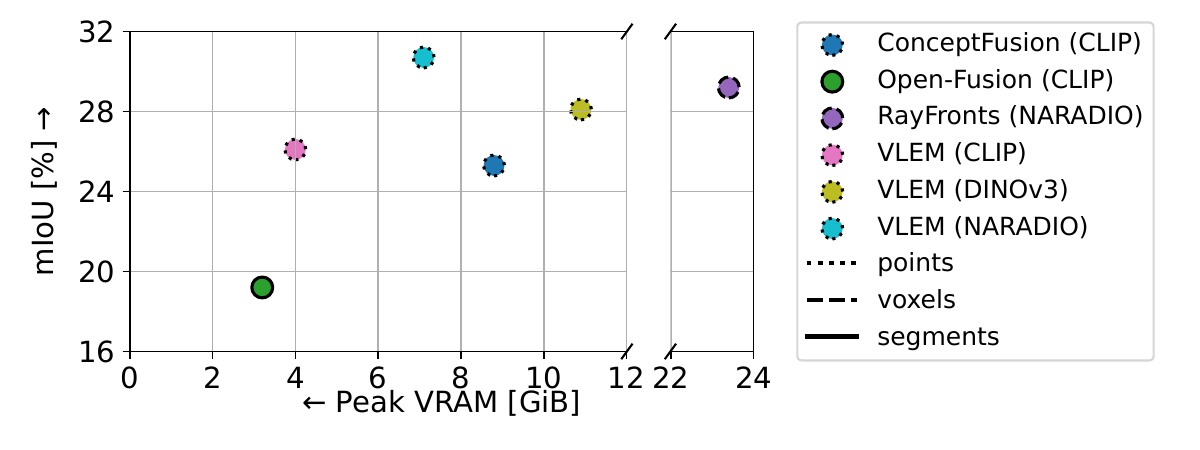}
    \caption{Embedding spatial resolution and performance on ScanNet compared to memory consumption.}
    \label{fig:vram_vs_miou}
    \vspace{-0.5cm}
\end{figure}

The comparison in \Cref{tab:scannet_miou_mem} and \Cref{fig:vram_vs_miou} shows that the segmentation performance mainly depends on the choice of VLM and embedding granularity, and the size of the semantic map is mainly determined by the 3D representation.

% RayFronts and VLEM with NARADIO (RF-NR vs. VLEM-NR) perform best in terms of segmentation, while our uniformly sampled point cloud consumes less memory. RayFronts operates on the VRAM limits of the GPU and the adjustment to 5cm voxel size makes this approach unsuitable for manipulation tasks. We can further see that 3D point embeddings from CLIP (CF-CL, VLEM-CL) perform better than 3D segment embeddings from Open-Fusion (OF-CL), albeit at a higher memory consumption, and that our CLIP embedding integration (VLEM-CL) performs better than ConceptFusion (CF-CL) while reducing the memory consumption by a large margin.

VLEM with any VLM backend provides the best trade-off between segmentation performance, spatial resolution, and memory requirements.
VLEM~(NARADIO) provides the best segmentation performance. RayFronts, using the same VLM (NARADIO), is the second-best approach, albeit using more than three times the peak VRAM of VLEM~(NARADIO).
Open-Fusion is the only approach that maintains embeddings for 3D segments instead of more refined points or voxels. This is the most compact representation of the evaluated approaches at the cost of also achieving the lowest segmentation performance due to the loss of spatial embedding resolution. VLEM~(CLIP) uses a less compact but more detailed spatial representation, thus using the second-lowest amount of peak VRAM but achieving a significant better segmentation performance than Open-Fusion.

\begin{table}
\centering
\caption{3D segmentation performance (mIoU [\%]) and peak VRAM consumption [GiB] on ScanNet for different integration approaches and embeddings (\underline{CL}IP, \underline{N}A\underline{R}ADIO, \underline{DI}NO). Averaged mIoU is provided over all NYU40 and ScanNet20 classes, excluding the background \texttt{unlabeled} and unspecific \texttt{other} classes.}
\begin{tabular}{lS[table-format=2.1]S[table-format=2.1]S[table-format=2.1]S[table-format=2.1]S[table-format=2.1]S[table-format=2.1]}
\toprule
                       & \textbf{CF}   & \textbf{OF}   & \textbf{RF}   & \multicolumn{3}{c}{\textbf{VLEM}} \\
\textbf{VLM}           & \textit{CL}   & \textit{CL}   & \textit{NR}   & \textit{CL}    & \textit{DI}   & \textit{NR} \\
\midrule
\textit{wall}          & 13.5          & \textbf{28.6} & 14.9          & 13.6 & 12.3          & 12.4          \\
\textit{floor}         & 15.2          & \textbf{16.5} & 09.2          & 13.4 & 13.8          & 11.2          \\
\textit{cabinet}       & 12.3          & 06.2          & 12.6          & 13.1 & 12.6          & \textbf{14.2} \\
\textit{bed}           & 36.1          & 26.1          & 44.6          & 38.9 & 41.0          & \textbf{44.9} \\
\textit{chair}         & \textbf{05.8} & 02.1          & 02.7          & 04.1 & 04.0          & 04.5          \\
\textit{sofa}          & 19.0          & 14.0          & 19.9          & 18.9 & 19.4          & \textbf{20.7} \\
\textit{table}         & 18.8          & 07.3          & 18.3          & 17.8 & 21.2          & \textbf{22.2} \\
\textit{door}          & 07.3          & 08.3          & 07.8          & 07.1 & 10.2          & \textbf{13.8} \\
\textit{window}        & 14.2          & 08.3          & 15.6          & 14.2 & 14.1          & \textbf{15.9} \\
\textit{bookshelf}     & 27.3          & 18.1          & 33.0          & 30.3 & 31.5          & \textbf{33.8} \\
\textit{picture}       & 23.8          & 19.8          & 29.5          & 24.6 & 26.8          & \textbf{33.6} \\
\textit{counter}       & 21.0          & 12.0          & 25.9          & 23.1 & 23.6          & \textbf{26.3} \\
\textit{blinds}        & 36.0          & 29.5          & 42.0          & 36.9 & 40.5          & \textbf{42.5} \\
\textit{desk}          & 08.5          & 04.0          & 08.8          & 08.8 & 09.5          & \textbf{10.6} \\
\textit{shelves}       & 16.5          & 10.1          & 18.9          & 17.0 & 21.3          & \textbf{19.5} \\
\textit{curtain}       & 35.5          & 27.8          & 42.6          & 37.5 & 40.7          & \textbf{43.3} \\
\textit{dresser}       & 36.1          & 20.8          & 43.9          & 38.7 & 40.7          & \textbf{44.1} \\
\textit{pillow}        & 28.1          & 18.7          & 29.3          & 25.5 & 27.1          & \textbf{36.8} \\
\textit{mirror}        & 37.2          & 30.9          & 42.9          & 37.4 & 40.2          & \textbf{43.7} \\
\textit{floormat}      & 30.0          & 18.3          & 31.9          & 30.6 & 30.2          & \textbf{32.1} \\
\textit{clothes}       & 36.6          & 30.5          & 45.1          & 38.5 & 41.3          & \textbf{45.5} \\
\textit{ceiling}       & 23.3          & 14.3          & 12.2          & 24.0 & \textbf{25.2} & 23.9          \\
\textit{books}         & 19.0          & 14.5          & 23.2          & 19.7 & 21.5          & \textbf{24.6} \\
\textit{refrigerator}  & 18.9          & 12.9          & 23.5          & 20.1 & 22.0          & \textbf{27.4} \\
\textit{television}    & 27.4          & 22.5          & 33.1          & 26.3 & 30.6          & \textbf{34.1} \\
\textit{paper}         & 26.6          & 14.6          & \textbf{32.9} & 28.0 & 30.6          & \textbf{32.9} \\
\textit{towel}         & 36.1          & 26.8          & 44.4          & 38.1 & 41.3          & \textbf{44.7} \\
\textit{showercurtain} & 36.3          & 31.6          & 43.2          & 37.8 & 42.6          & \textbf{44.3} \\
\textit{box}           & 22.0          & 15.3          & 25.6          & 22.2 & 24.0          & \textbf{25.8} \\
\textit{whiteboard}    & 35.7          & 23.3          & 43.9          & 36.6 & 40.4          & \textbf{44.1} \\
\textit{person}        & 37.0          & 34.2          & 47.6          & 37.9 & 42.7          & \textbf{48.3} \\
\textit{nightstand}    & 36.3          & 30.0          & 44.0          & 39.1 & 40.8          & \textbf{44.1} \\
\textit{toilet}        & 36.1          & 26.5          & 45.0          & 38.4 & 41.7          & \textbf{45.5} \\
\textit{sink}          & 24.1          & 21.4          & 26.3          & 22.3 & 24.1          & \textbf{26.7} \\
\textit{lamp}          & 18.2          & 16.9          & \textbf{22.3} & 19.7 & 20.6          & \textbf{22.3} \\
\textit{bathtub}       & 37.2          & 31.3          & 45.5          & 39.5 & 42.1          & \textbf{46.2} \\
\textit{bag}           & 24.0          & 15.8          & 28.8          & 26.1 & 26.8          & \textbf{28.9} \\
\midrule
NYU40 avg.             & 25.3          & 19.2          & \underline{29.2}    & 26.1 & 28.1          & \textbf{30.7} \\
ScanNet20 avg.         & 21.6          & 17.0          & \underline{24.7}    & 22.3 & 23.9          & \textbf{26.2} \\
\midrule
peak VRAM [GiB]        &  8.8          & \textbf{3.2}  & 23.4          & \underline{4.0} & 10.9          &  7.1          \\
\bottomrule
\end{tabular}
\label{tab:scannet_miou_mem}
\vspace{-0.5cm}
\end{table}

\subsection{Qualitative Results}

\begin{figure*}
% \vspace{-3mm}
\centering
\newcommand{\w}{0.19} % new size
\newcommand{\imgwidth}{\w\textwidth}

\setlength{\tabcolsep}{1pt}

\makebox[\textwidth][c]{
\begin{tabular}{cccccc}
\toprule
& \textbf{ConceptFusion} & \textbf{LangSplat} & \textbf{RayFronts} & \textbf{Open-Fusion} & \textbf{VLEM~(CLIP)} \\
\midrule

\rotatebox[origin=c]{90}{colour} &
\adjustbox{valign=c}{\begin{tikzpicture}
    \node[anchor=south west,inner sep=0] at (0,0) {\includegraphics[width=\imgwidth,align=c]{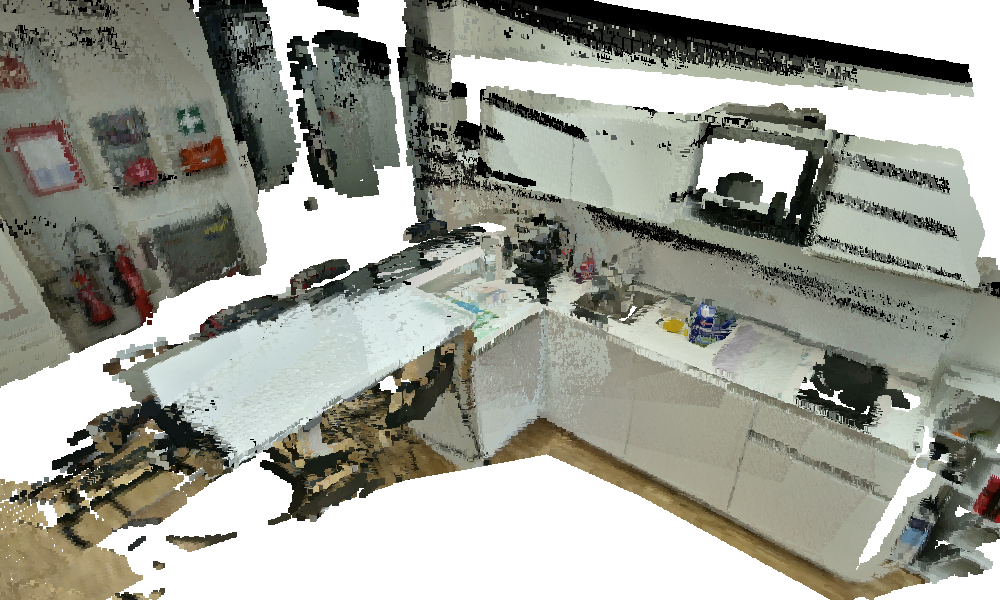}};
    \draw[red!80,thick,rounded corners=0.2mm] (\w*4*2.2,\w*4*1.35) rectangle (\w*4*2.55,\w*4*1.85); % coffee machine
    \draw[red!80,thick,rounded corners=0.2mm] (\w*4*2.5,\w*4*1.15) rectangle (\w*4*3.0,\w*4*1.55); % sink
    \draw[red!80,thick,rounded corners=0.2mm] (\w*4*3.0,\w*4*1.1) rectangle (\w*4*3.3,\w*4*1.4); % milk
    \draw[red!80,thick,rounded corners=0.2mm] (\w*4*3.2,\w*4*1.3) rectangle (\w*4*3.55,\w*4*1.5); % socket
\end{tikzpicture}}
&
\adjustbox{valign=c}{\begin{tikzpicture}
    \node[anchor=south west,inner sep=0] at (0,0) {\includegraphics[width=\imgwidth,align=c]{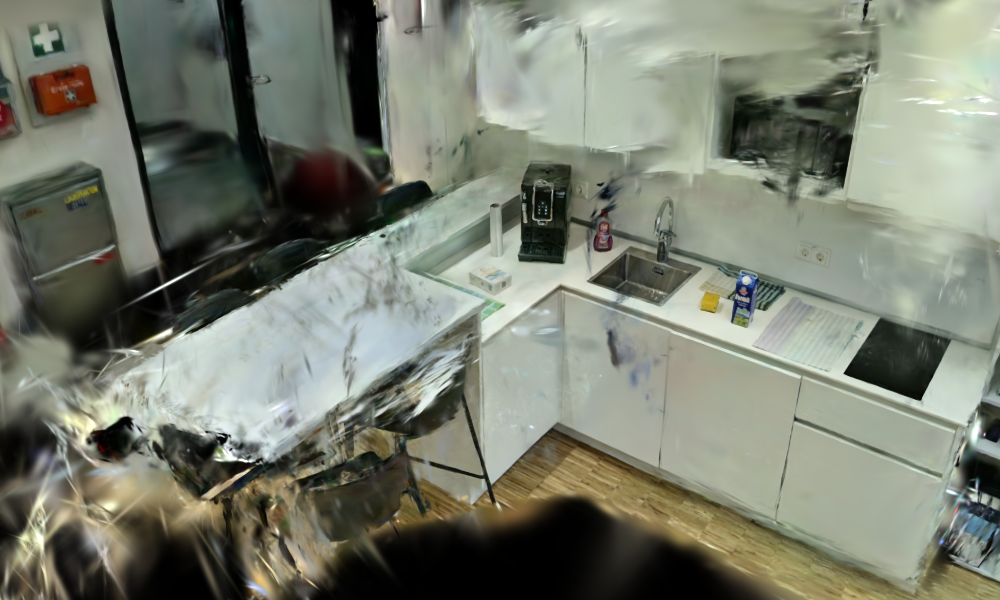}};
    \draw[red!80,thick,rounded corners=0.2mm] (\w*4*2.25,\w*4*1.45) rectangle (\w*4*2.65,\w*4*2.05); % coffee machine
    \draw[red!80,thick,rounded corners=0.2mm] (\w*4*2.6,\w*4*1.25) rectangle (\w*4*3.2,\w*4*1.65); % sink
    \draw[red!80,thick,rounded corners=0.2mm] (\w*4*3.2,\w*4*1.15) rectangle (\w*4*3.45,\w*4*1.55); % milk
    \draw[red!80,thick,rounded corners=0.2mm] (\w*4*3.5,\w*4*1.45) rectangle (\w*4*3.75,\w*4*1.65); % socket
\end{tikzpicture}}
&
\adjustbox{valign=c}{\begin{tikzpicture}
    \node[anchor=south west,inner sep=0] at (0,0) {\includegraphics[width=\imgwidth,align=c]{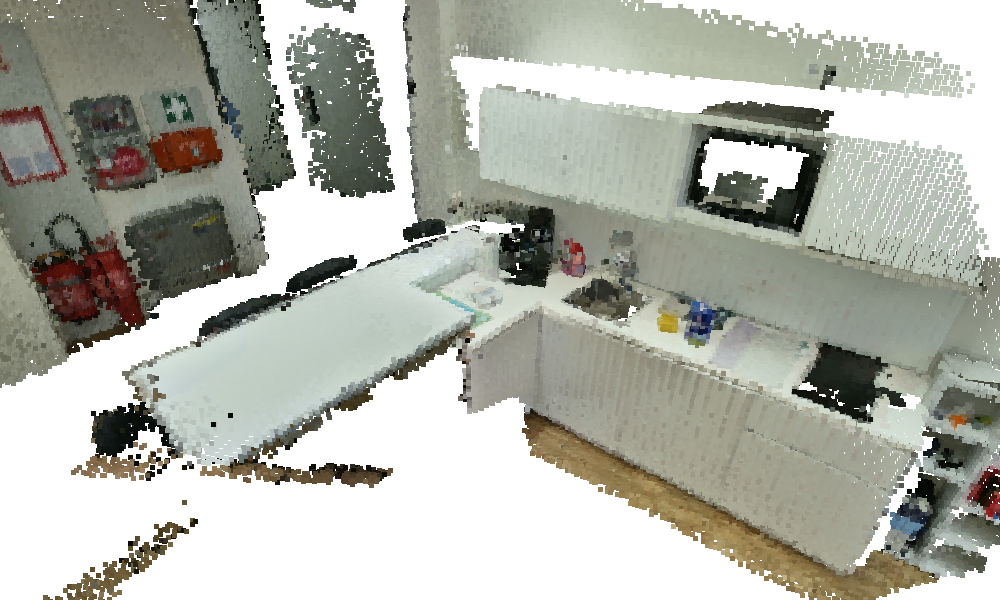}};
    \draw[red!80,thick,rounded corners=0.2mm] (\w*4*2.2,\w*4*1.35) rectangle (\w*4*2.55,\w*4*1.85); % coffee machine
    \draw[red!80,thick,rounded corners=0.2mm] (\w*4*2.5,\w*4*1.15) rectangle (\w*4*3.0,\w*4*1.55); % sink
    \draw[red!80,thick,rounded corners=0.2mm] (\w*4*3.0,\w*4*1.1) rectangle (\w*4*3.3,\w*4*1.4); % milk
    \draw[red!80,thick,rounded corners=0.2mm] (\w*4*3.2,\w*4*1.3) rectangle (\w*4*3.55,\w*4*1.5); % socket
\end{tikzpicture}}
&
\adjustbox{valign=c}{\begin{tikzpicture}
    \node[anchor=south west,inner sep=0] at (0,0) {\includegraphics[width=\imgwidth,align=c]{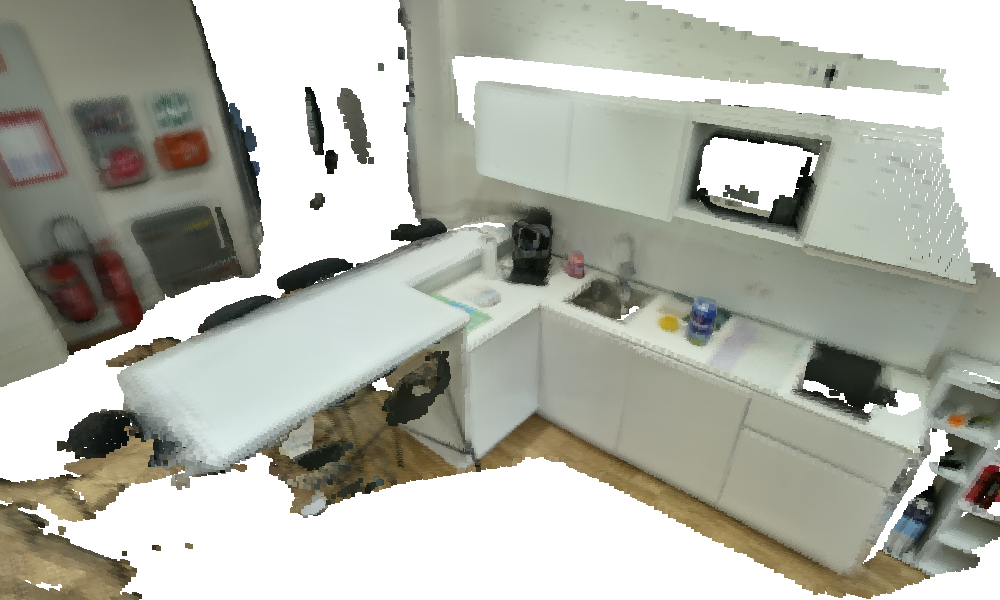}};
    \draw[red!80,thick,rounded corners=0.2mm] (\w*4*2.2,\w*4*1.35) rectangle (\w*4*2.55,\w*4*1.85); % coffee machine
    \draw[red!80,thick,rounded corners=0.2mm] (\w*4*2.5,\w*4*1.15) rectangle (\w*4*3.0,\w*4*1.55); % sink
    \draw[red!80,thick,rounded corners=0.2mm] (\w*4*3.0,\w*4*1.1) rectangle (\w*4*3.3,\w*4*1.4); % milk
    \draw[red!80,thick,rounded corners=0.2mm] (\w*4*3.2,\w*4*1.3) rectangle (\w*4*3.55,\w*4*1.5); % socket
\end{tikzpicture}}
&
\adjustbox{valign=c}{\begin{tikzpicture}
    \node[anchor=south west,inner sep=0] at (0,0) {\includegraphics[width=\imgwidth,align=c]{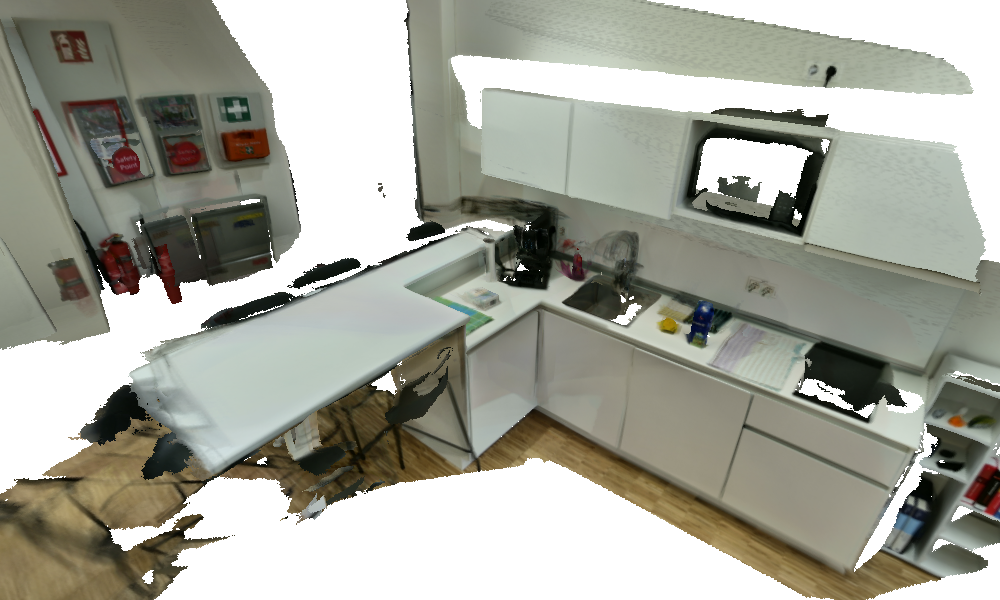}};
    \draw[red!80,thick,rounded corners=0.2mm] (\w*4*2.2,\w*4*1.35) rectangle (\w*4*2.55,\w*4*1.85); % coffee machine
    \draw[red!80,thick,rounded corners=0.2mm] (\w*4*2.5,\w*4*1.15) rectangle (\w*4*3.0,\w*4*1.55); % sink
    \draw[red!80,thick,rounded corners=0.2mm] (\w*4*3.0,\w*4*1.1) rectangle (\w*4*3.3,\w*4*1.4); % milk
    \draw[red!80,thick,rounded corners=0.2mm] (\w*4*3.2,\w*4*1.3) rectangle (\w*4*3.55,\w*4*1.5); % socket
\end{tikzpicture}}
\\

\rotatebox[origin=c]{90}{\textit{coffee machine}} &
\adjustbox{valign=c}{\begin{tikzpicture}
    \node[anchor=south west,inner sep=0] at (0,0) {\includegraphics[width=\imgwidth,align=c]{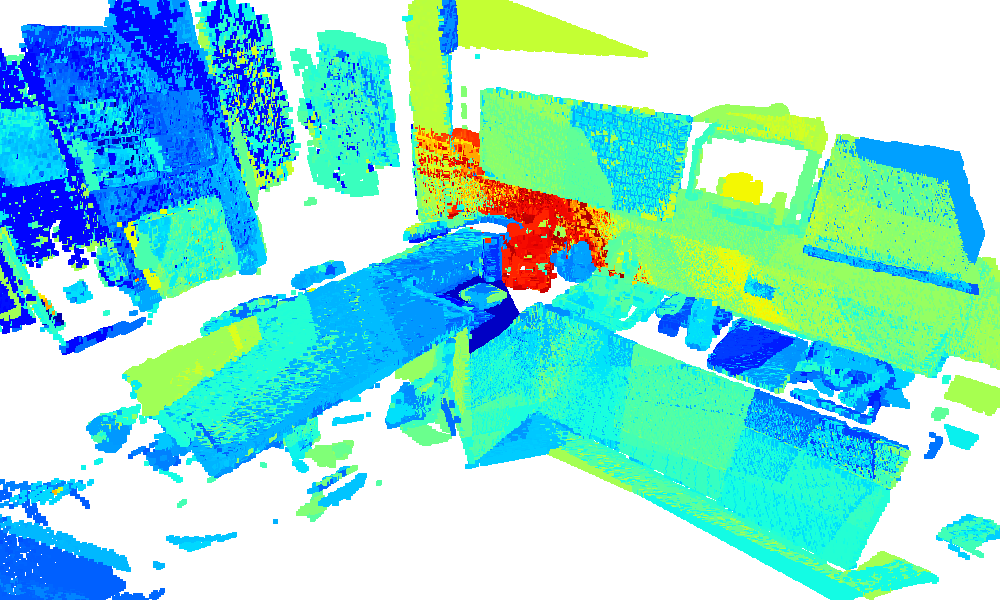}};
    \draw[black,thick,rounded corners=0.2mm] (\w*4*2.2,\w*4*1.35) rectangle (\w*4*2.55,\w*4*1.85); % coffee machine
\end{tikzpicture}}
&
\adjustbox{valign=c}{\begin{tikzpicture}
    \node[anchor=south west,inner sep=0] at (0,0) {\includegraphics[width=\imgwidth,align=c]{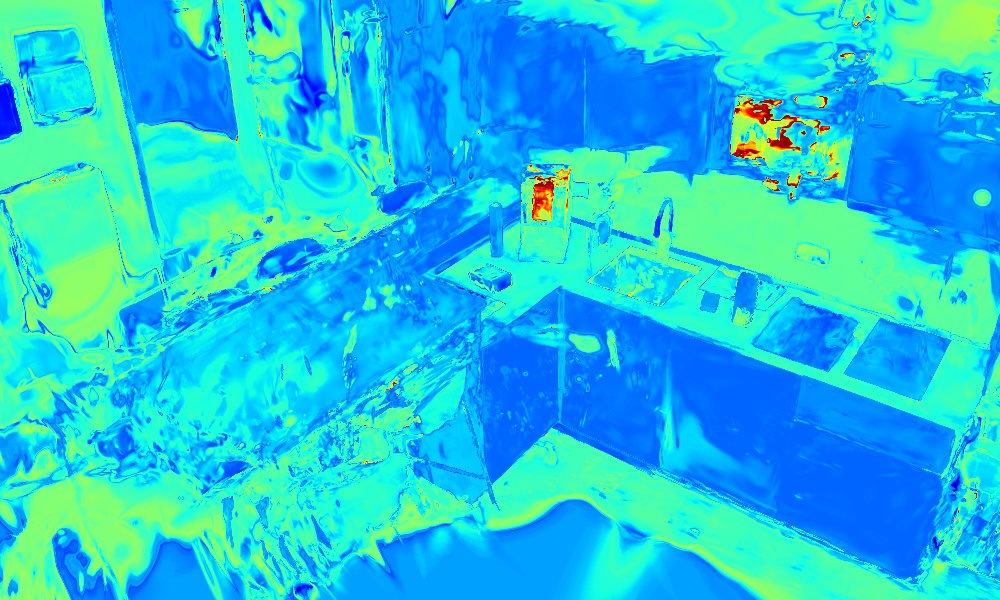}};
    \draw[black,thick,rounded corners=0.2mm] (\w*4*2.25,\w*4*1.45) rectangle (\w*4*2.65,\w*4*2.05); % coffee machine
\end{tikzpicture}}
&
\adjustbox{valign=c}{\begin{tikzpicture}
    \node[anchor=south west,inner sep=0] at (0,0) {\includegraphics[width=\imgwidth,align=c]{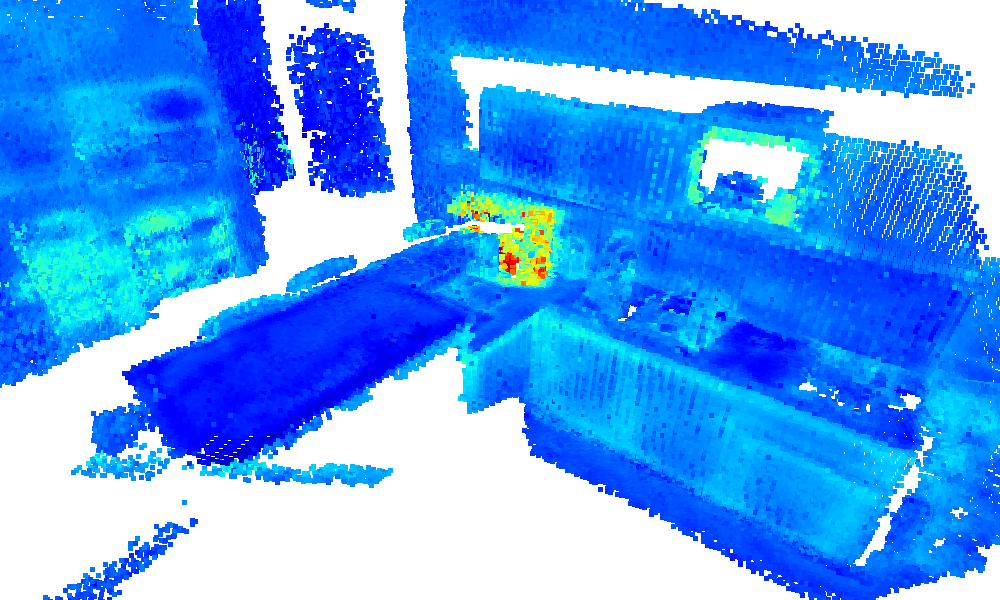}};
    \draw[black,thick,rounded corners=0.2mm] (\w*4*2.2,\w*4*1.35) rectangle (\w*4*2.55,\w*4*1.85); % coffee machine
\end{tikzpicture}}
&
\adjustbox{valign=c}{\begin{tikzpicture}
    \node[anchor=south west,inner sep=0] at (0,0) {\includegraphics[width=\imgwidth,align=c]{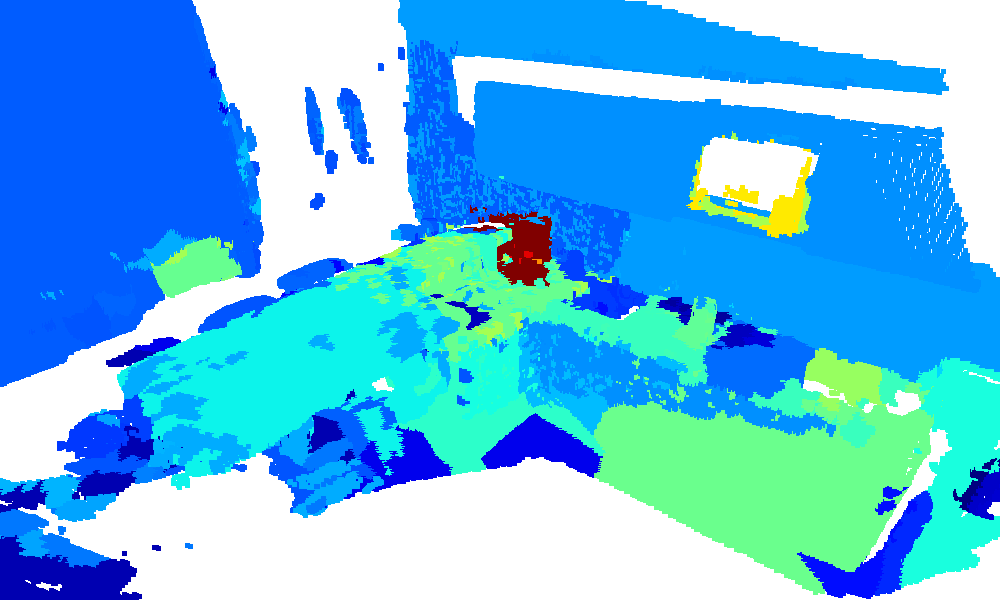}};
    \draw[black,thick,rounded corners=0.2mm] (\w*4*2.2,\w*4*1.35) rectangle (\w*4*2.55,\w*4*1.85); % coffee machine
\end{tikzpicture}}
&
\adjustbox{valign=c}{\begin{tikzpicture}
    \node[anchor=south west,inner sep=0] at (0,0) {\includegraphics[width=\imgwidth,align=c]{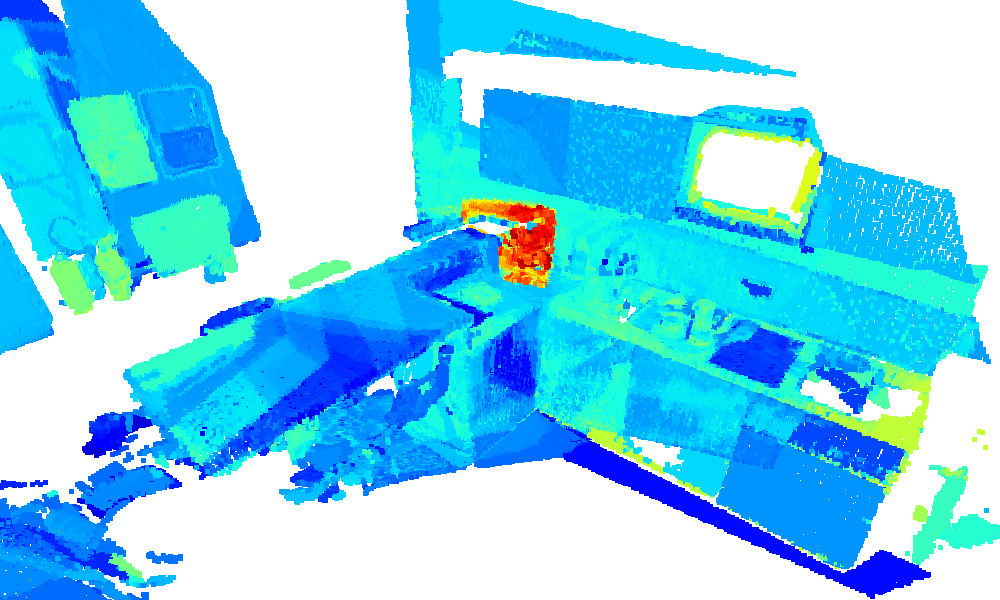}};
    \draw[black,thick,rounded corners=0.2mm] (\w*4*2.2,\w*4*1.35) rectangle (\w*4*2.55,\w*4*1.85); % coffee machine
\end{tikzpicture}}
\\

\rotatebox[origin=c]{90}{\textit{milk}} &
\adjustbox{valign=c}{\begin{tikzpicture}
    \node[anchor=south west,inner sep=0] at (0,0) {\includegraphics[width=\imgwidth,align=c]{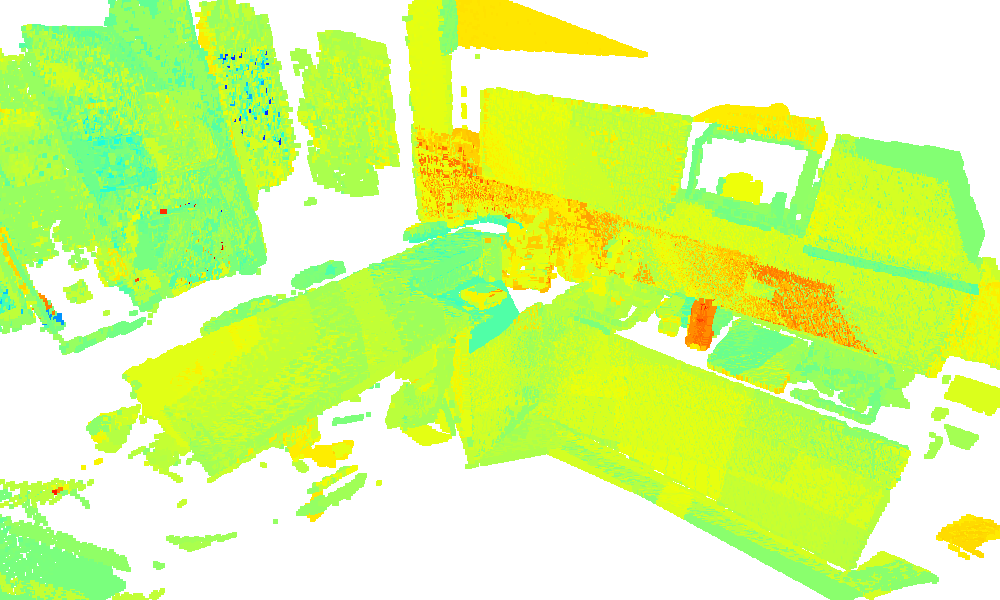}};
    \draw[black,thick,rounded corners=0.2mm] (\w*4*3.0,\w*4*1.1) rectangle (\w*4*3.3,\w*4*1.4); % milk
\end{tikzpicture}}
&
\adjustbox{valign=c}{\begin{tikzpicture}
    \node[anchor=south west,inner sep=0] at (0,0) {\includegraphics[width=\imgwidth,align=c]{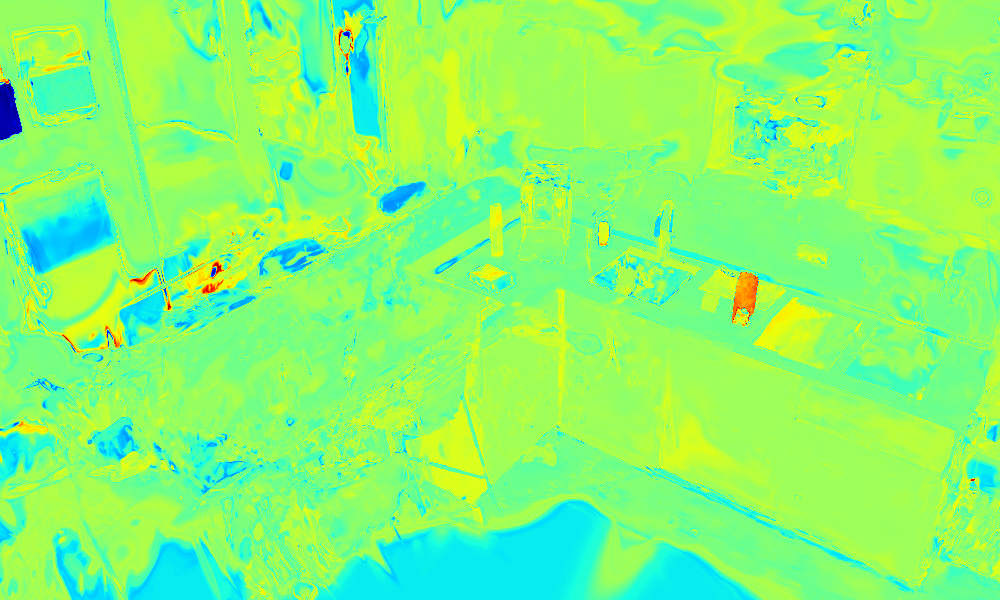}};
    \draw[black,thick,rounded corners=0.2mm] (\w*4*3.2,\w*4*1.15) rectangle (\w*4*3.45,\w*4*1.55); % milk
\end{tikzpicture}}
&
\adjustbox{valign=c}{\begin{tikzpicture}
    \node[anchor=south west,inner sep=0] at (0,0) {\includegraphics[width=\imgwidth,align=c]{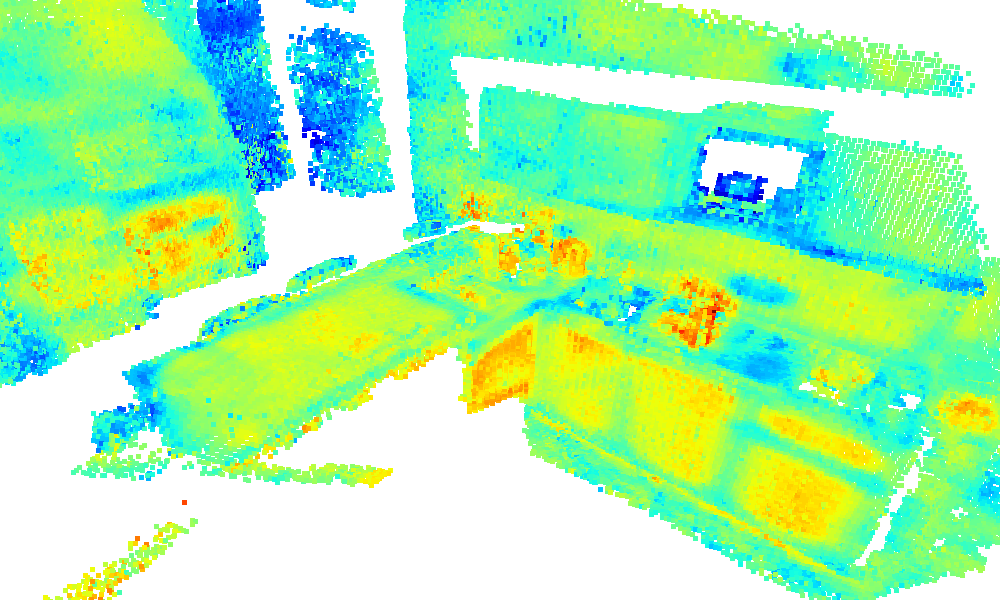}};
    \draw[black,thick,rounded corners=0.2mm] (\w*4*3.0,\w*4*1.1) rectangle (\w*4*3.3,\w*4*1.4); % milk
\end{tikzpicture}}
&
\adjustbox{valign=c}{\begin{tikzpicture}
    \node[anchor=south west,inner sep=0] at (0,0) {\includegraphics[width=\imgwidth,align=c]{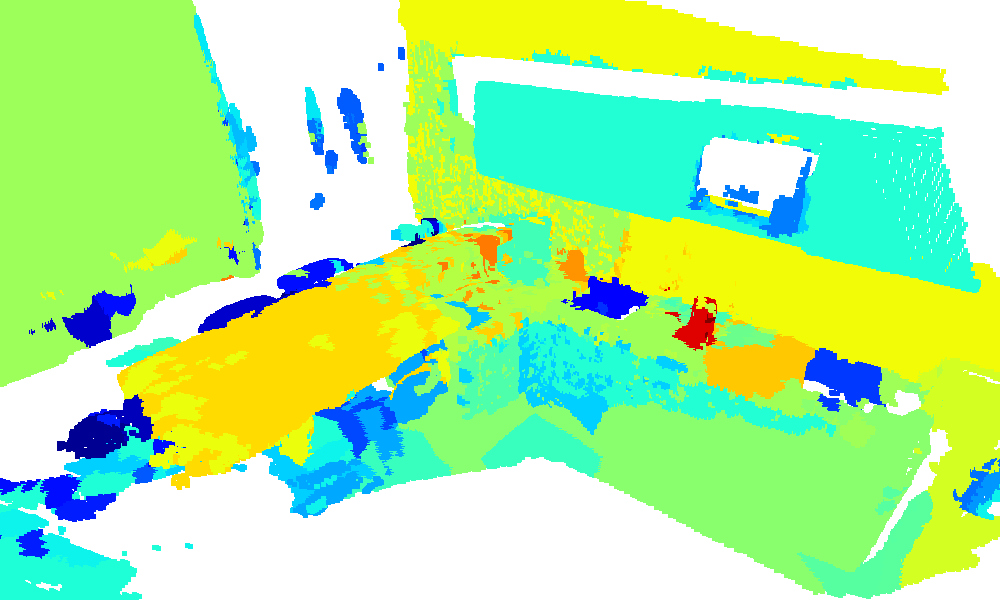}};
    \draw[black,thick,rounded corners=0.2mm] (\w*4*3.0,\w*4*1.1) rectangle (\w*4*3.3,\w*4*1.4); % milk
\end{tikzpicture}}
&
\adjustbox{valign=c}{\begin{tikzpicture}
    \node[anchor=south west,inner sep=0] at (0,0) {\includegraphics[width=\imgwidth,align=c]{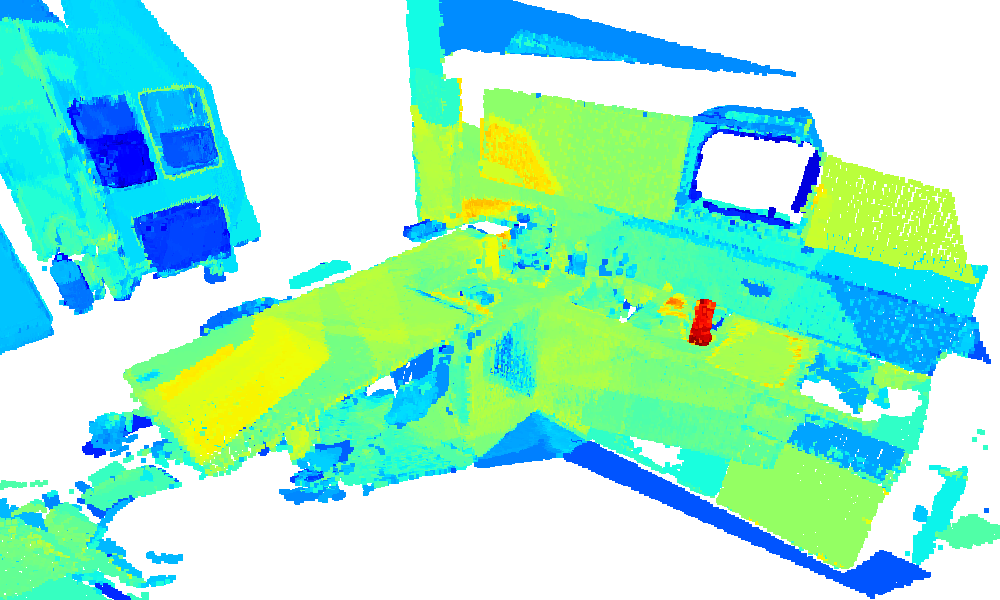}};
    \draw[black,thick,rounded corners=0.2mm] (\w*4*3.0,\w*4*1.1) rectangle (\w*4*3.3,\w*4*1.4); % milk
\end{tikzpicture}}
\\

\rotatebox[origin=c]{90}{\textit{sink}} &
\adjustbox{valign=c}{\begin{tikzpicture}
    \node[anchor=south west,inner sep=0] at (0,0) {\includegraphics[width=\imgwidth,align=c]{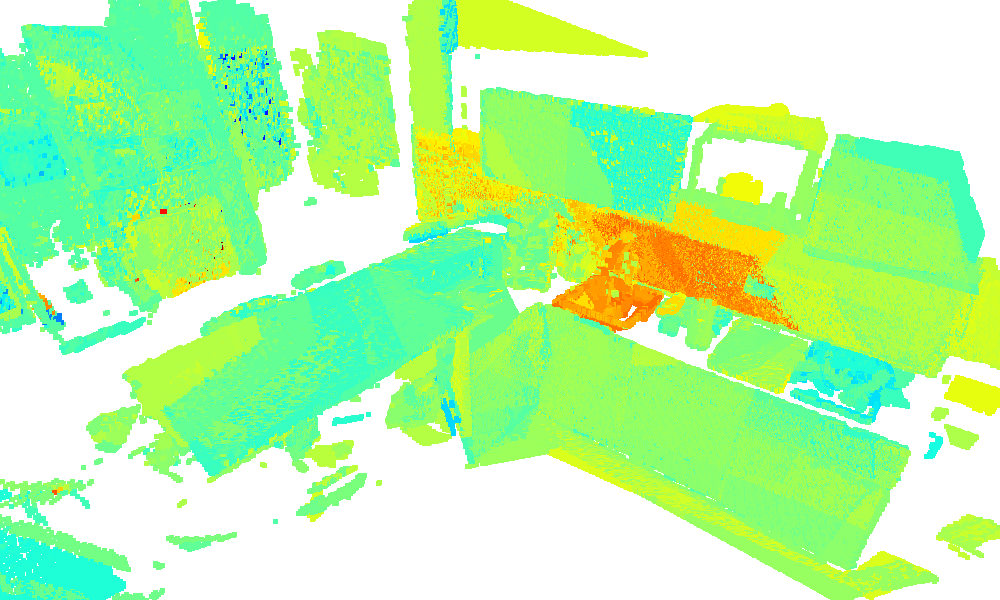}};
    \draw[black,thick,rounded corners=0.2mm] (\w*4*2.5,\w*4*1.15) rectangle (\w*4*3.0,\w*4*1.55); % sink
\end{tikzpicture}}
&
\adjustbox{valign=c}{\begin{tikzpicture}
    \node[anchor=south west,inner sep=0] at (0,0) {\includegraphics[width=\imgwidth,align=c]{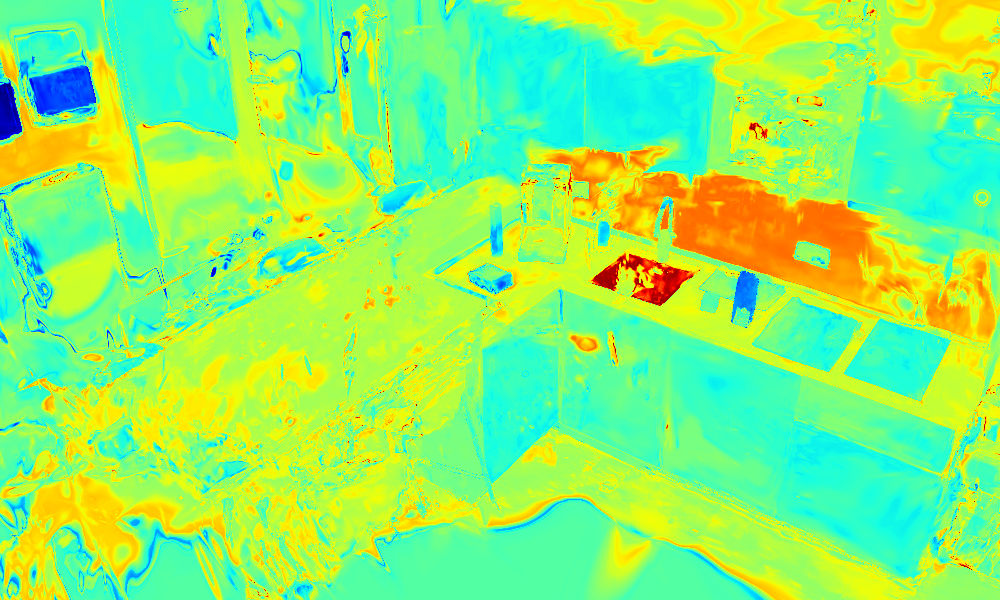}};
    \draw[black,thick,rounded corners=0.2mm] (\w*4*2.6,\w*4*1.25) rectangle (\w*4*3.2,\w*4*1.65); % sink
\end{tikzpicture}}
&
\adjustbox{valign=c}{\begin{tikzpicture}
    \node[anchor=south west,inner sep=0] at (0,0) {\includegraphics[width=\imgwidth,align=c]{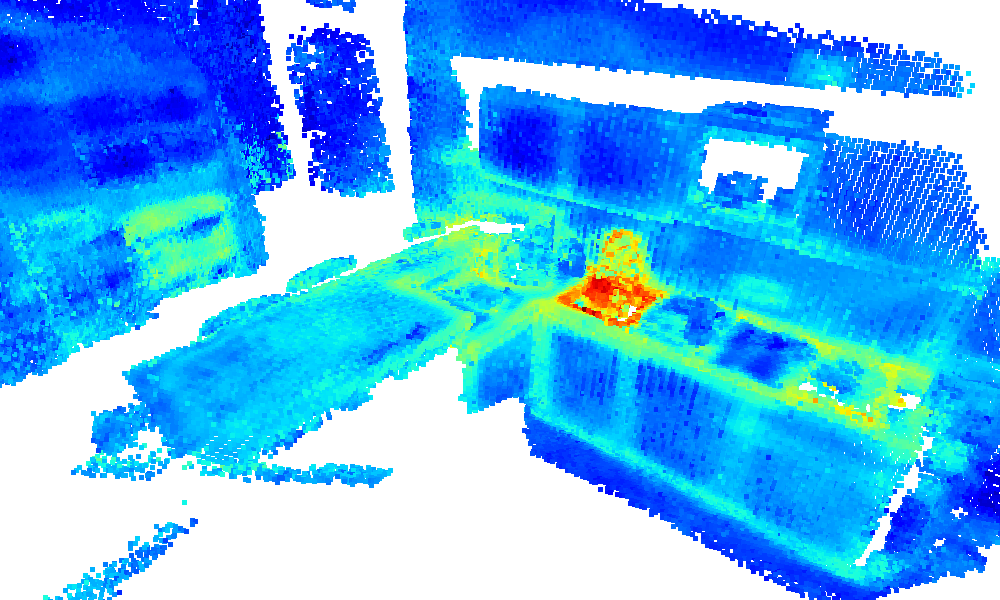}};
    \draw[black,thick,rounded corners=0.2mm] (\w*4*2.5,\w*4*1.15) rectangle (\w*4*3.0,\w*4*1.55); % sink
\end{tikzpicture}}
&
\adjustbox{valign=c}{\begin{tikzpicture}
    \node[anchor=south west,inner sep=0] at (0,0) {\includegraphics[width=\imgwidth,align=c]{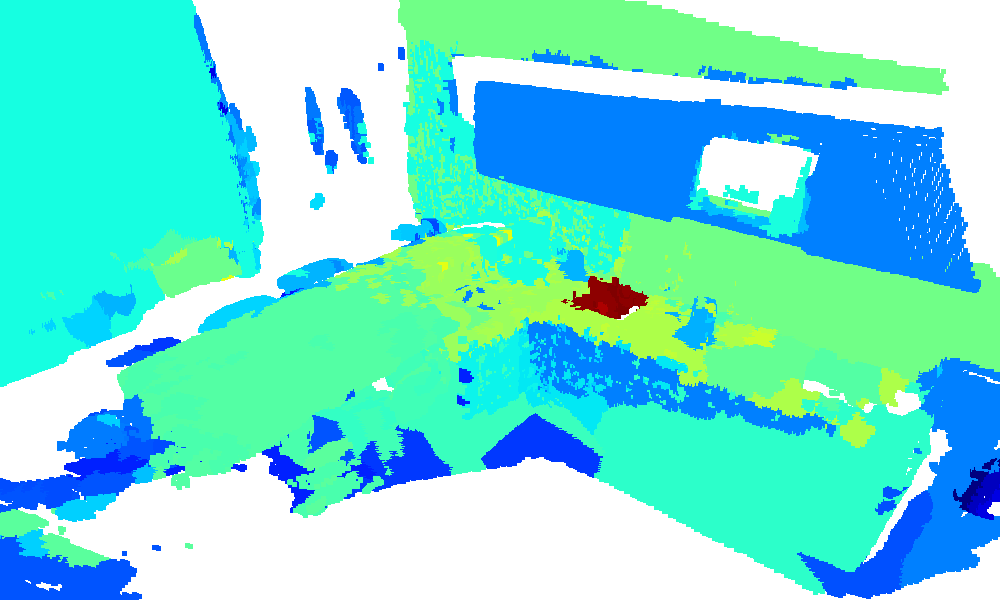}};
    \draw[black,thick,rounded corners=0.2mm] (\w*4*2.5,\w*4*1.15) rectangle (\w*4*3.0,\w*4*1.55); % sink
\end{tikzpicture}}
&
\adjustbox{valign=c}{\begin{tikzpicture}
    \node[anchor=south west,inner sep=0] at (0,0) {\includegraphics[width=\imgwidth,align=c]{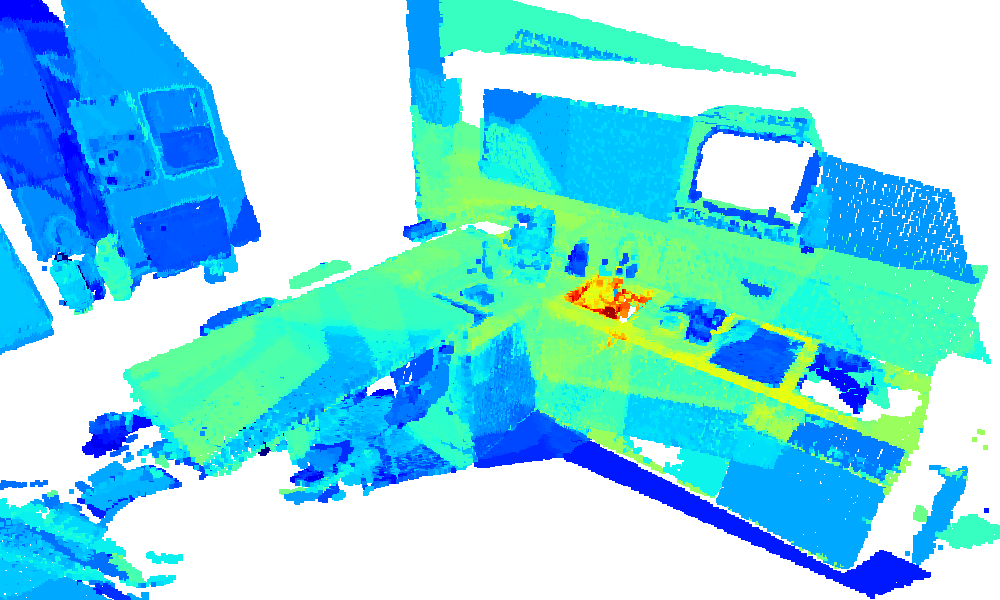}};
    \draw[black,thick,rounded corners=0.2mm] (\w*4*2.5,\w*4*1.15) rectangle (\w*4*3.0,\w*4*1.55); % sink
\end{tikzpicture}}
\\

\rotatebox[origin=c]{90}{\textit{socket}} &
\adjustbox{valign=c}{\begin{tikzpicture}
    \node[anchor=south west,inner sep=0] at (0,0) {\includegraphics[width=\imgwidth,align=c]{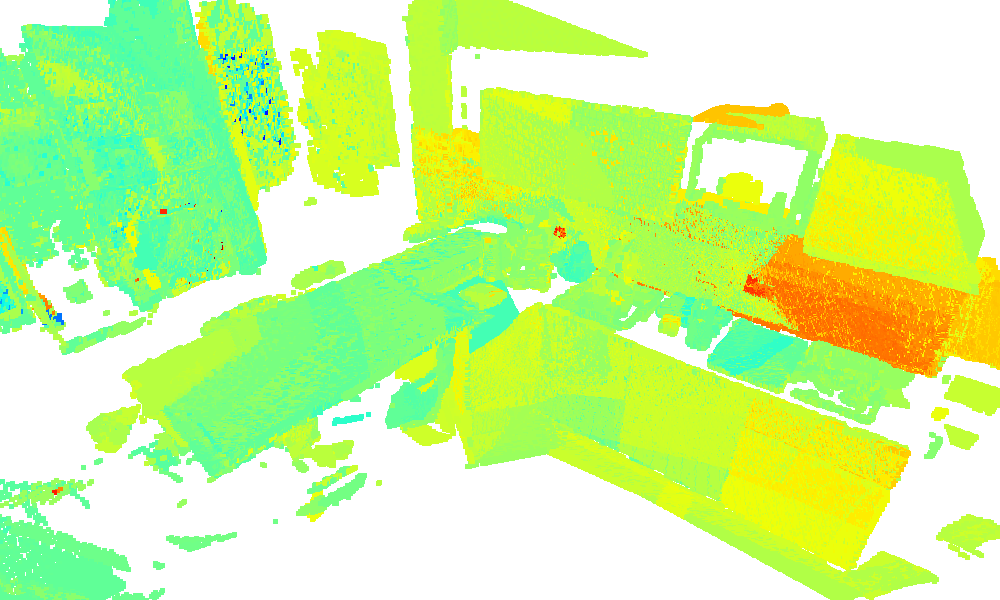}};
    \draw[black,thick,rounded corners=0.2mm] (\w*4*3.2,\w*4*1.3) rectangle (\w*4*3.55,\w*4*1.5); % socket
\end{tikzpicture}}
&
\adjustbox{valign=c}{\begin{tikzpicture}
    \node[anchor=south west,inner sep=0] at (0,0) {\includegraphics[width=\imgwidth,align=c]{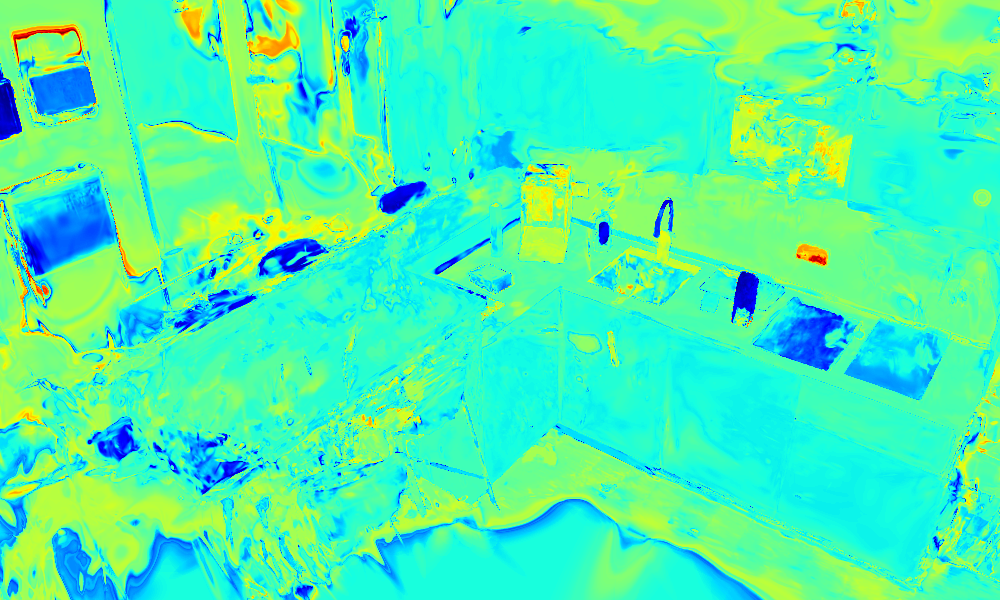}};
    \draw[black,thick,rounded corners=0.2mm] (\w*4*3.5,\w*4*1.45) rectangle (\w*4*3.75,\w*4*1.65); % socket
\end{tikzpicture}}
&
\adjustbox{valign=c}{\begin{tikzpicture}
    \node[anchor=south west,inner sep=0] at (0,0) {\includegraphics[width=\imgwidth,align=c]{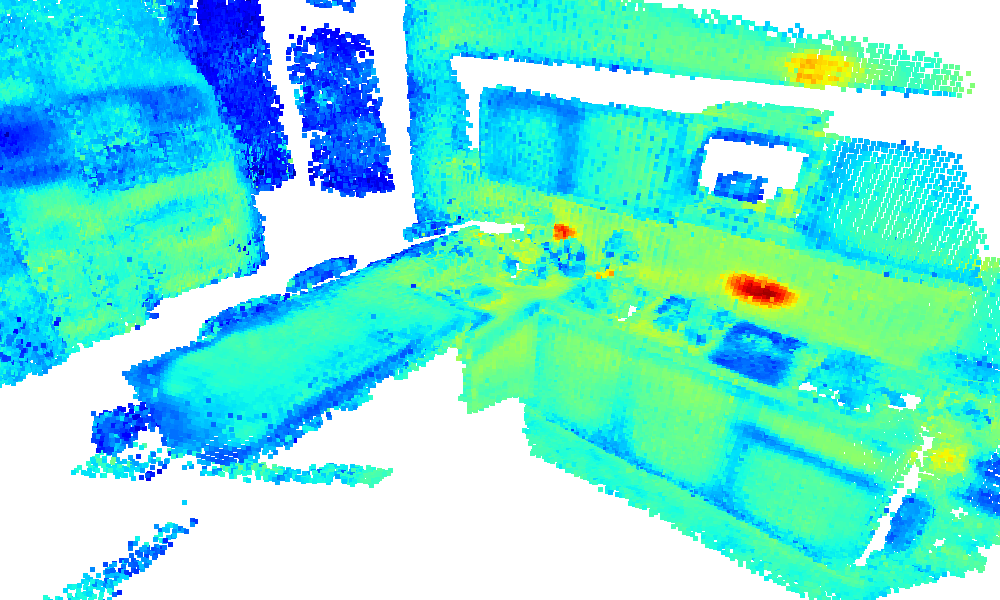}};
    \draw[black,thick,rounded corners=0.2mm] (\w*4*3.2,\w*4*1.3) rectangle (\w*4*3.55,\w*4*1.5); % socket
\end{tikzpicture}}
&
\adjustbox{valign=c}{\begin{tikzpicture}
    \node[anchor=south west,inner sep=0] at (0,0) {\includegraphics[width=\imgwidth,align=c]{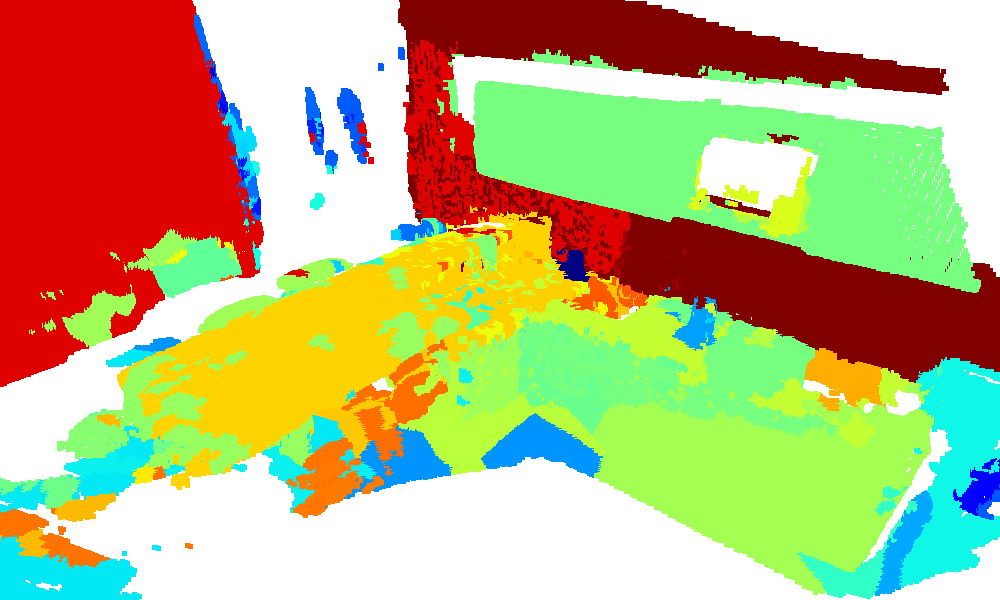}};
    \draw[black,thick,rounded corners=0.2mm] (\w*4*3.2,\w*4*1.3) rectangle (\w*4*3.55,\w*4*1.5); % socket
\end{tikzpicture}}
&
\adjustbox{valign=c}{\begin{tikzpicture}
    \node[anchor=south west,inner sep=0] at (0,0) {\includegraphics[width=\imgwidth,align=c]{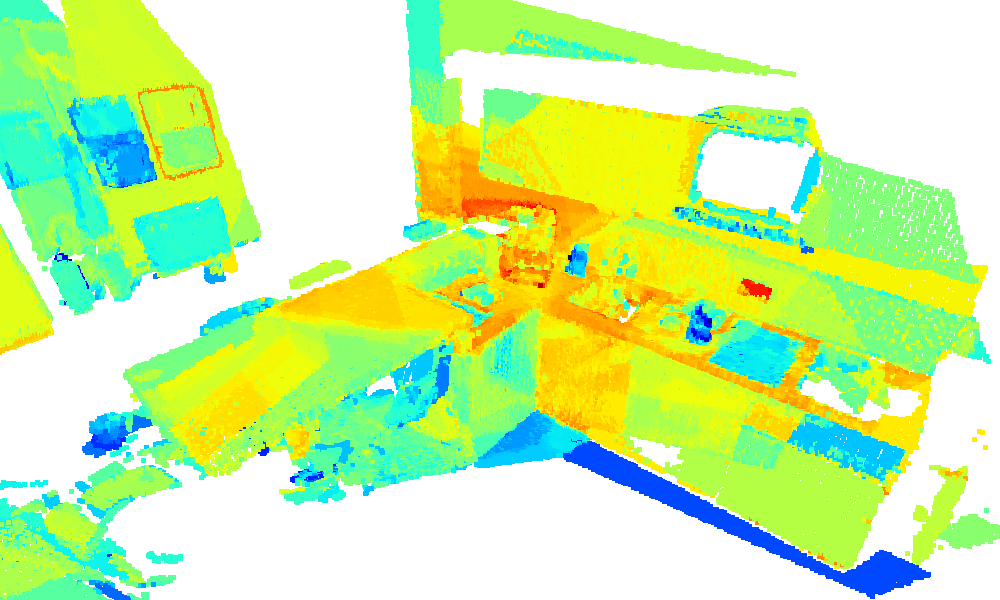}};
    \draw[black,thick,rounded corners=0.2mm] (\w*4*3.2,\w*4*1.3) rectangle (\w*4*3.55,\w*4*1.5); % socket
\end{tikzpicture}}
\\

\rotatebox[origin=c]{90}{\textit{sponge}} &
\adjustbox{valign=c}{\begin{tikzpicture}
    \node[anchor=south west,inner sep=0] at (0,0) {\includegraphics[width=\imgwidth,align=c]{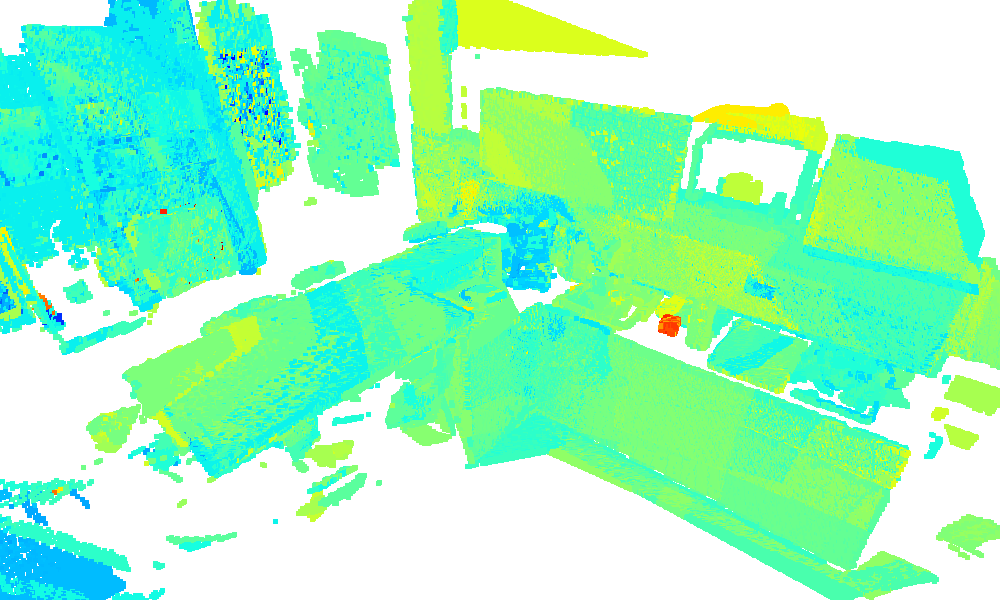}};
\end{tikzpicture}}
&
\adjustbox{valign=c}{\begin{tikzpicture}
    \node[anchor=south west,inner sep=0] at (0,0) {\includegraphics[width=\imgwidth,align=c]{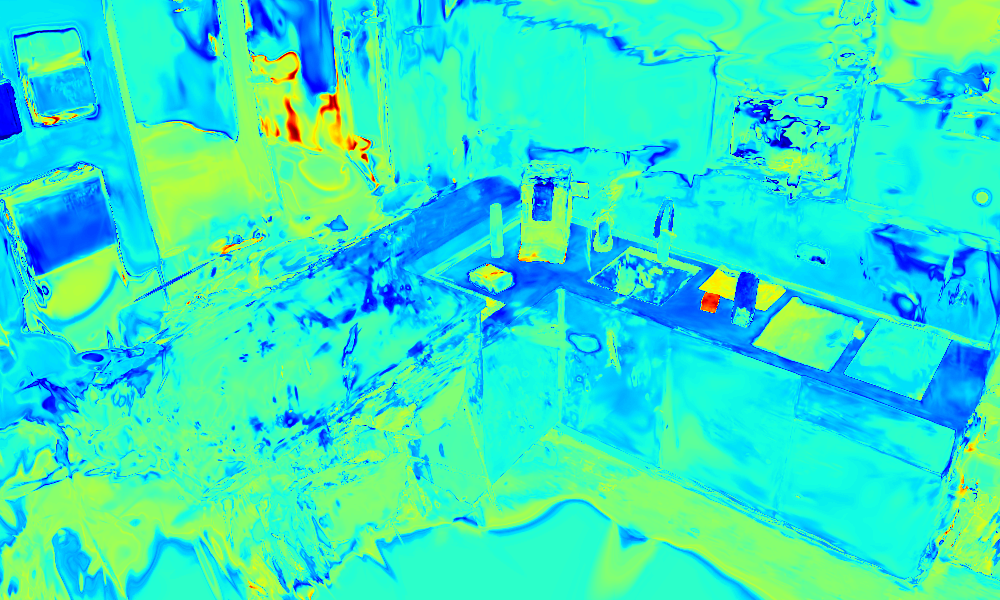}};
\end{tikzpicture}}
&
\adjustbox{valign=c}{\begin{tikzpicture}
    \node[anchor=south west,inner sep=0] at (0,0) {\includegraphics[width=\imgwidth,align=c]{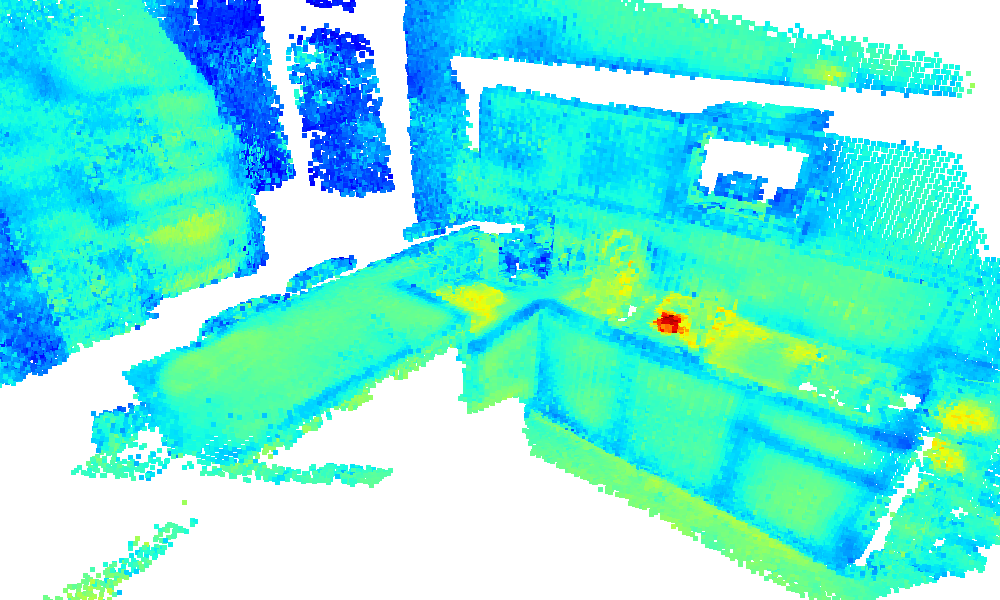}};
\end{tikzpicture}}
&
\adjustbox{valign=c}{\begin{tikzpicture}
    \node[anchor=south west,inner sep=0] at (0,0) {\includegraphics[width=\imgwidth,align=c]{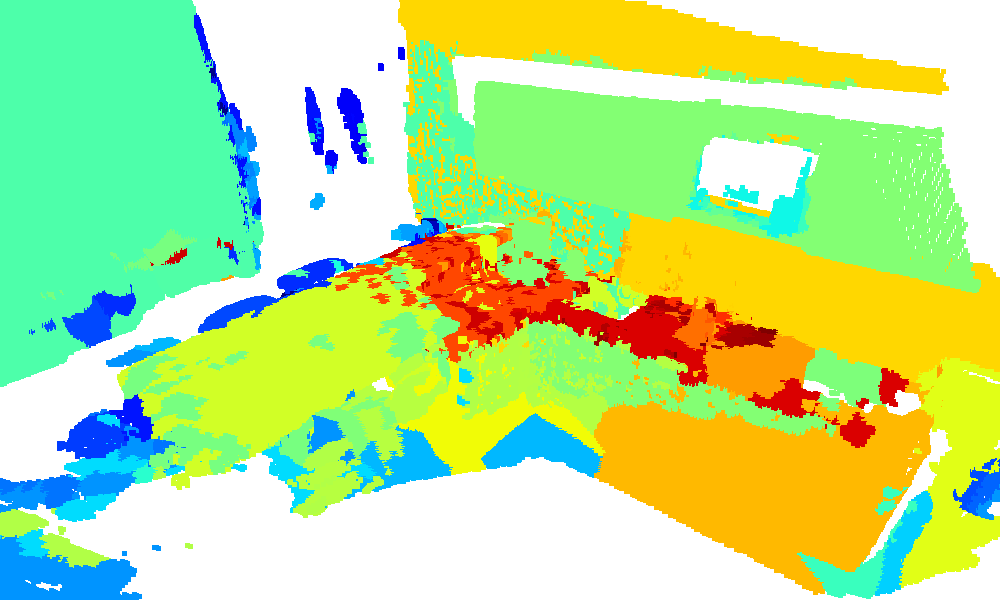}};
\end{tikzpicture}}
&
\adjustbox{valign=c}{\begin{tikzpicture}
    \node[anchor=south west,inner sep=0] at (0,0) {\includegraphics[width=\imgwidth,align=c]{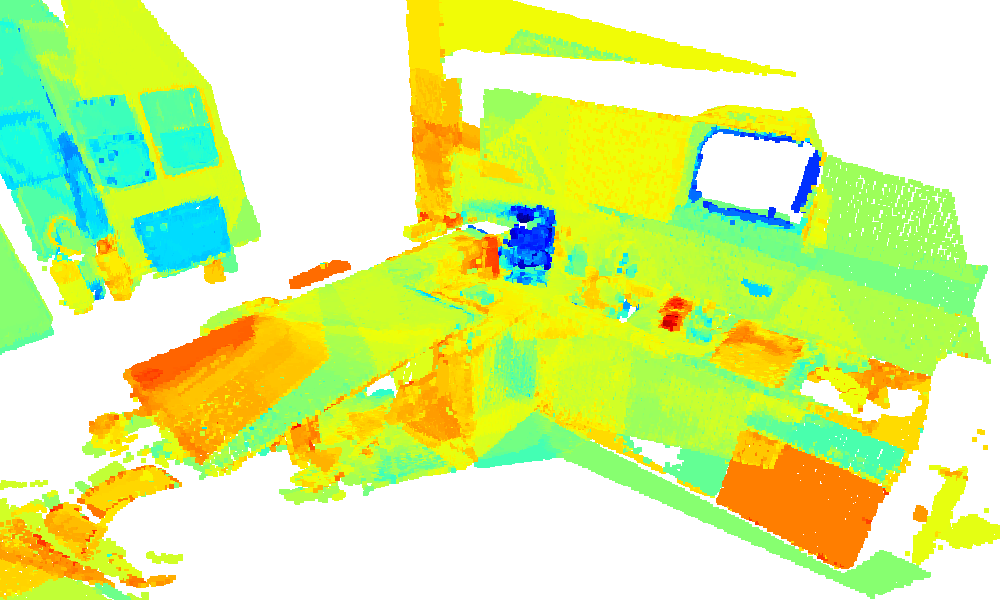}};
\end{tikzpicture}}
\\

\rotatebox[origin=c]{90}{\textit{table top}} &
\adjustbox{valign=c}{\begin{tikzpicture}
    \node[anchor=south west,inner sep=0] at (0,0) {\includegraphics[width=\imgwidth,align=c]{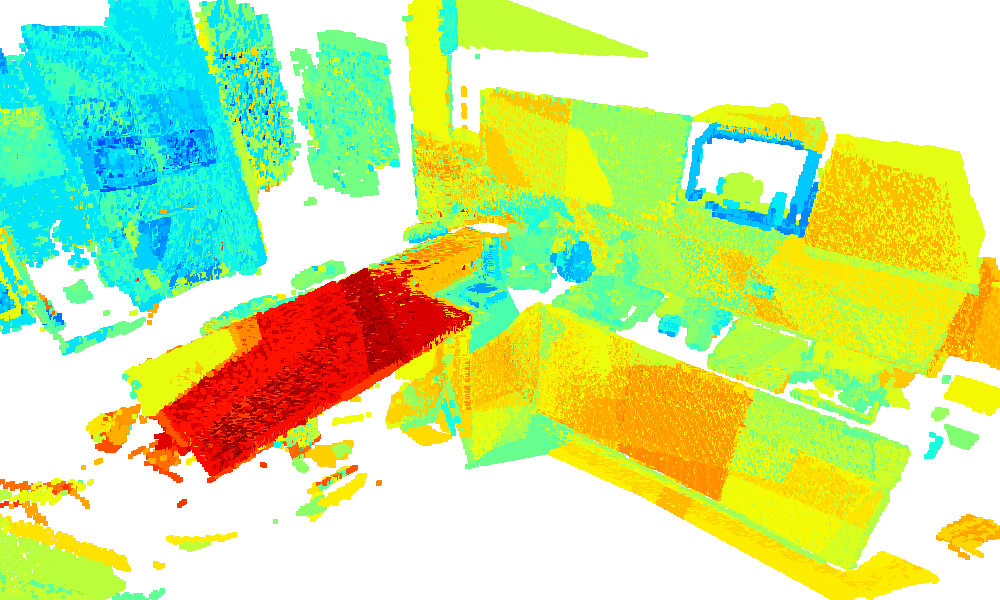}};
\end{tikzpicture}}
&
\adjustbox{valign=c}{\begin{tikzpicture}
    \node[anchor=south west,inner sep=0] at (0,0) {\includegraphics[width=\imgwidth,align=c]{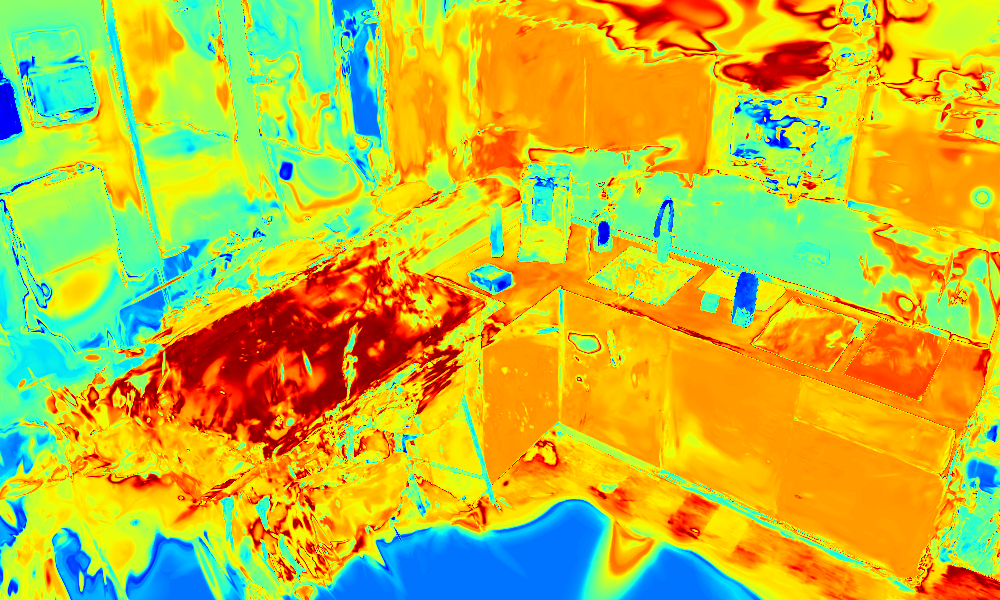}};
\end{tikzpicture}}
&
\adjustbox{valign=c}{\begin{tikzpicture}
    \node[anchor=south west,inner sep=0] at (0,0) {\includegraphics[width=\imgwidth,align=c]{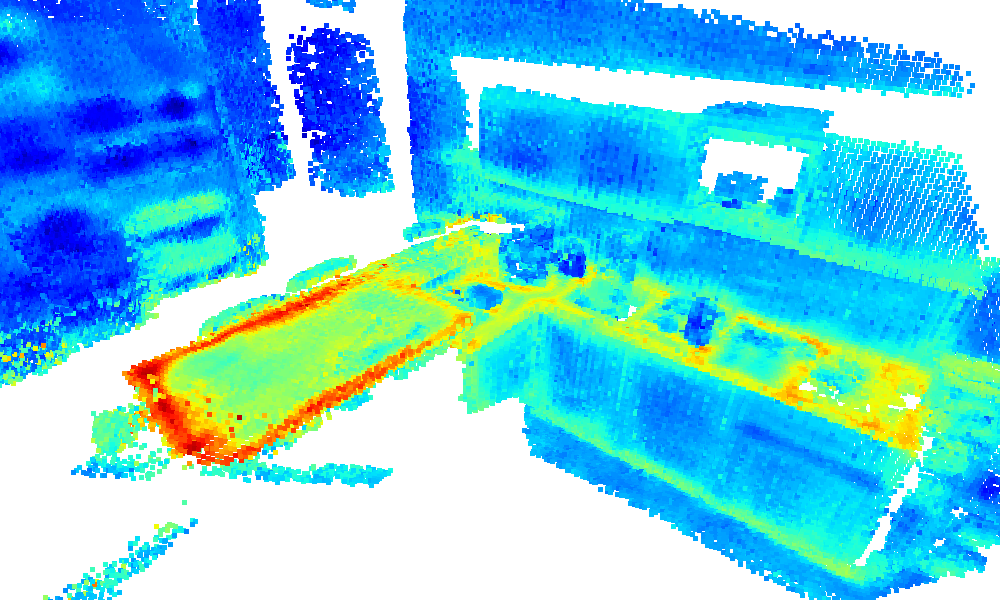}};
\end{tikzpicture}}
&
\adjustbox{valign=c}{\begin{tikzpicture}
    \node[anchor=south west,inner sep=0] at (0,0) {\includegraphics[width=\imgwidth,align=c]{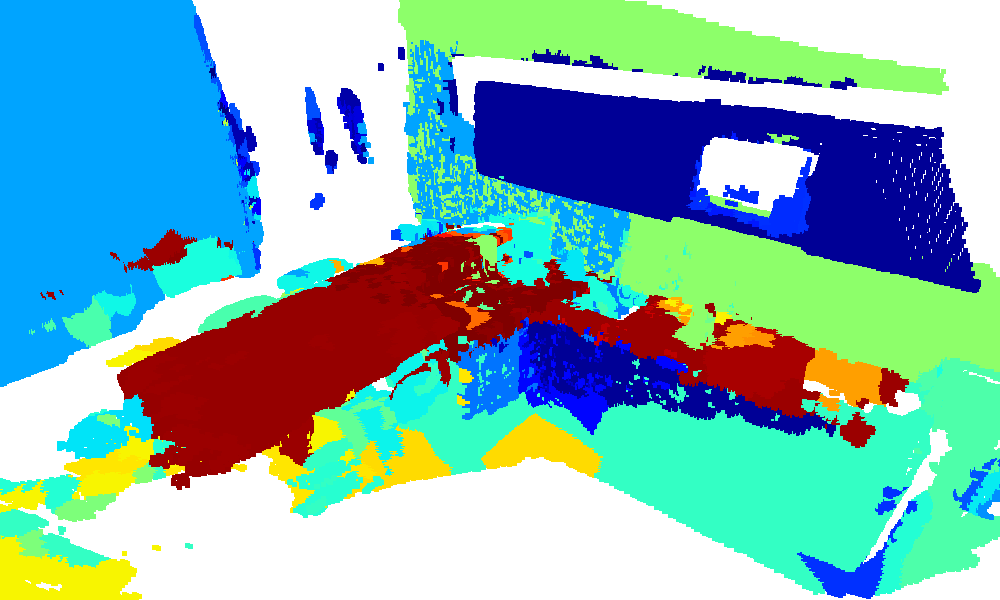}};
\end{tikzpicture}}
&
\adjustbox{valign=c}{\begin{tikzpicture}
    \node[anchor=south west,inner sep=0] at (0,0) {\includegraphics[width=\imgwidth,align=c]{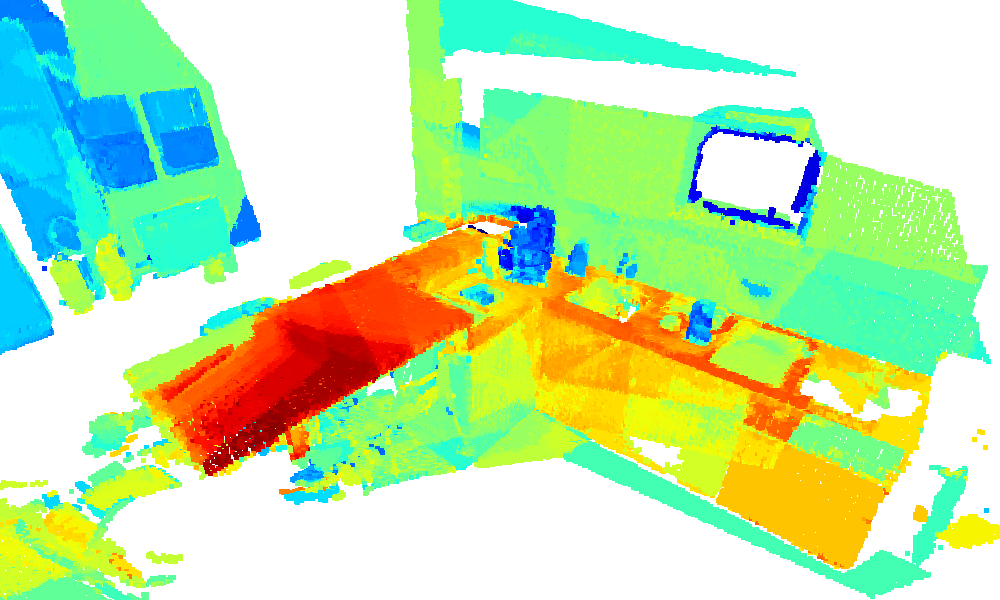}};
\end{tikzpicture}}
\\

% \rotatebox[origin=c]{90}{\textit{towel}} &
% \adjustbox{valign=c}{\begin{tikzpicture}
%     \node[anchor=south west,inner sep=0] at (0,0) {\includegraphics[width=\imgwidth,align=c]{results/3d/cf/render_towel.png}};
% \end{tikzpicture}}
% &
% \adjustbox{valign=c}{\begin{tikzpicture}
%     \node[anchor=south west,inner sep=0] at (0,0) {\includegraphics[width=\imgwidth,align=c]{results/3d/ls/ls_lvl1/render_towel.png}};
% \end{tikzpicture}}
% &
% \adjustbox{valign=c}{\begin{tikzpicture}
%     \node[anchor=south west,inner sep=0] at (0,0) {\includegraphics[width=\imgwidth,align=c]{results/3d/rf_vitb_2cm/render_towel.png}};
% \end{tikzpicture}}
% &
% \adjustbox{valign=c}{\begin{tikzpicture}
%     \node[anchor=south west,inner sep=0] at (0,0) {\includegraphics[width=\imgwidth,align=c]{results/3d/of/render_towel.png}};
% \end{tikzpicture}}
% &
% \adjustbox{valign=c}{\begin{tikzpicture}
%     \node[anchor=south west,inner sep=0] at (0,0) {\includegraphics[width=\imgwidth,align=c]{results/3d/xf/render_towel.png}};
% \end{tikzpicture}}
% \\

\bottomrule
\end{tabular}
} % makebox
\caption{3D environment reconstruction of the \textit{kitchen} sequence with similarity heatmap of text queries (blue: low, red: high). The queries \textit{coffee machine}, \textit{milk}, \textit{sink}, and \textit{socket} are highlighted with red and black bounding boxes, respectively. Due to the masked local embeddings, our method has much sharper query responses than ConceptFusion, and fewer outliers than LangSplat. Our method also has fewer pixelation and blurry artifacts in the dense colour model (first row).}
\label{fig:similarity_3d}
\vspace{-0.5cm}
\end{figure*}

We additionally provide qualitative comparisons of VLEM on custom RGB-D sequences against ConceptFusion~(CF), LangSplat~(LS), Open-Fusion~(OF), and RayFronts~(RF) in \Cref{fig:similarity_3d}.
% VLEM, Open-Fusion, and RayFronts run in real-time, while ConceptFusion and LangSplat process all frames offline. VLEM additionally runs pose estimation and colour mapping in parallel (\Cref{sec:localisation}). All other methods use pseudo ground truth poses provided by our SLAM system offline.
LangSplat process all frames offline, while all other methods run online in real-time. VLEM runs camera pose estimation in parallel using the remaining VRAM, while all other methods use pseudo ground truth poses provided offline by our SLAM system.

% We qualitatively compare the query response of VLEM with CLIP in the \textit{kitchen} sequence against ConceptFusion~(CF), LangSplat~(LS), Open-Fusion~(OF), and RayFronts~(RF) (\Cref{fig:similarity_3d}).

The integration of the colour images into our multivalued hash map exhibits fewer artifacts than the point cloud model used by ConceptFusion and also has fewer artifacts than LangSplat in areas close to unobserved space. Additionally, LangSplat does not provide a metric-accurate geometric representation as required for robotic applications. RayFronts' voxel-representation causes pixelation artifacts, while Open-Fusion shows reduced details.

\begin{figure}[htbp]
    \centering
    \begin{tikzpicture}
        \node[anchor=north west,inner sep=0] at (0,0) {\includegraphics[width=0.33\linewidth,frame]{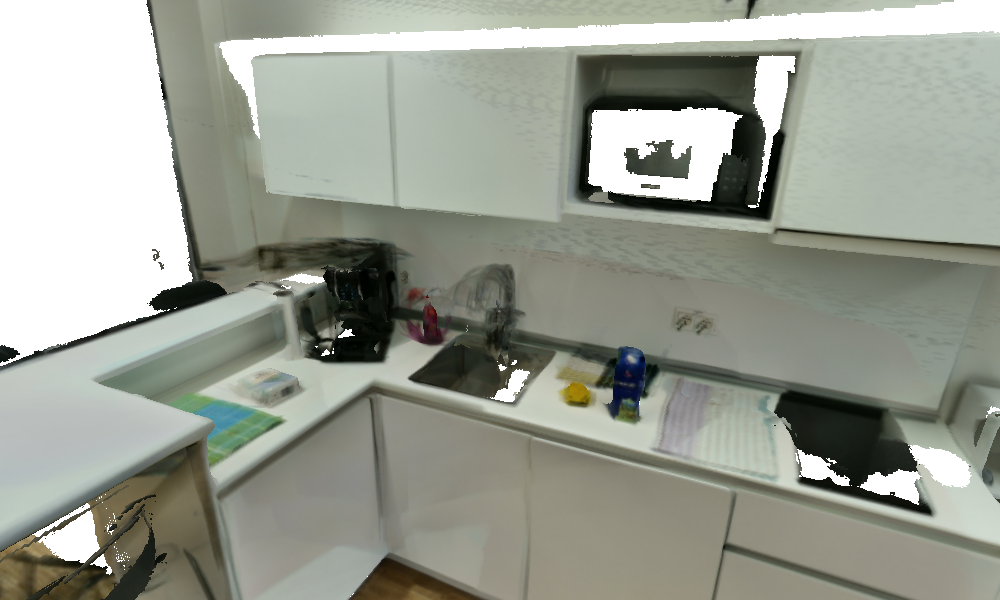}};
        \node[fill=white,above right] at (current bounding box.south west) {kitchen};
    \end{tikzpicture}%
    \begin{tikzpicture}
        \node[anchor=north west,inner sep=0] at (0,0) {\includegraphics[width=0.33\linewidth,frame]{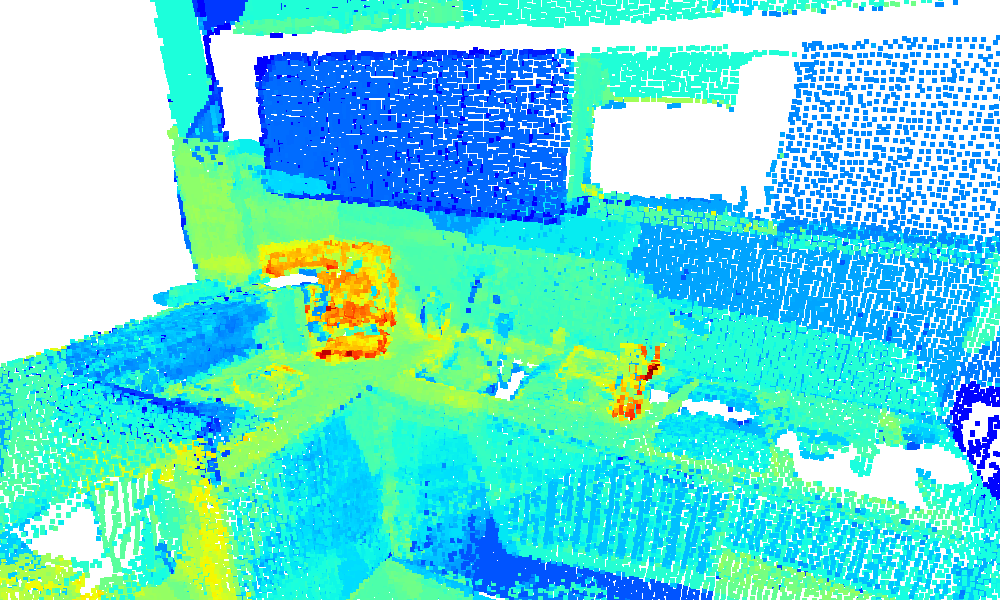}};
        \node[fill=white,above right] at (current bounding box.south west) {\textit{coffee}};
    \end{tikzpicture}%
    \begin{tikzpicture}
        \node[anchor=north west,inner sep=0] at (0,0) {\includegraphics[width=0.33\linewidth,frame]{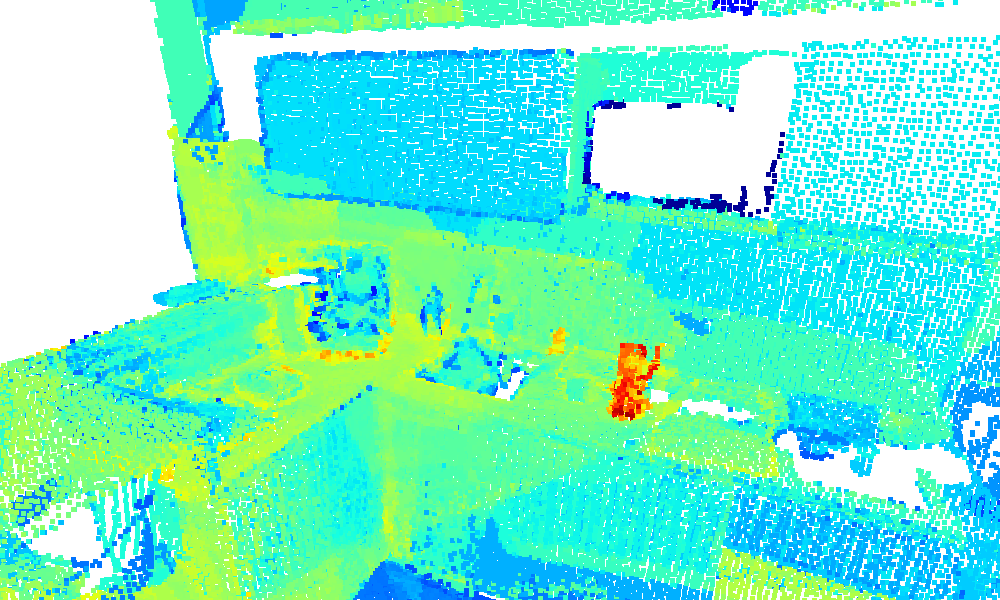}};
        \node[fill=white,above right] at (current bounding box.south west) {\textit{milk}};
    \end{tikzpicture}%
    
    \begin{tikzpicture}
        \node[anchor=north west,inner sep=0] at (0,0) {\includegraphics[width=0.33\linewidth,frame]{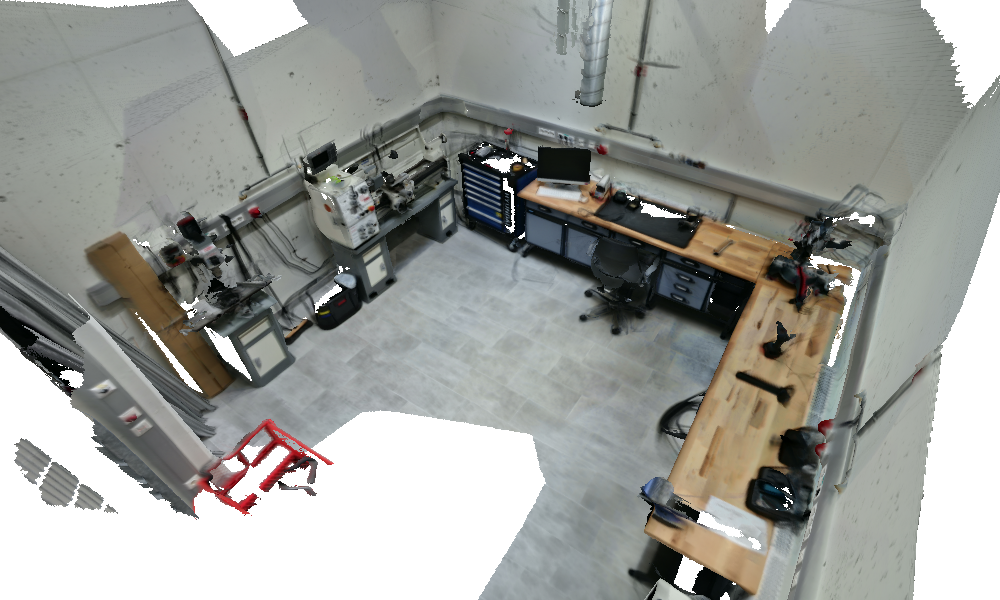}};
        \node[fill=white,above right] at (current bounding box.south west) {workshop};
    \end{tikzpicture}%
    \begin{tikzpicture}
        \node[anchor=north west,inner sep=0] at (0,0) {\includegraphics[width=0.33\linewidth,frame]{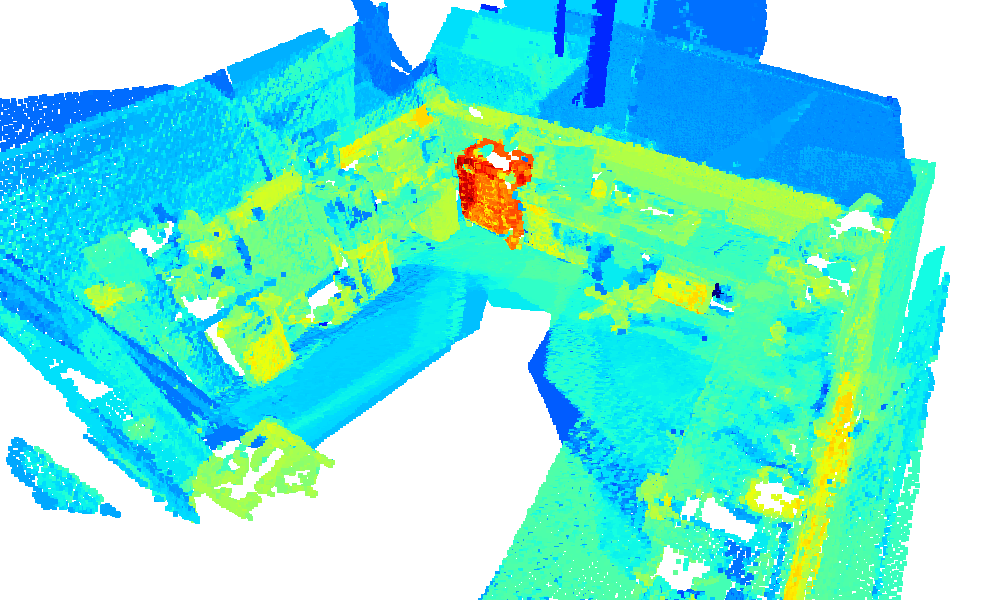}};
        \node[fill=white,above right] at (current bounding box.south west) {\textit{toolbox}};
    \end{tikzpicture}%
    \begin{tikzpicture}
        \node[anchor=north west,inner sep=0] at (0,0) {\includegraphics[width=0.33\linewidth,frame]{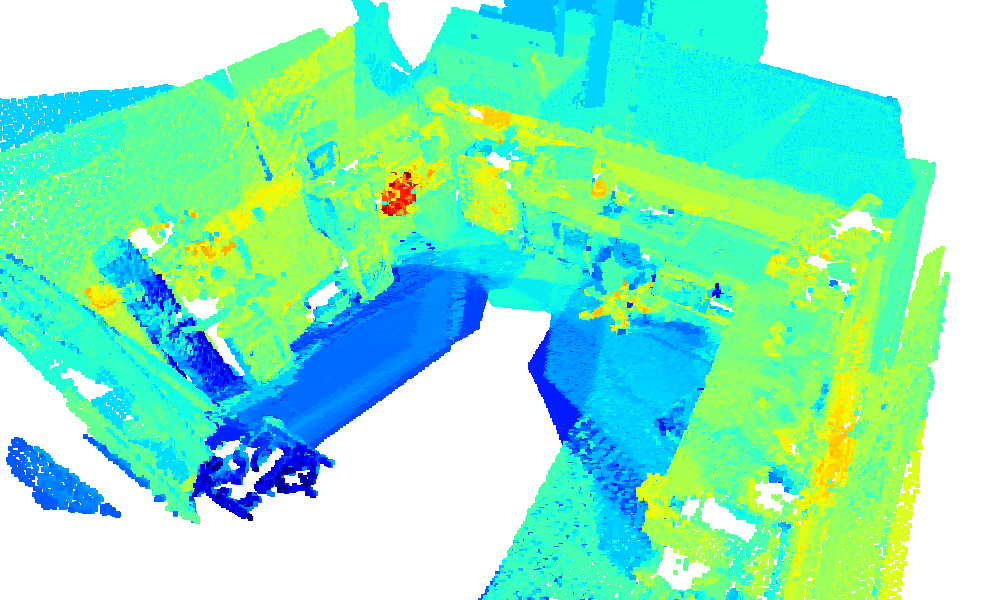}};
        \node[fill=white,above right] at (current bounding box.south west) {\textit{lathe}};
    \end{tikzpicture}%
    
    \begin{tikzpicture}
        \node[anchor=north west,inner sep=0] at (0,0) {\includegraphics[width=0.33\linewidth,frame]{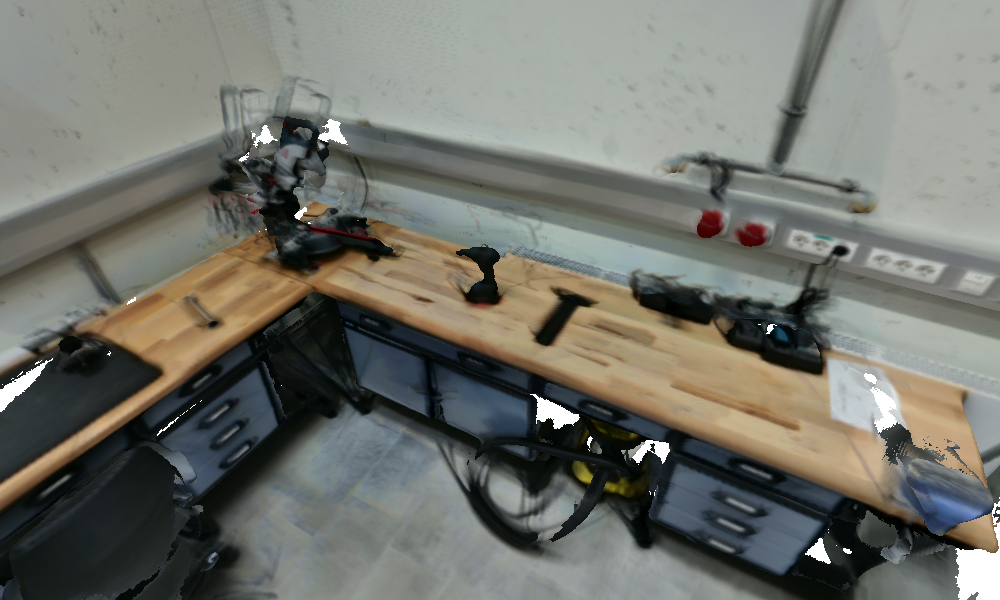}};
        \node[fill=white,above right] at (current bounding box.south west) {workshop};
    \end{tikzpicture}%
    \begin{tikzpicture}
        \node[anchor=north west,inner sep=0] at (0,0) {\includegraphics[width=0.33\linewidth,frame]{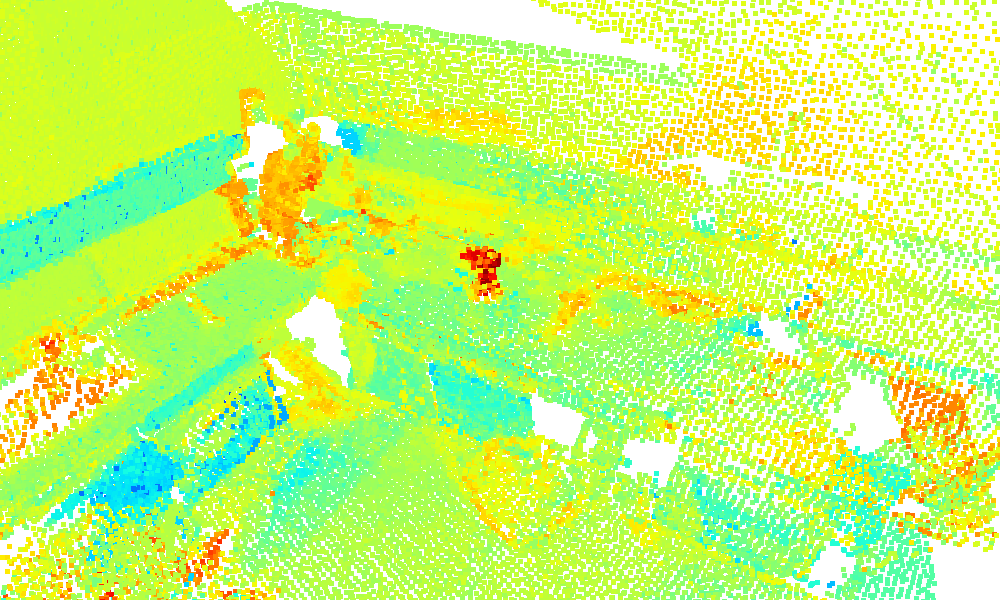}};
        \node[fill=white,above right] at (current bounding box.south west) {\textit{drill}};
    \end{tikzpicture}%
    \begin{tikzpicture}
        \node[anchor=north west,inner sep=0] at (0,0) {\includegraphics[width=0.33\linewidth,frame]{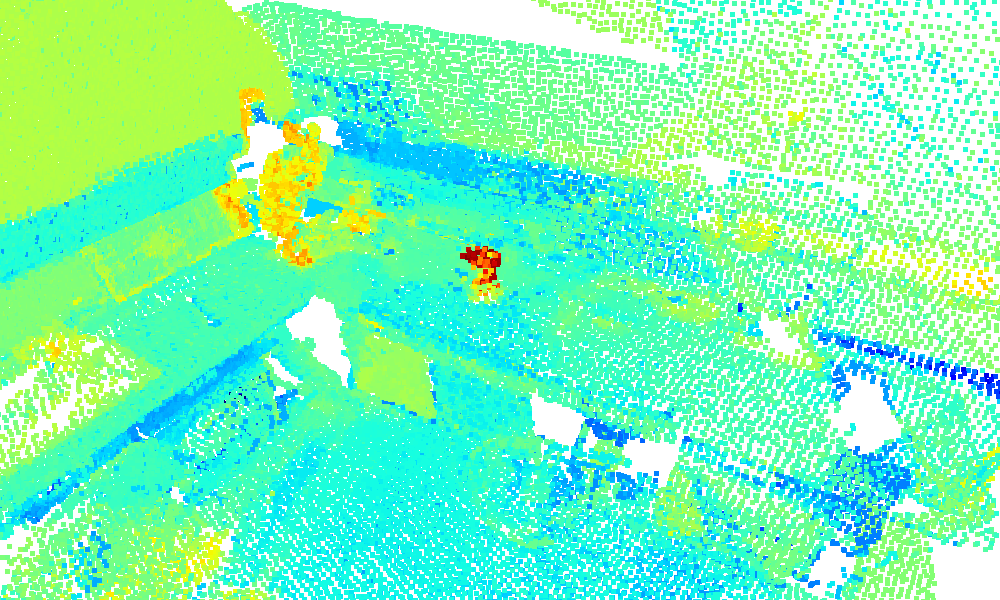}};
        \node[fill=white,above right] at (current bounding box.south west) {\textit{Bosch drill}};
    \end{tikzpicture}%
    
    \begin{tikzpicture}
        \node[anchor=north west,inner sep=0] at (0,0) {\includegraphics[width=0.33\linewidth,frame]{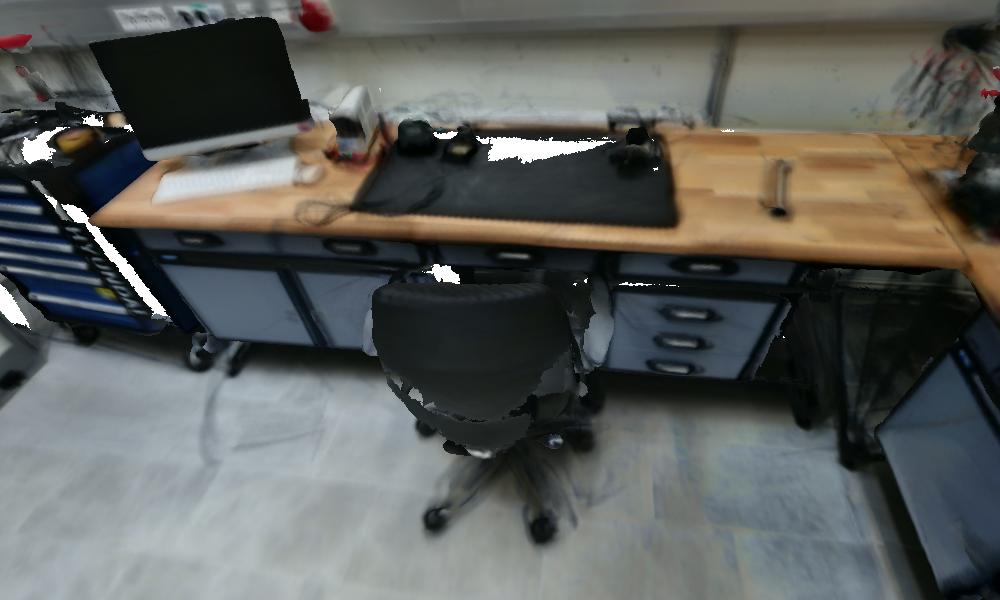}};
        \node[fill=white,above right] at (current bounding box.south west) {workshop};
    \end{tikzpicture}%
    \begin{tikzpicture}
        \node[anchor=north west,inner sep=0] at (0,0) {\includegraphics[width=0.33\linewidth,frame]{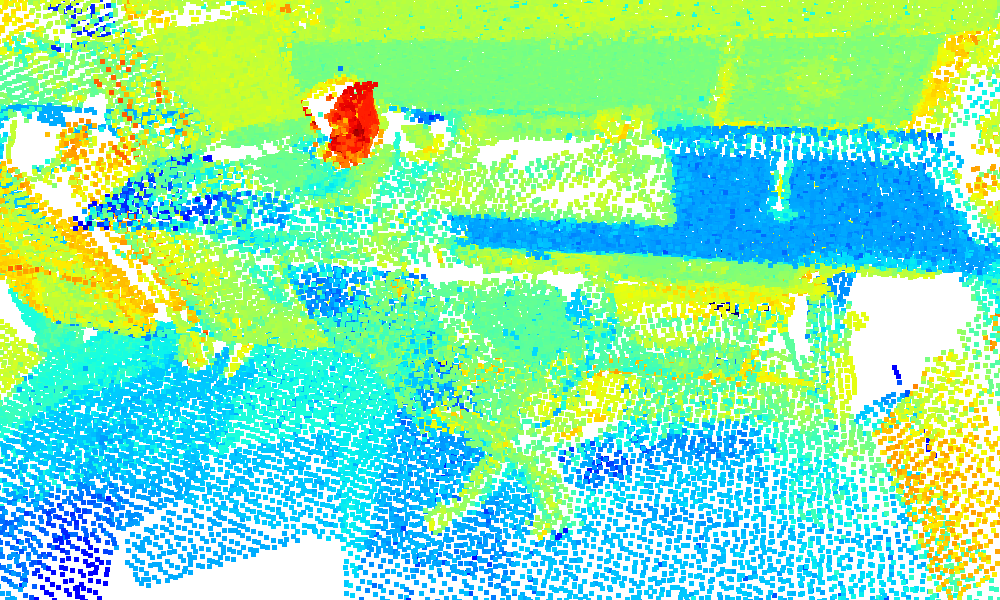}};
        \node[fill=white,above right] at (current bounding box.south west) {\textit{power supply}};
    \end{tikzpicture}%
    \begin{tikzpicture}
        \node[anchor=north west,inner sep=0] at (0,0) {\includegraphics[width=0.33\linewidth,frame]{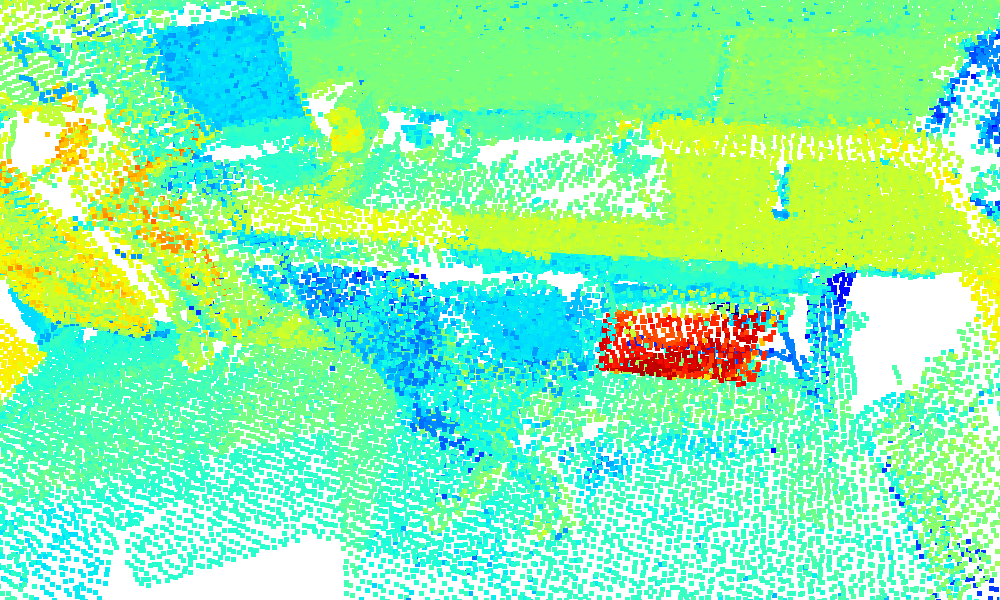}};
        \node[fill=white,above right] at (current bounding box.south west) {\textit{drawer}};
    \end{tikzpicture}%

    \begin{tikzpicture}
        \node[anchor=north west,inner sep=0] at (0,0) {\includegraphics[width=0.33\linewidth,frame]{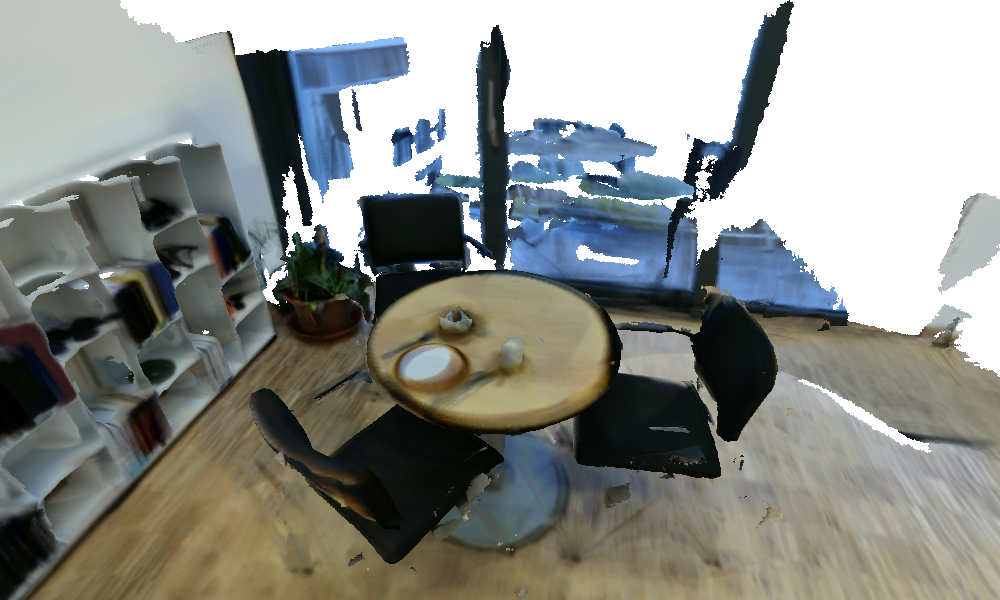}};
        \node[fill=white,above right] at (current bounding box.south west) {table};
    \end{tikzpicture}%
    \begin{tikzpicture}
        \node[anchor=north west,inner sep=0] at (0,0) {\includegraphics[width=0.33\linewidth,frame]{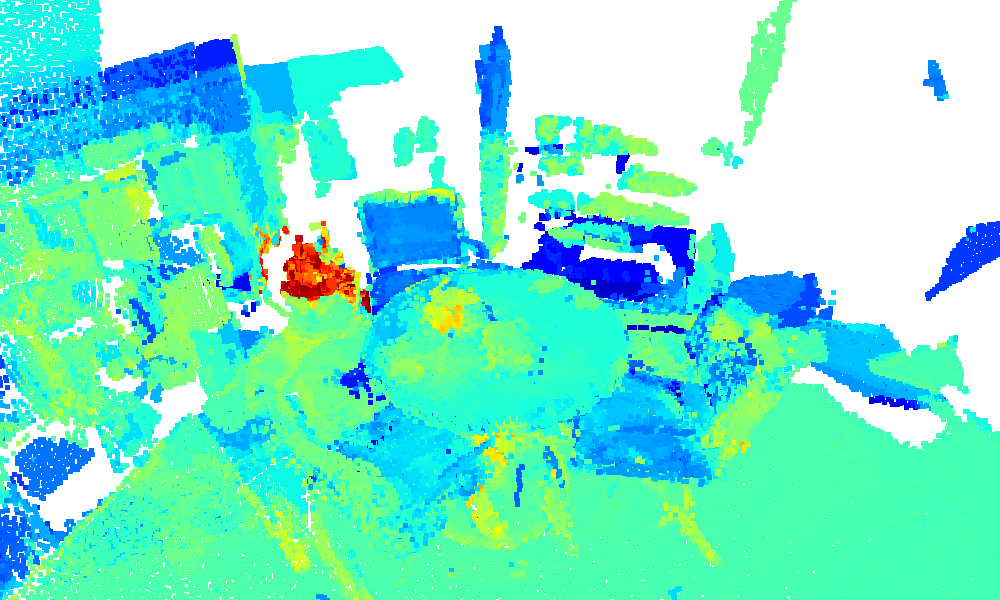}};
        \node[fill=white,above right] at (current bounding box.south west) {\textit{plant}};
    \end{tikzpicture}%
    \begin{tikzpicture}
        \node[anchor=north west,inner sep=0] at (0,0) {\includegraphics[width=0.33\linewidth,frame]{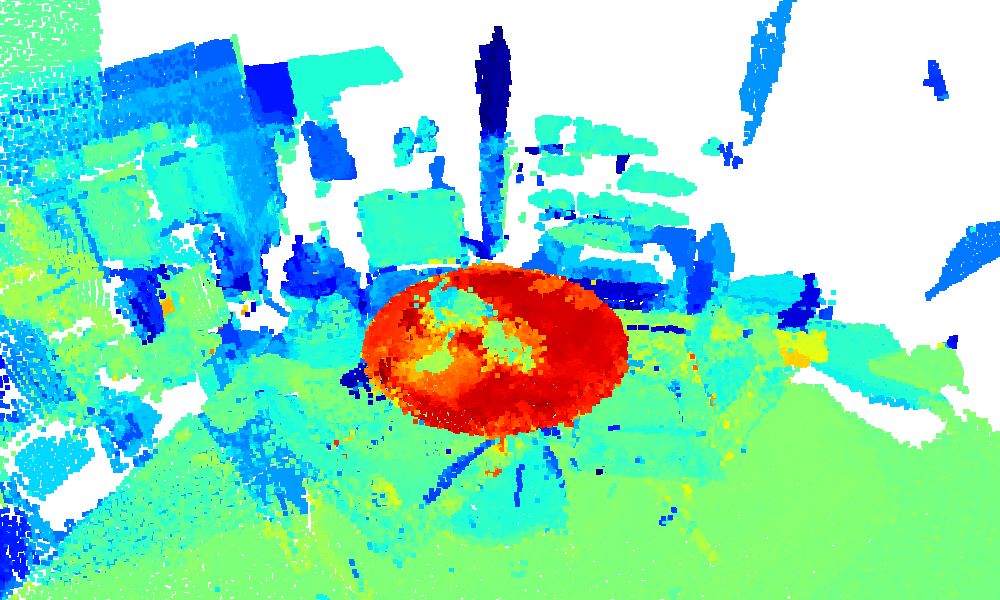}};
        \node[fill=white,above right] at (current bounding box.south west) {\textit{table}};
    \end{tikzpicture}%

    %\caption{Qualitative examples of reconstructions and similarity heatmaps are presented for the sequences \textit{kitchen}, \textit{workshop}, and \textit{table}. The specificity of the query notably affects the shape of the heatmap responses. For instance, a generic query for \textit{coffee} in the \textit{kitchen} sequence highlights multiple relevant objects including the coffee machine and milk carton, both essential to preparing coffee, whereas querying specifically for \textit{milk} emphasises only the milk carton (first row). Similarly, in the \textit{workshop}, querying precisely for the \textit{Bosch drill} yields a more pronounced response in the heatmap compared to a generic \textit{drill} query (third row). This adaptiveness highlights the open-set property of our method, enabling interaction with varying environments at varying semantic levels.}
    \caption{Qualitative results for VLEM reconstructions and similarity heatmaps for sequences \textit{kitchen} (68.2s), \textit{workshop} (121.31s), and \textit{table} (51.14s). The specificity of the query notably affects the shape of the heatmap responses.}
    % For instance, a generic query for \textit{coffee} in the \textit{kitchen} sequence highlights multiple relevant objects including the coffee machine and milk carton, whereas querying specifically for \textit{milk} emphasises only the milk carton (first row). Similarly, querying precisely for the \textit{Bosch drill} yields a more pronounced response in the heatmap compared to a generic \textit{drill} query (third row).
    \label{fig:qualitative_examples}
    \vspace{-0.5cm}
\end{figure}

The qualitative results in \Cref{fig:qualitative_examples} demonstrate the versatility of our method and the effect of the query specificity on the shape of the heatmap responses. For instance, a generic query for \textit{coffee} in the \textit{kitchen} sequence highlights multiple relevant objects including the coffee machine and milk carton, whereas querying specifically for \textit{milk} emphasises only the milk carton (first row). Similarly, querying precisely for the \textit{Bosch drill} yields a more pronounced response in the heatmap compared to a generic \textit{drill} query (third row).

\begin{figure}
    \vspace{0.2cm}
    \centering

    \setlength{\tabcolsep}{0pt}
    \renewcommand{\arraystretch}{0}

    \begin{tabular}{cccc}

    \multicolumn{2}{c}{\scriptsize \textbf{RayFronts}} & \multicolumn{2}{c}{\scriptsize \textbf{VLEM~(CLIP)}} \\
    
    \begin{tikzpicture}
        \node[anchor=north west,inner sep=0] at (0,0) {\includegraphics[width=0.25\linewidth]{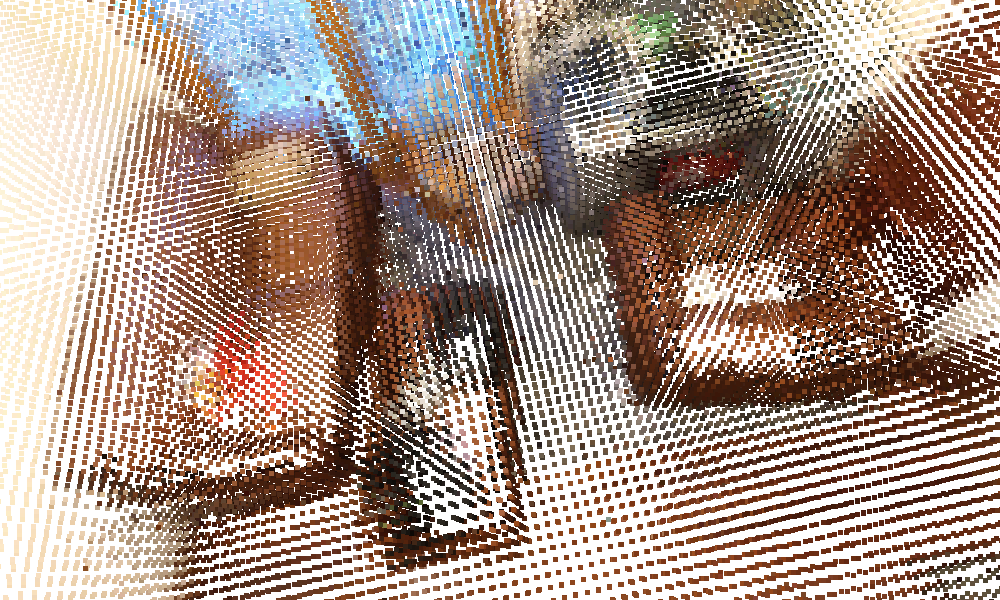}};%
        % \node[fill=white,above left] at (current bounding box.south east) {\scriptsize RF RGB};
    \end{tikzpicture}
    &
    \begin{tikzpicture}
        \node[anchor=north west,inner sep=0] at (0,0) {\includegraphics[width=0.25\linewidth]{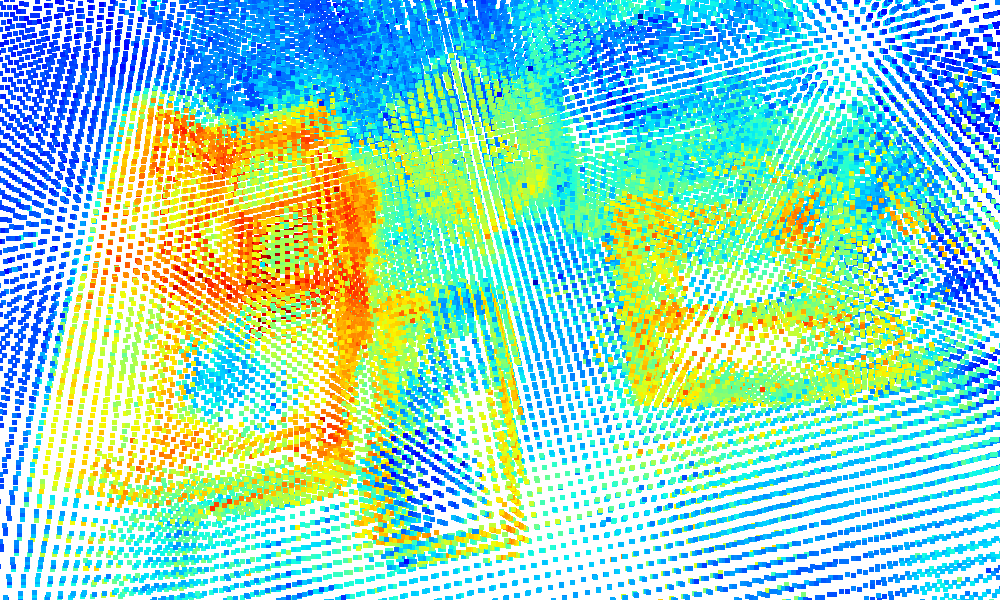}};%
        \node[fill=white,above left] at (current bounding box.south east) {\scriptsize \textit{sofa}};
    \end{tikzpicture}
    &
    \begin{tikzpicture}
        \node[anchor=north west,inner sep=0] at (0,0) {\includegraphics[width=0.25\linewidth]{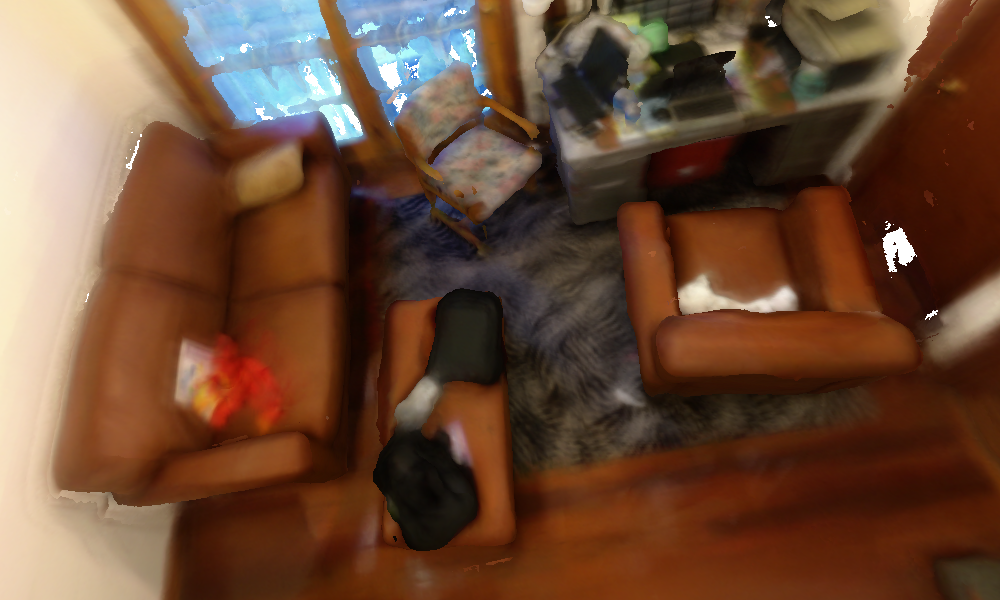}};%
        % \node[fill=white,above left] at (current bounding box.south east) {\scriptsize VLEM RGB};
    \end{tikzpicture}
    &
    \begin{tikzpicture}
        \node[anchor=north west,inner sep=0] at (0,0) {\includegraphics[width=0.25\linewidth]{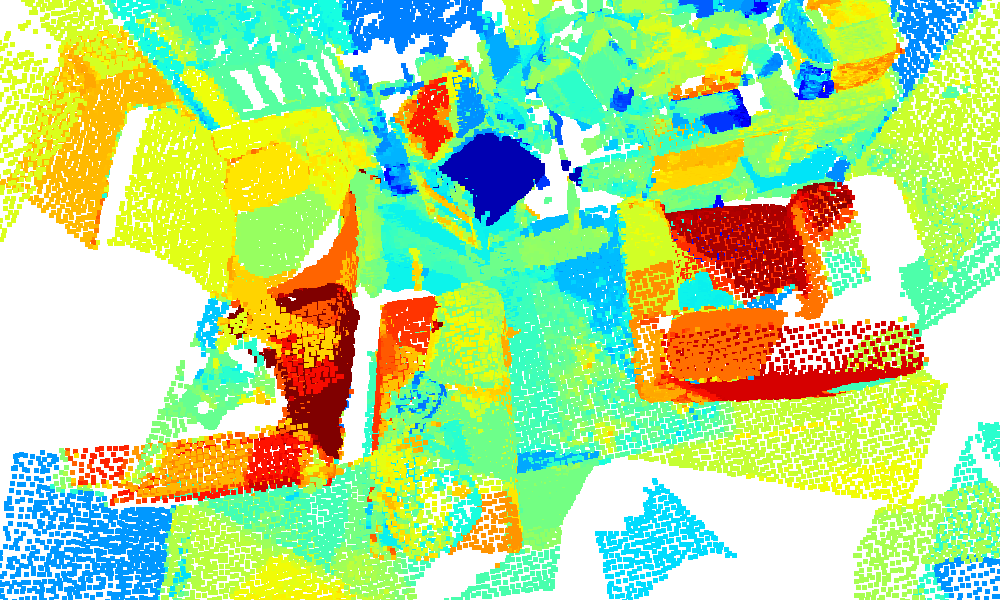}};%
        \node[fill=white,above left] at (current bounding box.south east) {\scriptsize \textit{sofa}};
    \end{tikzpicture}
    \\

    \begin{tikzpicture}
        \node[anchor=north west,inner sep=0] at (0,0) {\includegraphics[width=0.25\linewidth]{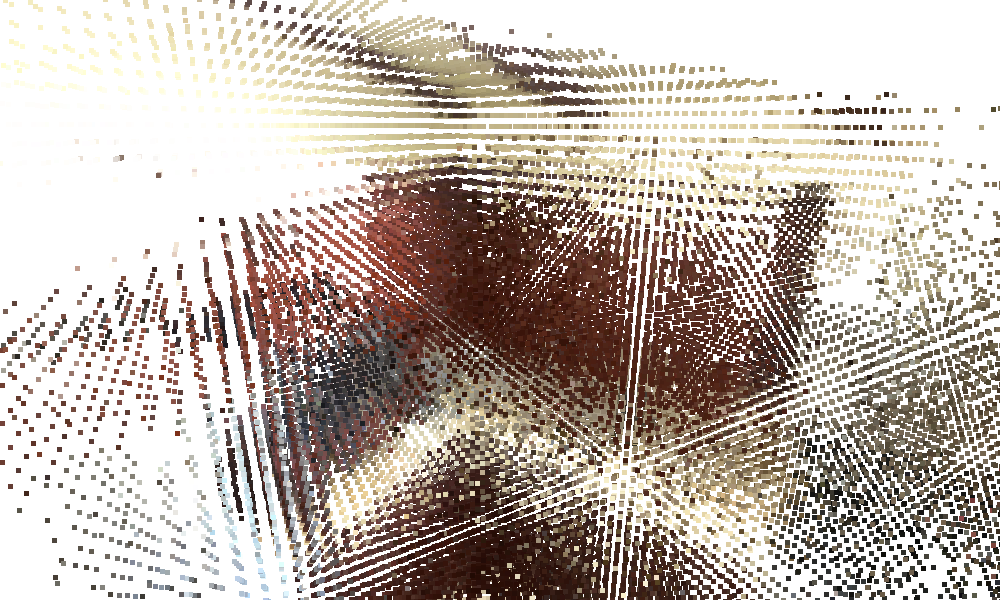}};%
        % \node[fill=white,below right] at (current bounding box.north west) {\scriptsize RF RGB};
    \end{tikzpicture}
    &
    \begin{tikzpicture}
        \node[anchor=north west,inner sep=0] at (0,0) {\includegraphics[width=0.25\linewidth]{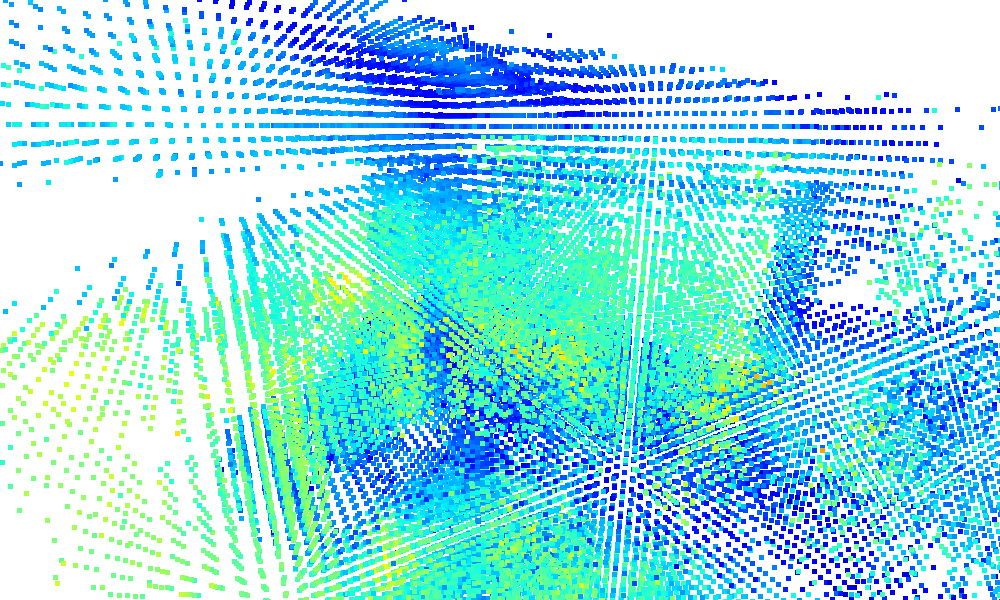}};%
        \node[fill=white,below right] at (current bounding box.north west) {\scriptsize \textit{door}};
    \end{tikzpicture}
    &
    \begin{tikzpicture}
        \node[anchor=north west,inner sep=0] at (0,0) {\includegraphics[width=0.25\linewidth]{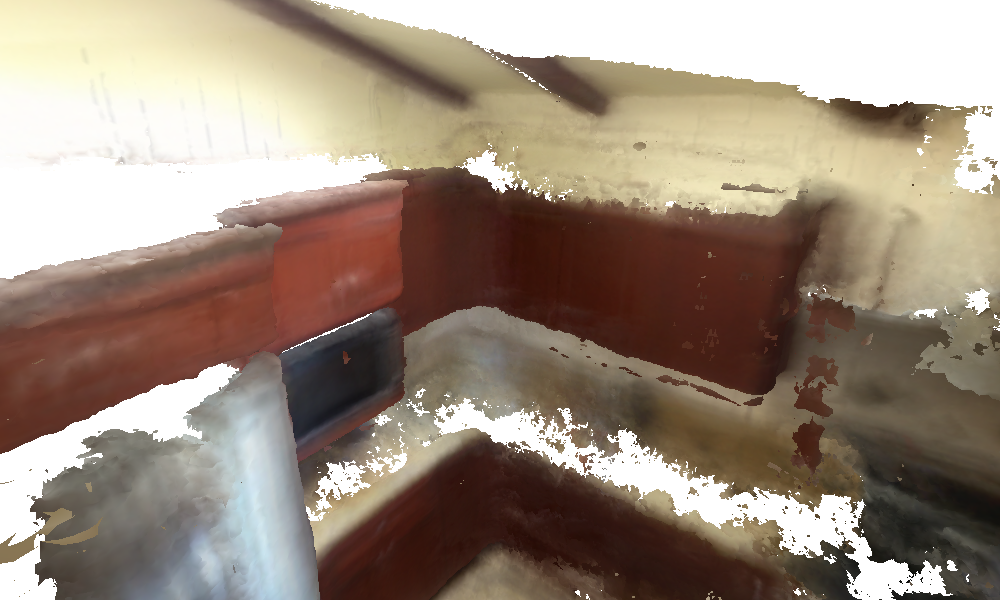}};%
        % \node[fill=white,below right] at (current bounding box.north west) {\scriptsize VLEM RGB};
    \end{tikzpicture}
    &
    \begin{tikzpicture}
        \node[anchor=north west,inner sep=0] at (0,0) {\includegraphics[width=0.25\linewidth]{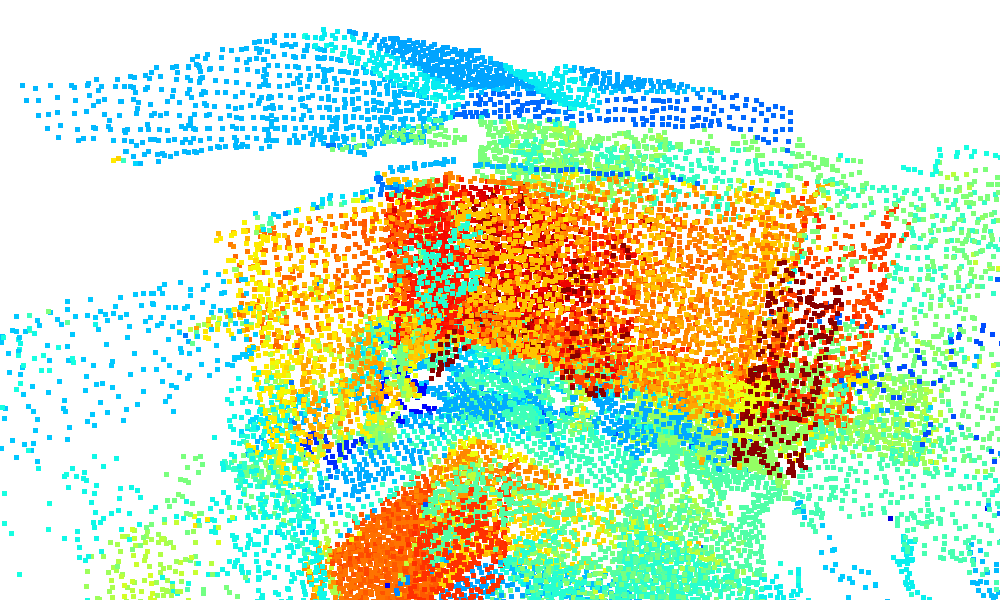}};%
        \node[fill=white,below right] at (current bounding box.north west) {\scriptsize \textit{door}};
    \end{tikzpicture}%
    \\
    
    \begin{tikzpicture}
        \node[anchor=north west,inner sep=0] at (0,0) {\includegraphics[width=0.25\linewidth]{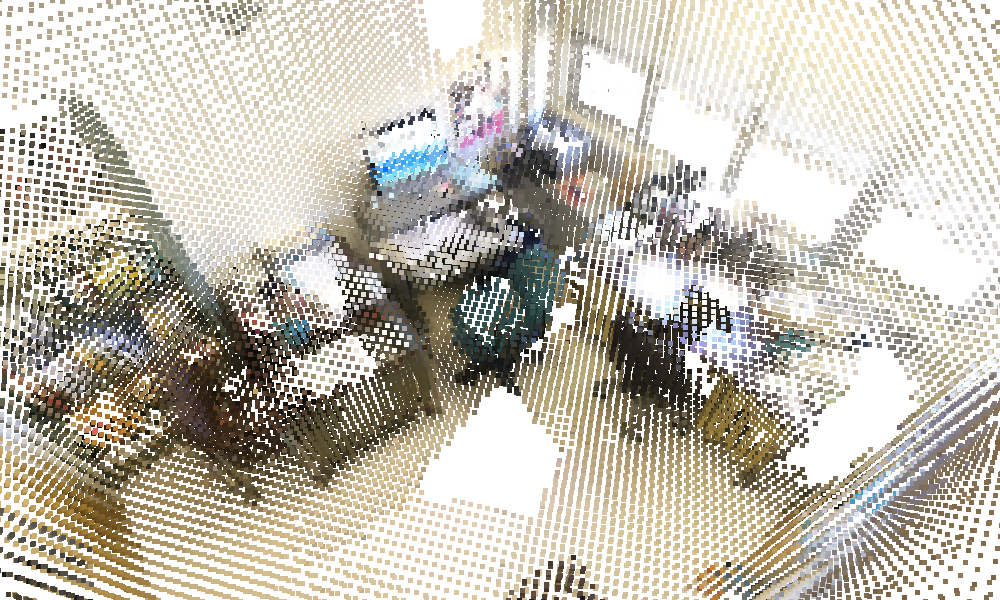}};%
        % \node[fill=white,below right] at (current bounding box.north west) {\scriptsize RF RGB};
    \end{tikzpicture}
    &
    \begin{tikzpicture}
        \node[anchor=north west,inner sep=0] at (0,0) {\includegraphics[width=0.25\linewidth]{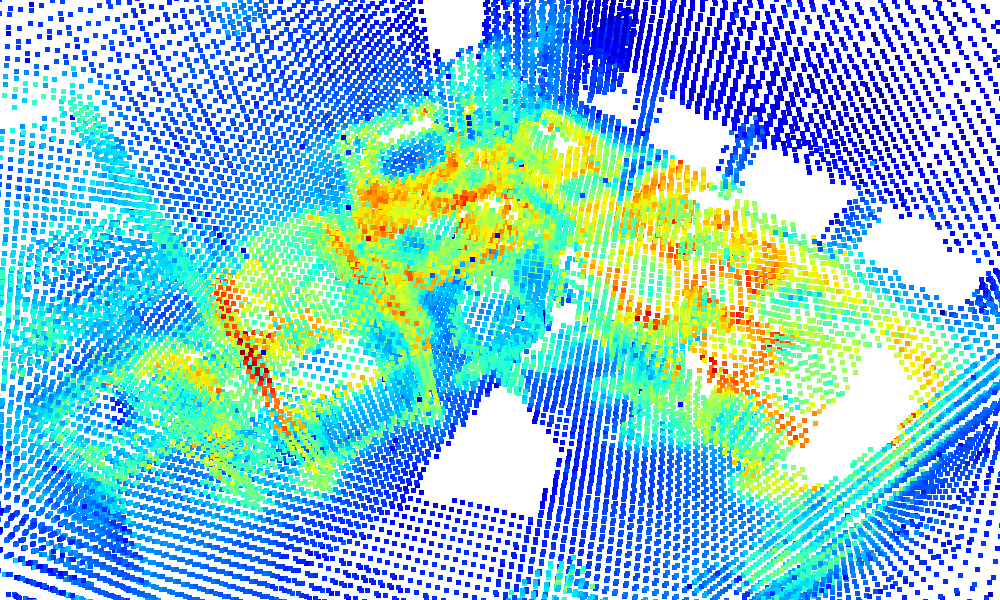}};%
        \node[fill=white,below right] at (current bounding box.north west) {\scriptsize \textit{desk}};
    \end{tikzpicture}
    &
    \begin{tikzpicture}
        \node[anchor=north west,inner sep=0] at (0,0) {\includegraphics[width=0.25\linewidth]{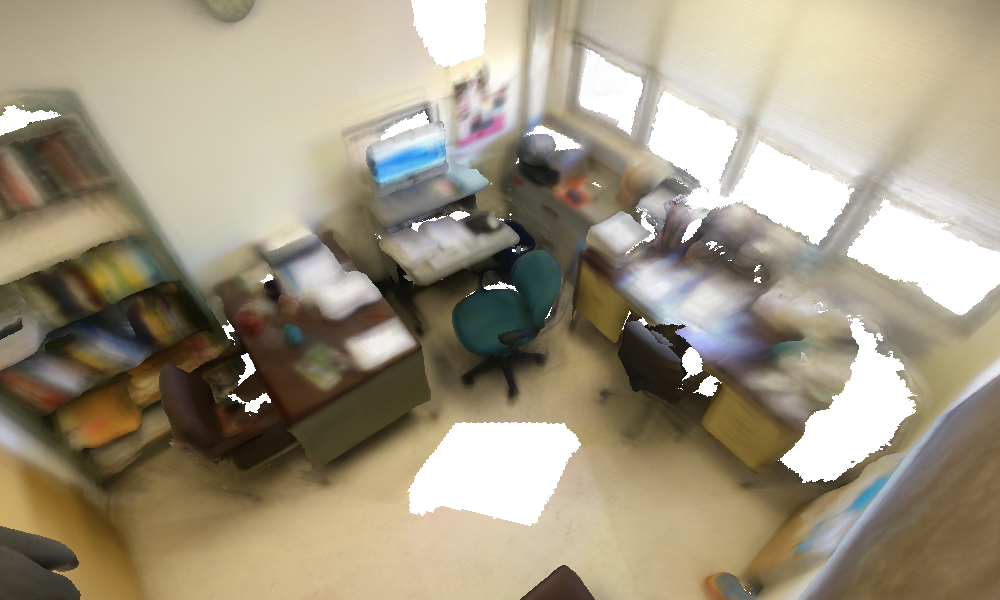}};%
        % \node[fill=white,below right] at (current bounding box.north west) {\scriptsize VLEM RGB};
    \end{tikzpicture}
    &
    \begin{tikzpicture}
        \node[anchor=north west,inner sep=0] at (0,0) {\includegraphics[width=0.25\linewidth]{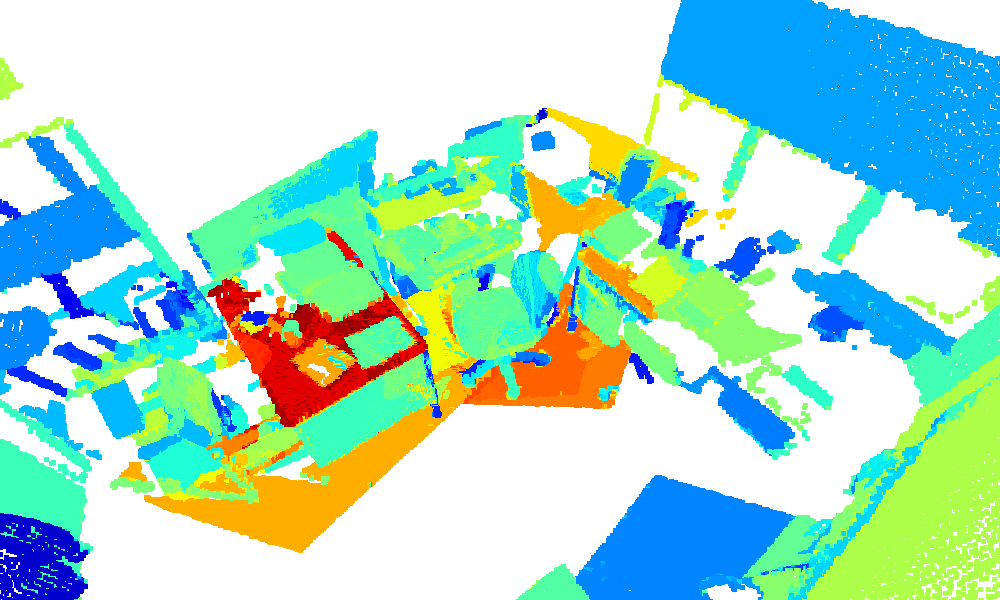}};%
        \node[fill=white,below right] at (current bounding box.north west) {\scriptsize \textit{desk}};
    \end{tikzpicture}%
    \\

    \end{tabular}

    \caption{ScanNet scene (50, 231, 378) reconstruction and similarity for patch- (left) and segment-embeddings (right).}
    \label{fig:scannet_qualitative}
    \vspace{-0.5cm}
\end{figure}

Although novel VLMs with semantic meaningful patch-embeddings provide higher segmentation performance, \Cref{fig:scannet_qualitative} shows that this stems from covering multiple instances, while segment-embeddings provide a more consistent single instance segmentation.

\subsection{Robotic Applications}

% This section demonstrates the application of our approach to various robotic applications. See the supplementary video for details.

Our requirements on a consistent segmentation and spatial embedding resolution of arbitrarily queried objects excludes approaches such as RayFronts, which requires reducing the resolution beyond the size of objects we want to manipulate, and Open-Fusion, which generally exhibited inferior segmentation accuracy. We further run experiments in real-time on an NVIDIA GeForce RTX 4070 and are thus limited to 12GiB VRAM. Given these constraints, we selected VLEM~(CLIP) and simultaneously estimate the camera pose, reconstruct the colour map and merge embeddings in 3D. Once a user queries an object, we compute the similarity between points in the embedding map and the text embeddings, and segment the point cloud into point sets.

% Once a user queries the pick and place targets, we compute the similarity between points in the embedding map and the text embeddings, and segment the point cloud into point sets (\Cref{sec:segmentation}).

\subsubsection{Interactive Manipulation}

\begin{figure}[htbp]
    \centering
    \begin{tikzpicture}
        \node[anchor=north west,inner sep=0] at (0,0) {\includegraphics[width=\linewidth]{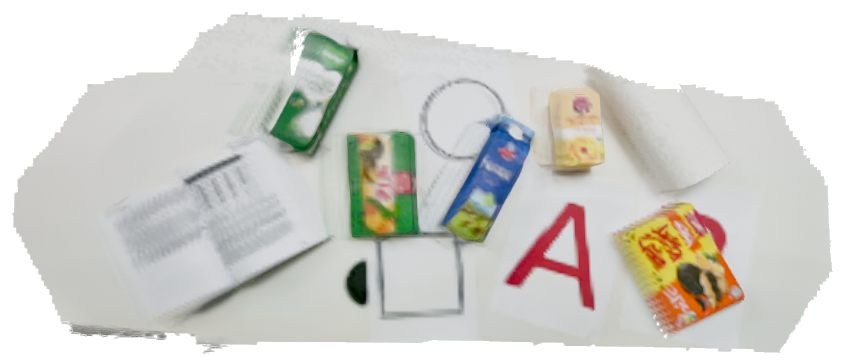}};
        \node[fill=white,below right] at (current bounding box.north west) {reconstruction};
    \end{tikzpicture}
    \begin{tikzpicture}
        \node[anchor=north west,inner sep=0] at (0,0) {\includegraphics[width=\linewidth]{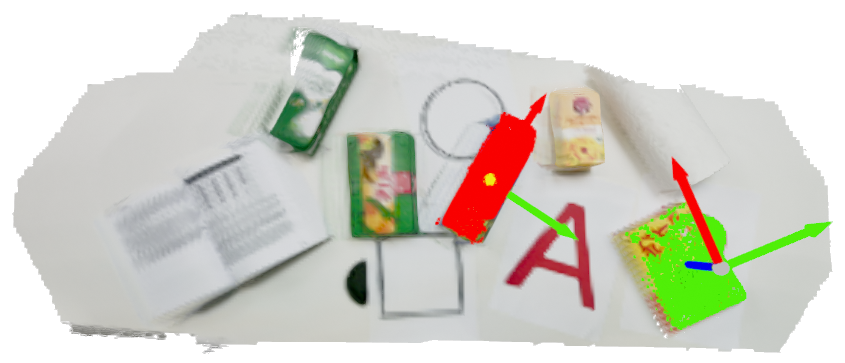}};
        \node[fill=white,below right] at (current bounding box.north west) {pick \textit{Formil milk}, place \textit{jaffa cake} (english)};
    \end{tikzpicture}
    \begin{tikzpicture}
        \node[anchor=north west,inner sep=0] at (0,0) {\includegraphics[width=\linewidth]{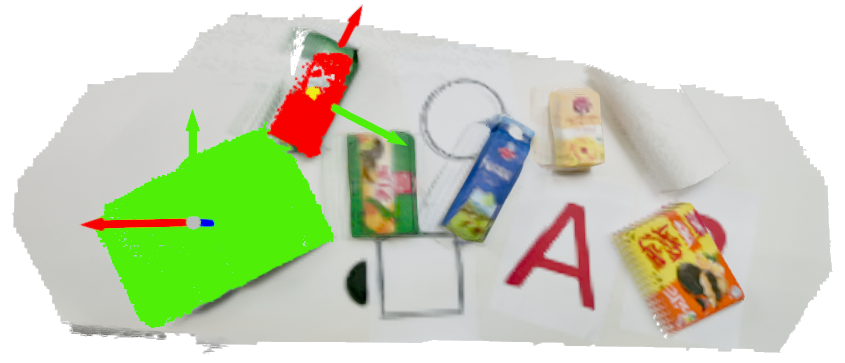}};
        \node[fill=white,below right] at (current bounding box.north west) {pick \textit{leche entera}, place \textit{libro} (spanish)};
    \end{tikzpicture}
    \caption{Reconstruction and end-effector poses (coordinate frames) for queried pick (red) and place (green) locations.}
    \label{fig:pick_place}
    \vspace{-0.5cm}
\end{figure}

\Cref{fig:pick_place} demonstrates the application of our method to an interactive pick and place task, where a user queries objects to pick and locations or objects where to place a grasped object.
The grasp pose for a parallel gripper is extracted by 1) projecting these 3D points on a plane orthogonal to an approach direction and located at a gripper-specific depth from the highest point, and 2) selecting the shortest extent of the projected point set.

\subsubsection{Mobile Platform}

\begin{figure}
     \centering
     \includegraphics[width=\linewidth]{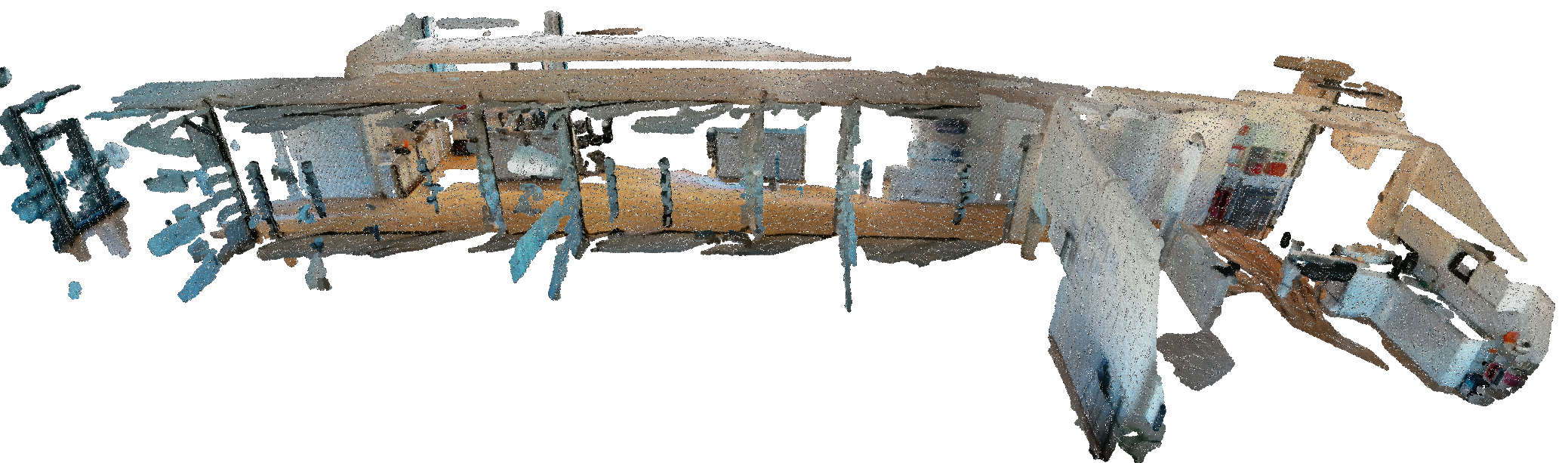}\\%
     \begin{tikzpicture}
        \node[anchor=north west,inner sep=0] at (0,0) {\includegraphics[width=\linewidth]{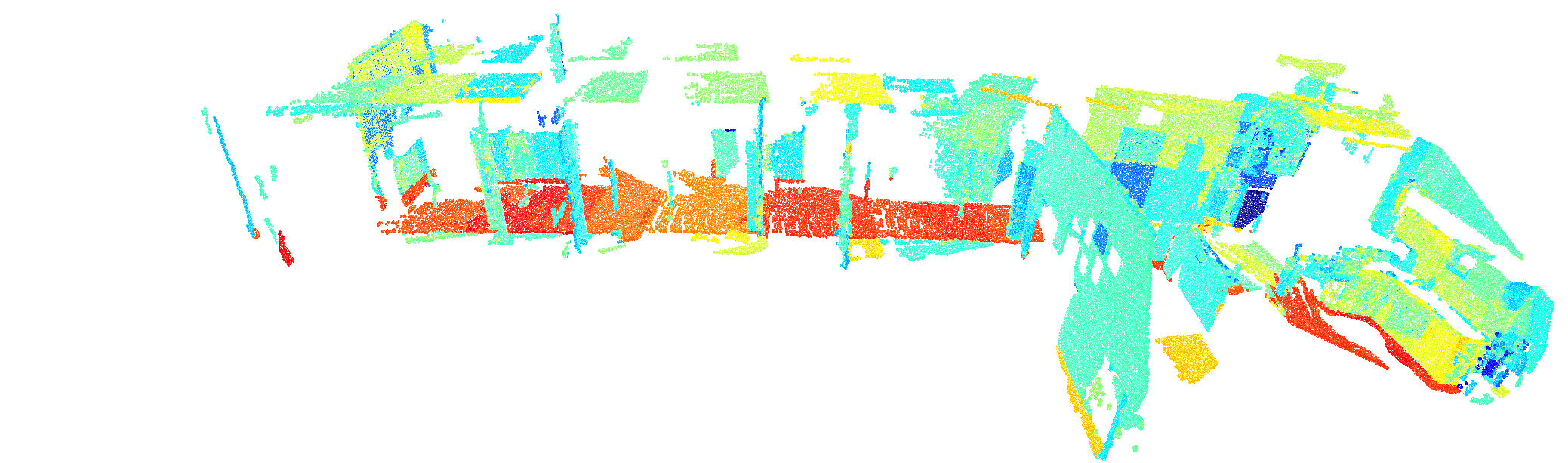}};
        \node[fill=white,below right] at (current bounding box.north west) {\textit{floor}};
    \end{tikzpicture}\\%
    % \begin{tikzpicture}
    %     \node[anchor=north west,inner sep=0] at (0,0) {\includegraphics[width=\linewidth]{results/navigation/floor3_sim_ceiling.png}};
    %     \node[fill=white,below right] at (current bounding box.north west) {\textit{ceiling}};
    % \end{tikzpicture}\\%
     
    \begin{tikzpicture}
        \node[anchor=north west,inner sep=0] at (0,0) {\includegraphics[width=0.5\linewidth,frame]{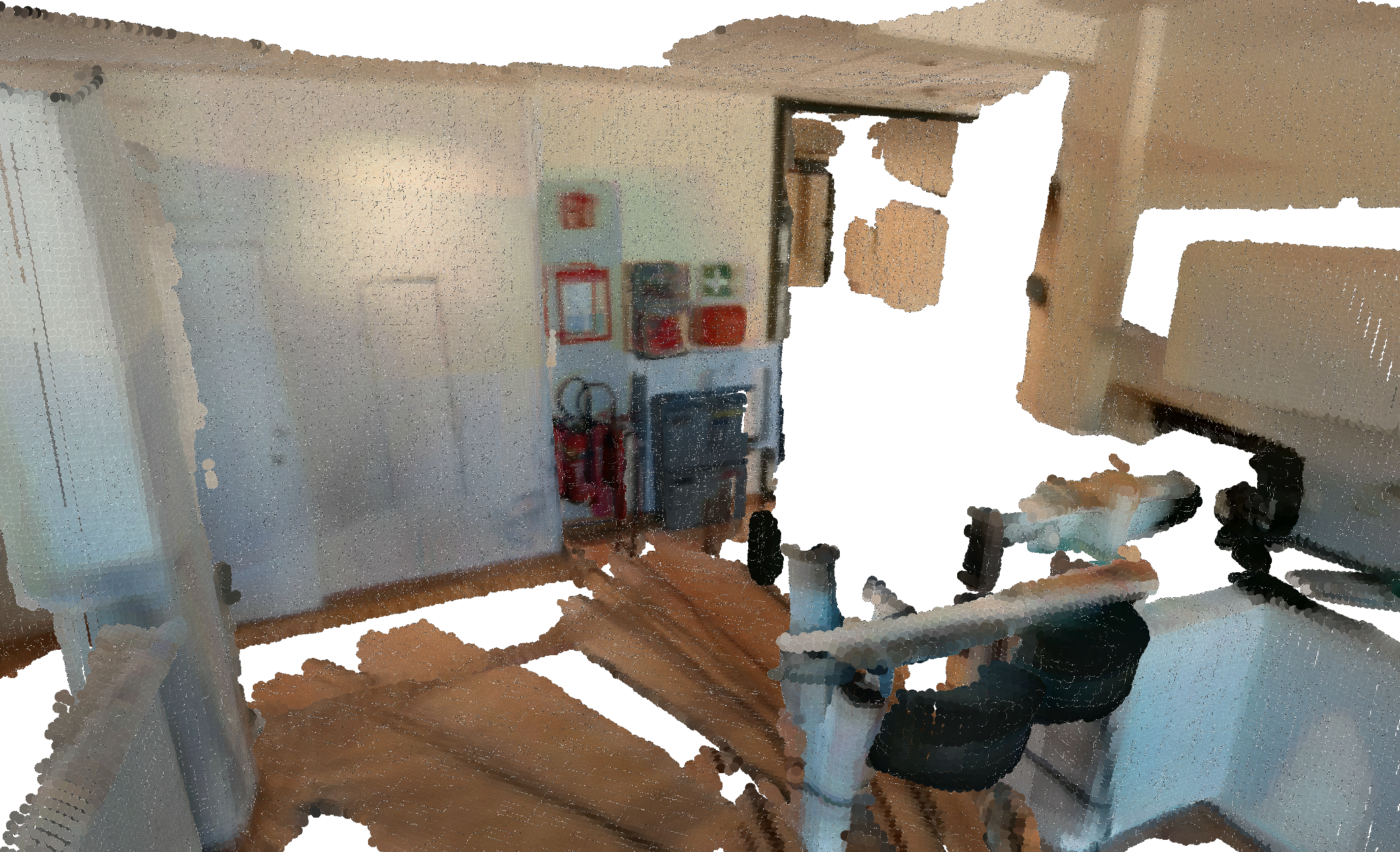}};
    \end{tikzpicture}%
    \begin{tikzpicture}
        \node[anchor=north west,inner sep=0] at (0,0) {\includegraphics[width=0.5\linewidth,frame]{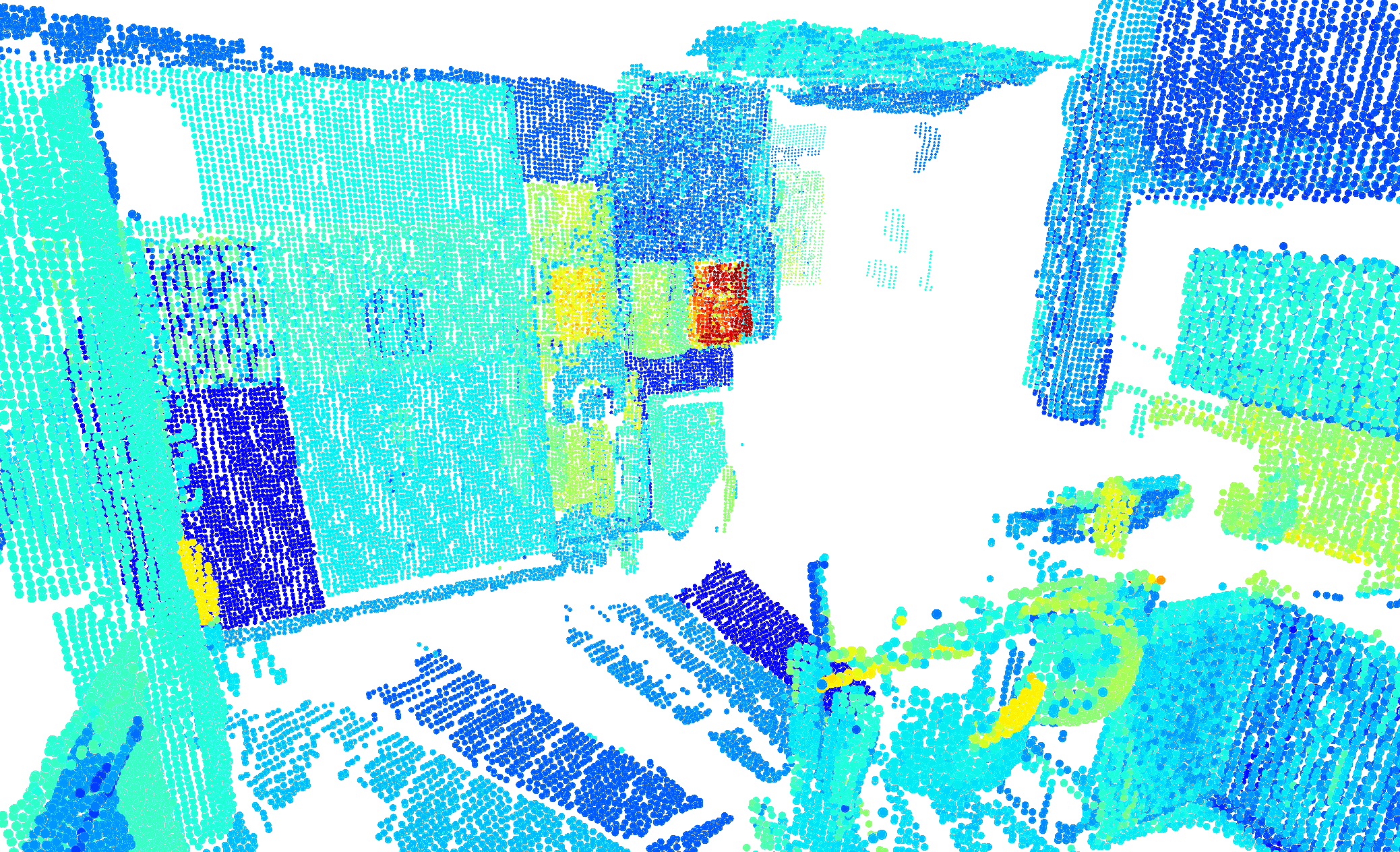}};
        \node[fill=white,below right] at (current bounding box.north west) {\textit{first aid kit}};
    \end{tikzpicture}%

    \begin{tikzpicture}
        \node[anchor=north west,inner sep=0] at (0,0) {\includegraphics[width=0.5\linewidth,frame]{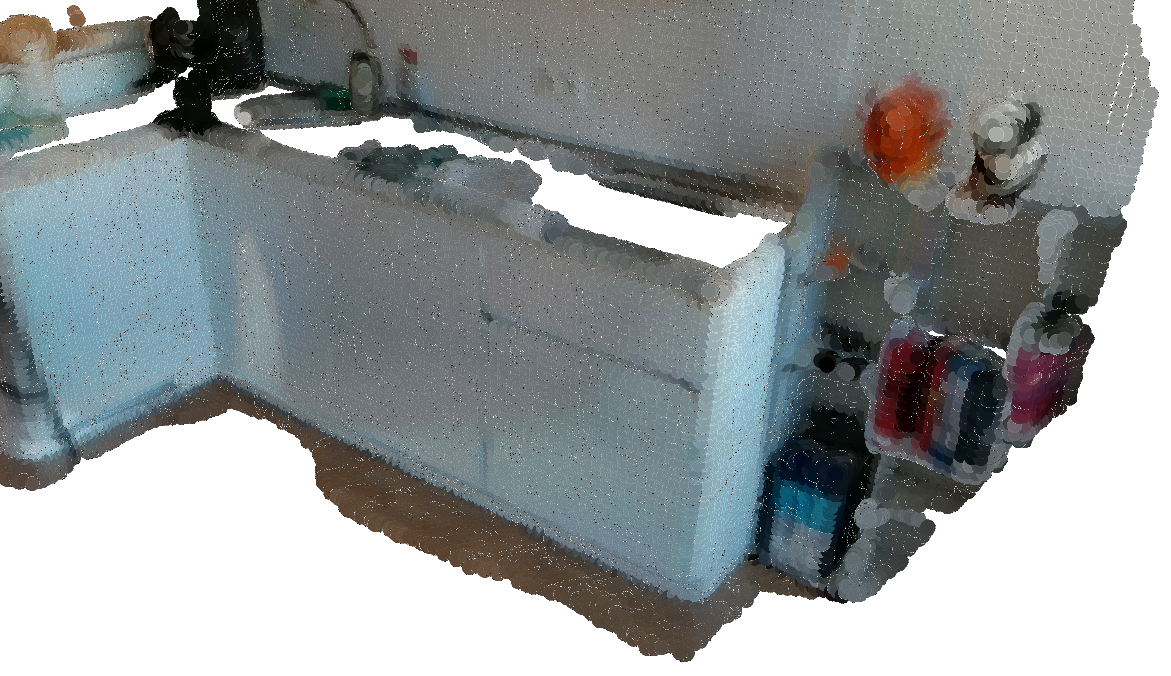}};
    \end{tikzpicture}%
    \begin{tikzpicture}
        \node[anchor=north west,inner sep=0] at (0,0) {\includegraphics[width=0.5\linewidth,frame]{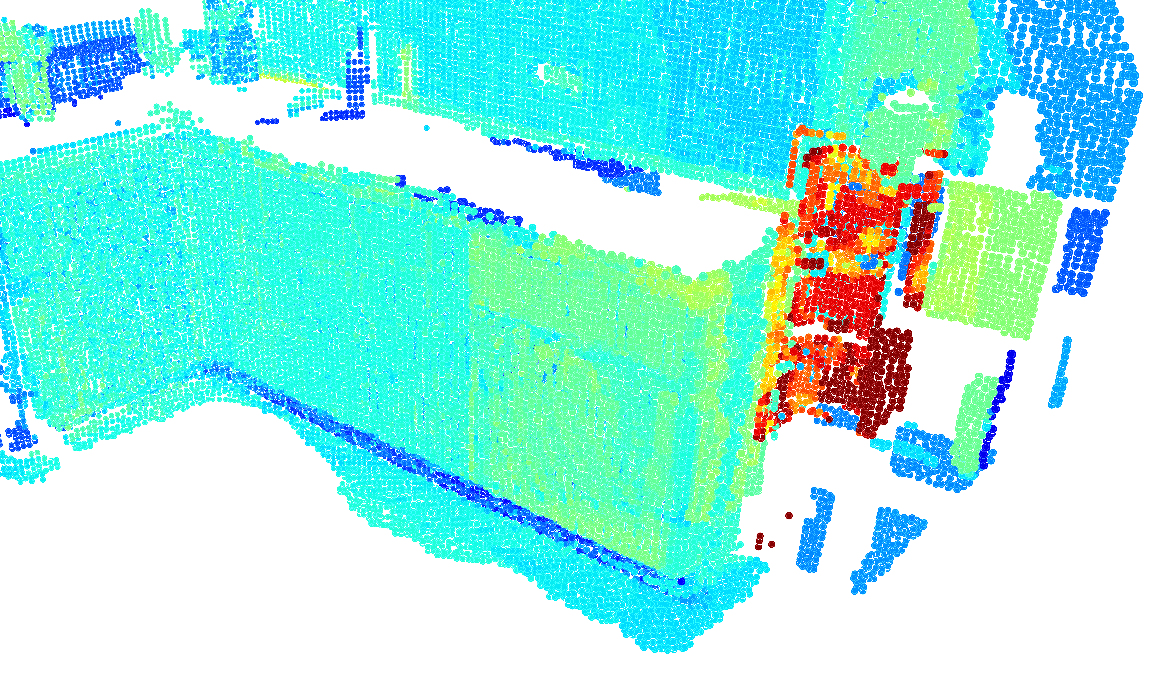}};
        \node[fill=white,below right] at (current bounding box.north west) {\textit{book shelf}};
    \end{tikzpicture}%
     
     \caption{Reconstruction of an office environment with multi-room- and object-level queries. In the same model, we are able to identify semantic concepts that span over multiple rooms (top rows, \textit{floor}) as well as individual objects (bottom rows, \textit{first aid kit} and \textit{book shelf}). Scenes are shown at different stages of the live reconstruction.}
     \label{fig:mobile_robot}
     \vspace{-0.5cm}
\end{figure}

While local manipulation tasks require high accuracy only on an object-level, mobile manipulation tasks have to trade off room- or even building-level with object-level resolution and accuracy, such that a larger area can be represented while keeping track of objects and their affordances.
\Cref{fig:mobile_robot} shows the reconstruction and embedding integration in a larger office floor environment using a mobile robot. Using uniform density across the whole map, we are able to interactively identify regions spanning multiple rooms as well as individual objects.

\section{Conclusion}

% This work investigated different approaches for real-time integration of vision-language embeddings in 3D for open-set semantic segmentation.
This work investigated the real-time integration of vision-language embeddings into an open-set semantic 3D representation.
Our findings suggest that the segmentation performance mainly depends on the VLM backbone and the choice of 3D representation, and that the embedding granularity mainly determines the task space memory efficiency.
% Our analysis shows that recent patch-level embeddings perform better in terms of segmentation and runtime performance when integrated in 3D, than classic class-agnostic segment-embeddings. This removes the limitation on the granularity of the segments imposed by a pre-segmentation.

% We presented VLEM, a real-time framework for 3D integration and retrieval of vision-language embeddings as a building block for various interactive downstream tasks, and demonstrated its efficiency and versatility against baseline approaches in interactive robotic tasks.

We presented VLEM, an efficient real-time framework for the 3D integration and retrieval of vision-language embeddings, and demonstrated its efficiency and versatility as a building block for interactive robotic tasks covering manipulation and mobile platforms. VLEM provides a much more compact semantic environment representation while also providing superior segmentation performance.
% We presented VLEM, a real-time framework that integrates masked, pixel-aligned embeddings directly into a resampled 3D point cloud without requiring ground-truth camera poses or pre-training on the target environment.
% Compared to RayFronts and Open-Fusion, VLEM enables higher task space resolution and maintains real-time capability and memory efficiency, while achieving comparable or better performance using identical VLM backbones.

Our current approach of uniformly sampling the map and using segments prevents representing different regions at different levels of detail. VLEM, and its baselines, currently cannot represent relations between objects. Future work will focus on dynamic embedding granularity and take hierarchical and spatial relations into account.

\bibliography{rtvlem.bib}

\end{document}